\pdfoutput=1
\documentclass[11pt]{article}

\usepackage[]{acl}
\usepackage[normalem]{ulem}
\usepackage{times}
\usepackage{latexsym}
\usepackage{amssymb}
\usepackage{array}
\usepackage{booktabs} % For professional-looking tables
\usepackage{multirow} % For multirow cells
\usepackage{enumitem}
\usepackage[T1]{fontenc}
\usepackage{fontawesome5} % For icons like the pointing finger
\usepackage{xcolor}       % For colored text or symbols
\usepackage{ragged2e} 
\usepackage[utf8]{inputenc}

\usepackage{microtype}

\usepackage{inconsolata}

\usepackage{graphicx}
\usepackage{booktabs}
\usepackage{multirow}
\usepackage{graphicx}
\usepackage{xcolor}
\usepackage[many]{tcolorbox} 
\usepackage{setspace}               % for LINE SPACING
\usepackage{multicol}               % for MULTICOLUMNS
\usepackage{tikz}
\usepackage{amsmath}
\usepackage{amsmath,amssymb}%      pour les maths
\usepackage{enumitem}
\usepackage{varwidth}
\usepackage{paralist}
\usetikzlibrary{calc}
\usepackage{mdframed}
\usepackage{xcolor}

\definecolor{main}{HTML}{5989cf}    % setting main color to be used
\definecolor{sub}{HTML}{cde4ff}     % setting sub color to be used

\newcommand{\mistral}[1]{\textsc{Mistral}}
\newcommand{\tower}[1]{\textsc{TowerInstruct}}
\newcommand{\gemma}[1]{\textsc{Gemma}}
\newcommand{\hathi}[1]{\textsc{OpenHathi}}
\newcommand{\tamil}[1]{\textsc{Tamil-Llama}}
\newcommand{\kan}[1]{\textsc{Kan-Llama}}
\newcommand{\ctr}[1]{\textbf{CounterFact}}
\newcommand{\zsre}[1]{\textbf{ZsRE}}
\definecolor{RoyalBlue}{RGB}{65, 105, 225} % RGB values for RoyalBlue
\usepackage{graphicx}       % For scaling tables
\definecolor{scorelow}{HTML}{FFCCCC}  % Light red
\definecolor{scoremed}{HTML}{FFFF99}  % Light yellow
\definecolor{scorehigh}{HTML}{CCFF99} % Light green
\tcbset{
    sharp corners,
    colback = white,
    before skip = 0.1cm,    % add extra space before the box
    after skip = 0.2cm      % add extra space after the box
}   

\newmdenv[
  topline=false,
  bottomline=false,
  skipabove=\topsep,
  skipbelow=\topsep,
  leftline=true,
  rightline=true,
  linecolor=RoyalBlue,
  linewidth=2pt,
  innertopmargin=10pt,
  innerbottommargin=10pt,
  innerrightmargin=10pt,
  innerleftmargin=5pt,
  backgroundcolor=gray!10,
  roundcorner=10pt
]{stylishframe}

\newtcolorbox{boxH}{
    colback = sub, 
    colframe = main, 
    boxrule = 0pt, 
    leftrule = 1pt, % left rule weight
    left=1pt,
    right=3pt
}

\usepackage{xcolor}

\definecolor{etonblue}{rgb}{0.59, 0.78, 0.64}
\definecolor{lightblue}{rgb}{0.68, 0.85, 0.9}
\definecolor{lightgreen}{rgb}{0.56, 0.93, 0.56}
\definecolor{increase}{rgb}{0,0.5,0} 
\definecolor{Dandelion}{RGB}{240, 225, 48}

\mdfsetup{skipabove=\topskip,skipbelow=\topskip}
\newcounter{theo}[section]
\newenvironment{theo}[1][]{%
\stepcounter{theo}%
\ifstrempty{#1}%
 {\mdfsetup{%
   frametitle={%
    \tikz[baseline=(current bounding box.east),outer sep=0pt]
    \node[anchor=east,rectangle,fill=purple!60]
         {\strut Key observations};}}
 }%
{\mdfsetup{%
  frametitle={%
   \tikz[baseline=(current bounding box.east),outer sep=0pt]
   \node[anchor=east,rectangle,fill=purple!40]
        {\strut Prompt~\thetheo:~#1};}}%
 }%
\mdfsetup{innertopmargin=1pt,linecolor=purple!40,%
       linewidth=2pt,topline=true,
       frametitleaboveskip=\dimexpr-\ht\strutbox\relax,}
   \begin{mdframed}[]\relax%
}
{\end{mdframed}}

\title{Breaking Boundaries: Investigating the Effects of\\
Model Editing on Cross-linguistic Performance}

\author{%
  Somnath Banerjee$^\dagger$ 
  Avik Halder$^\dagger \thanks{These authors contributed equally to this work.}$ 
  Rajarshi Mandal$^\dagger \footnotemark[1]$ 
  Sayan Layek$^\dagger$\\
  \textbf{Ian Soboroff}$^\ddagger$
  \textbf{Rima Hazra}$^\mp$
  \textbf{Animesh Mukherjee}$^\dagger$\\
  $^\dagger$Indian Institute of Technology Kharagpur, India\\
  $^\ddagger$ National Institute of Standards and Technology, USA \thanks{Any mention of commercial is for information only; It does not imply recommendation or endorsement by NIST.}\\
  $^\mp$INSAIT, Sofia University ``St. Kliment Ohridski''\\
  \texttt{ \{som.iitkgpcse\}@kgpian.iitkgp.ac.in}\\
 }

\begin{document}
\maketitle
\begin{abstract}
% The integration of pretrained language models (PLMs) like BERT and GPT has revolutionized NLP, particularly for English, but it has also created linguistic imbalances. This paper strategically identifies the need for linguistic equity by examining several knowledge editing techniques in multilingual contexts. We evaluate the performance of models such as \mistral{}, \tower{}, \hathi{}, \tamil{}, and \kan{} across languages including English, German, French, Italian, Spanish, Hindi, Tamil, and Kannada. Our research identifies significant discrepancies in normal and merged models concerning cross-lingual consistency. We employ strategies like `each language for itself' (ELFI) and `each language for others' (ELFO) to stress-test these models. Our findings demonstrate the potential for LLMs to overcome linguistic barriers, laying the groundwork for future research in achieving linguistic inclusivity in AI technologies.
Pretrained language models (PLMs) have transformed natural language processing (NLP) but tend to exacerbate linguistic disparities in multilingual contexts. While earlier research has primarily focused on transformer-based models like BERT, this study shifts attention to large language models (LLMs) such as \mistral{}, \tower{}, \hathi{}, \tamil{}, and \kan{}. Through comprehensive evaluations across eight languages—including high-resource ones (English, German, French, Italian, Spanish) and low-resource ones (Hindi, Tamil, Kannada)—the research uncovers significant shortcomings in ensuring multilingual robustness and adaptability. Employing frameworks like ``each language for itself'' (ELFI) and ``each language for others" (ELFO), the analysis reveals that existing LLMs struggle to address linguistic inequities. Even strategies like model merging fail to close these gaps, highlighting fundamental deficiencies. These findings underscore the urgent need to redesign AI systems to achieve genuine linguistic inclusivity and balanced performance across diverse languages.
\end{abstract}

\section{Introduction}

\if{0}
In the dynamic and ever-evolving field of NLP, the emergence and integration of LLMs represent a watershed moment, heralding a new era of sophistication in how machines understand and generate human language. This revolution, sparked by the introduction of models such as BERT~\cite{devlin2019bert}, GPT~\cite{brown2020language}, and their successors, has not only expanded our computational toolkit but has also profoundly enriched our methodologies for harnessing and deploying factual knowledge within a myriad of applications. The pioneering work of parameter packing~\cite{roberts-etal-2020-much} and knowledge inheritance~\cite{qin-etal-2022-knowledge} has been instrumental in demonstrating the unparalleled efficacy of LLMs across diverse tasks, ranging from text generation to sentiment analysis, thereby cementing their status as quintessential instruments in the NLP arsenal.
  \begin{figure}
    \includegraphics[scale=0.23]{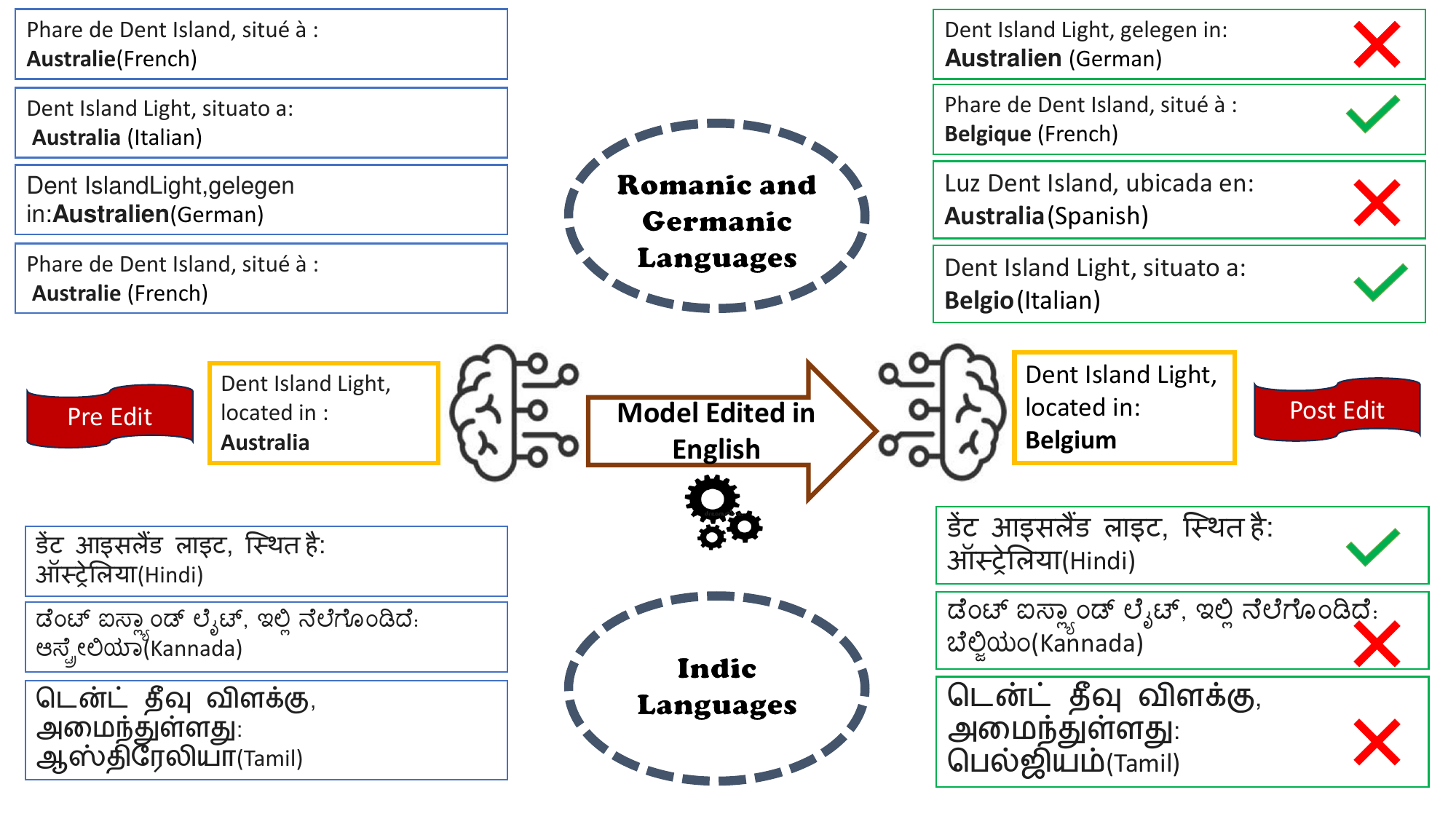} %[width=\linewidth]
    \caption{\label{fig:enhCon}Edited knowledge conflict across various languages for TowerInstruct.}
    %\vspace*{-0.6cm}
    %\vspace{-0.4cm}
  \end{figure}
However, this rapid advancement has also shone a spotlight on a critical oversight within the realm of NLP development: the predominant optimization of LLMs for English~\cite{huang2023transformerpatcher,meng2023massediting, zhao2023survey}. Such an anglocentric bias not only underscores a pressing issue of linguistic equity but also perpetuates a digital divide that disproportionately benefits English-speaking populations while marginalizing the linguistic and cultural diversity inherent to non-anglophone communities. The investigative efforts of X-FACTR~\cite{jiang-etal-2020-x} and M-LAMA~\cite{kassner-etal-2021-multilingual} have been pivotal in broadening our understanding of LLMs’ performance across a spectrum of languages, unveiling significant disparities in the treatment of factual knowledge that highlight the deep-seated issue of linguistic inequality entrenched in contemporary NLP technologies.
\fi

Handling multilinguality in language models remains a significant challenge, particularly when models are prompted in languages other than English. Tasks such as question answering~\cite{xu2024surveymultilinguallargelanguage}, addressing multilingual safety concerns~\cite{wang-etal-2024-languages, deng2024multilingual}, or performing knowledge edits~\cite{hazra-etal-2024-sowing} often reveal noticeable gaps in performance for low-resource languages. Despite the advancements in multilingual large language models (LLMs), disparities persist, especially for languages with fewer computational resources. A clear example of this issue arises in knowledge editing~\cite{Sinitsin2020Editable, de-cao-etal-2021-editing}. For instance, when an LLM is updated to correct a factual statement, ``\textit{\textbf{\textcolor{red}{The PM of the UK is Rishi Sunak}}}" to ``\textit{\textbf{\textcolor{teal}{The PM of the UK is Keir Starmer}}}" the model may apply the update accurately in well-represented languages like English or French~\cite{qi-etal-2023-cross,xu-etal-2023-language-anisotropic}. However, the same edit often fails to propagate when queried in low-resourced languages like Tamil or Hindi. This inconsistency highlights a critical weakness in the ability of LLMs to transfer factual updates across languages. Even advanced models like \mistral{} and \tower{}, while effective in European languages, struggle significantly with low-resource languages. This limitation undermines the broader goal of making language technologies universally accessible and equitable~\cite{wang2023crosslingual}.

\noindent This research aims to uncover the disparities in cross-lingual performance of LLMs to promote future linguistic inclusivity. While model editing techniques have advanced in monolingual settings, ensuring that factual updates made in one language are accurately reflected across others remains a major challenge~\cite{hazra-etal-2024-sowing,DBLP:journals/corr/abs-2402-15302}. This issue is particularly severe for low-resource languages, where models often fail to maintain reliability and consistency after edits. Such limitations reduce the utility of LLMs for these languages and widen existing linguistic inequities, leaving many communities underserved. Our work highlights these gaps, showing how current models struggle to manage multilingual updates, especially in underrepresented languages. By evaluating cross-lingual performance, we emphasize the need for more inclusive approaches to ensure that LLMs benefit users of all languages, not just those with abundant resources.

\noindent In this work, we conduct a comprehensive evaluation of how factual knowledge is transferred and maintained across eight linguistically diverse languages. We examine established knowledge editing techniques such as ROME~\cite{meng2022locating} and MEMIT~\cite{meng2023massediting} to assess their performance in multilingual contexts. Our research utilizes two strategies~\cite{10.1145/3511095.3531277}—``\textit{each language for itself}" (\textbf{ELFI}) and ``\textit{each language for others}" (\textbf{ELFO})—to rigorously test the ability of LLMs to preserve cross-lingual knowledge consistency. Through this evaluation, we reveal current models' limitations in maintaining consistent cross-lingual edits, emphasizing critical gaps to address for enhancing LLMs, particularly in low-resource languages. Our key contributions are as follows.\\ 
% \begin{compactitem}
%     \item We conduct extensive model editing experiments across eight languages—English (\textbf{En}), German (\textbf{De}), French (\textbf{Fr}), Italian (\textbf{It}), Spanish (\textbf{Es}), Hindi (\textbf{Hi}), Tamil (\textbf{Ta}), and Kannada (\textbf{Kn})—using \textbf{ELFI} and \textbf{ELFO}, focusing on decoder-only models' multilingual performance.
%     \item We evaluate 7B decoder-only models, including \mistral{}, \tower{}, \hathi{}, \tamil{}, and \kan{}, with editing methods \textbf{ROME} and \textbf{MEMIT}, advancing model editing research.
%     \item This is the first of it's kind work on LLM to reveal that model merging improves capabilities but struggles with cross-lingual consistency after editing.
% \end{compactitem}
\begin{stylishframe}
{\color{blue}\small\faHandPointRight} We conduct extensive model editing experiments across eight languages—English (\textbf{En}), German (\textbf{De}), French (\textbf{Fr}), Italian (\textbf{It}), Spanish (\textbf{Es}), Hindi (\textbf{Hi}), Tamil (\textbf{Ta}), and Kannada (\textbf{Kn})—using \textbf{ELFI} and \textbf{ELFO}, focusing on decoder-only models' multilingual performance.\\
{\color{blue}\small\faHandPointRight} We evaluate 7B decoder-only models, including \mistral{}, \tower{}, \hathi{}, \tamil{}, and \kan{}, with editing methods \textbf{ROME} and \textbf{MEMIT}, advancing model editing research.\\
{\color{blue}\small\faHandPointRight} This is the first of it's kind work on LLM to reveal that model merging improves capabilities but struggles with cross-lingual consistency after editing.
\end{stylishframe}
\section{Related work}
\noindent \textbf{Targeted parameter editing} modifies specific model components to integrate new information. \cite{dai-etal-2022-knowledge} introduced adjustments to `knowledge neurons' in transformers, while ROME~\cite{meng2022locating} updated neural weights to refresh LLM knowledge. MEMIT~\cite{meng2023massediting} expanded ROME for simultaneous updates, with further validation by~\cite{hase2023does,yao2023editing}.\\
\noindent \textbf{Multilingual knowledge editing} remains limited, focusing mainly on translating English prompts. X-FACTR~\cite{jiang-etal-2020-x} and M-LAMA~\cite{kassner-etal-2021-multilingual} exposed large knowledge gaps in non-English languages, often with $<10\%$ accuracy. GeoMLAMA~\cite{yin2022geomlama} revealed that native languages may not best access national knowledge.
\dashuline{We analyze cross-lingual consistency in multilingual LLMs, extending prior work mostly on BERT (pre LLM era) to diverse LLMs fine-tuned for specific languages}~\cite{wang2023crosslingual,beniwal-etal-2024-cross}.

\section{Task overview}
\noindent\textbf{Model editing}: Given a language model $\theta_{pre}$ and an edit descriptor <$kn$, $a_{new}$, $a_{old}$>, the model editing technique will create an edited model $\theta_{edit}$. So, for an input prompt $kn$, $\theta_{pre}$ has the old prediction $a_{old}$ and after editing $\theta_{pre}$, the edited model $\theta_{edit}$ has updated prediction $a_{new}$ without influencing model behaviour on other samples. Thus, given the edit input $kn$, $\theta_{pre}$ does not produce $a_{new}$; it is $\theta_{edit}$ that is designed to produce the output $a_{new}$.
\begin{equation}
\small
{\theta_{edit}}(kn) = 
\begin{cases}
a_{new} & \text{if } kn \in I(kn, a_{new}) \\
{\theta_{pre}}(kn) & \text{if } kn \in O(kn, a_{new})
\end{cases}
\end{equation}
% The scope of consideration $I(kn, a_{new})$ is generally inclusive of $kn$ as well as its equivalence neighborhood, which encompasses pertinent input/output pair associations.  The neighborhood contains rephrased prompts version of actual prompt $kn$. 
The scope of consideration, $I(kn, a_{new})$, includes $kn$ and similar versions of it. This means it covers the original input and any rephrased versions of it that still relate to the same topic. For example, if $kn$ is a question, this scope includes different ways of asking the same question. However, the excluded scope, $O(kn, a_{new})$, refers to inputs that are not related to the edit case provided. So, it leaves out any inputs that do not have anything to do with $kn$ or its related versions.
%\am{This is not clear; best is to give an example.}
%On the other hand, the scope excluded, $O(kn, a_{new})$, pertains to inputs that are unrelated with the edit case provided.
%The in-scope $I(kn, a_{new})$ usually encompasses $kn$ along with its equivalence neighborhood $\mathcal{N}(kn, a_{new})$, which includes related input/output pairs. In contrast, the out-of-scope $\mathcal{O}(kn, a_{new})$ consists of inputs that are unrelated to the edit example. 
Along with the updated information, the edited model should follow the four properties: \textcolor{red}{(i)} \textit{\textbf{reliability}} -- $\theta_{edit}$, produces the correct response for the specific edit scenario represented by ($kn$, $a_{new}$), \textcolor{red}{(ii)} \textit{\textbf{generalization}} -- the edited model $\theta_{edit}$ must uniformly apply edits to both the designated edit case ($kn$, $a_{new}$) and its semantically equivalent variations, guaranteeing a consistent output, $a_{new}$, across all rephrased iterations of $kn$, \textcolor{red}{(iii)} \textit{\textbf{locality}} -- $\theta_{edit}$ should not alter the output for examples outside its intended scope (O($kn$, $a_{new}$)), and \textcolor{red}{(iv)} \textit{\textbf{portability}} -- evaluates the capacity of edited model $\theta_{edit}$ for robust generalization, assessed through questions designed to test the edited model's reasoning with updated knowledge.

\if{0}\subsubsection{Reliability}
Recent studies by ~\cite{de-cao-etal-2021-editing, meng2022locating, HuangSZZR023} determine the reliability of an edit based on whether the edited model, denoted as $\theta_{edit}$, produces the correct response for the specific edit scenario represented by ($kn$, $a_{new}$). The reliability metric is quantified through the mean accuracy obtained for these edit instances. So, for edit case ($kn$, $a_{new}$) as input the edited model needs to give the $a_{new}$ as generated output. For example, the edited model must generate the consistent edited answer in any language, i.e., \emph{``2006''} for the prompt ~\emph{``When was the inception of IAAF Combined Events Challenge?''}.

% Previous works (Huang et al., 2023; De Cao et al., 2021; Meng et al., 2022) define a reliable edit when the post-edit model f gives the target answer for the case (xe, ye) to be edited. The reliability is measured as the average accuracy on the edit case:
% \begin{equation}
% \mathbb{E}_{x'_e,y'_e \sim (x_e,y_e)} \mathbb{1} \{ \arg\max_{y} f_{\theta_e} (y | x'_e) = y'_e \}
% \end{equation}

\subsubsection{Generalization}
The generalization property mandates that the edited model $\theta_{edit}$ must uniformly apply edits to both the designated edit case ($kn$, $a_{new}$) and its semantically equivalent variations, guaranteeing a consistent output, $a_{new}$, across all rephrased iterations of $kn$. For example, the edited model $\theta_{edit}$ must yield the consistent answer \emph{``2006''} to both the original prompt ~\emph{``When was the inception of the IAAF Combined Events Challenge?''} and its rephrased counterpart \emph{``When was the IAAF Combined Events Challenge launched?''}.

% \begin{equation}
% \mathbb{E}_{x'_e,y'_e \sim \mathcal{N}(x_e,y_e)} \mathbb{1} \left\{ \arg\max_{y} f_{\theta_e} (y | x'_e) = y'_e \right\}
% \end{equation}

\subsubsection{Locality}
The principle of locality, sometimes called specificity, emphasizes that modifications should only affect targeted areas, leaving the outcomes for non-targeted examples unchanged in the post-edit model $\theta_{edit}$. This means $\theta_{edit}$ should not alter the output for examples outside its intended scope (O($kn$, $a_{new}$)). The degree of locality is measured by how often $\theta_{edit}$ maintains the original model $\theta_{pre}$'s predictions for these outside examples.

% Locality also noted as Specificity in some work. Editing should be implemented locally, which means the post-edit model fe should not change the output of the irrelevant examples in the out-of scope O(xe, ye). Hence, the locality is evaluated by the rate at which the post-edit model fe’s predictions are unchanged as the pre-edit f model:

% \begin{equation}
% \mathbb{E}_{x'_e,y'_e \sim \mathcal{O}(x_e,y_e)} \mathbb{1} \left\{ f_{\theta_e} (y | x'_e) = f_{\theta} (y | x'_e) \right\}
% \end{equation}

\subsubsection{Portability}
Portability evaluates the capacity of edited model $\theta_{edit}$ for robust generalization, assessed through questions designed to test the edited model's reasoning with updated knowledge. Given such a portability prompt, edited model is expected to provide the actual answer, affirming its ability to learn the knowledge as opposed to merely adapting to trivial lexical adjustments. For instance, if the model is edited with prompt ~\emph{``When was the inception of IAAF Combined Events Challenge?''} and answer ~\emph{2006}, the portability prompt ~\emph{``What type of sports event is the IAAF Combined Events Challenge, which was established in 2006?''} to the model should provide the answer ~\emph{Athletics} by reasoning.\fi
\noindent\textbf{Multilingual knowledge editing}: Given a set of languages $\mathcal{L}$, we consider a language $l \in \mathcal{L}$ to edit the model $\theta_{pre}$ and obtain $\theta_{edit}^{l}$. We then test the edited model $\theta_{edit}^{l}$ with all the languages in $\mathcal{L}$. In the equations below, $s$ is the source language, and $t$ is the target language. The conditions are as follows: if $kn_s$ is in the inclusion scope $I(kn, a_{new})$, the model should output $a_{new}^s$. Otherwise, if $kn_s$ is in the exclusion scope $O(kn, a_{new})$, the model should output $\theta_{pre}(kn_s)$. For the target language, similar conditions apply with transformations $\mathcal{T}^t$.
\begin{equation}
\small
{\theta_{edit}}(kn_s) = 
\begin{cases}
a_{new}^s & \text{if } kn_s \in I(kn, a_{new}) \\
\theta_{pre}(kn_s) & \text{if } kn_s \in O(kn, a_{new})
\end{cases}
\end{equation}
\begin{equation}
\small
\theta_{edit}(kn_t) = 
\begin{cases}
\mathcal{T}^t(a_{new}^s) & \text{if } kn_t \in \mathcal{T}^t(I(kn, a_{new})) \\
\theta_{pre}(kn_t) & \text{if } kn_t \notin \mathcal{T}^t(O(kn, a_{new}))
\end{cases}
\end{equation}
$\mathcal{T}^t(.)$ transforms the target output of the source language to the target language with the same meaning. Therefore, after editing the model in one language, such as English, the effect of the edit should be reflected in other languages as well. This ensures that the specific edit is consistent across all languages, regardless of the language in which the edit was made.\\
\if{0}\subsection{Evaluation of the edit properties}
We assess edit consistency across languages by testing each edit made in language \(l\) on the edited model \(\theta_{\text{edit}}^l\), using knowledge from other languages \(\mathcal{L} \). This evaluation are used on four key edit properties. The evaluation involves translation  ($\mathcal{T}(.)$) of both ground truth and generated outputs into English (\textbf{En}), further employ the evaluation metrics on them.\fi
\noindent\textbf{Model merging}: In the specific case of Indic languages -- Hindi, Tamil and Kannada -- we have specialized LLMs for each unlike in the case of Western languages where the models we have used are known to be pretrained on all those languages. We investigate if the three LLMs for the Indic languages could be further unified to obtain a more powerful model \(\theta_{merged}\), which dynamically harnesses the specialized linguistic capabilities of each constituent models. %context of merging three large language models individually trained on Hindi (\textbf{Hi}), Tamil (\textbf{Ta}), and Kannada (\textbf{Kn}), we aim to create a unified model,  
This involves extracting language-specific unique task vectors from instruction-tuned models, i.e., \(\theta_{base-Hindi} \rightarrow \vec{v}_{Hindi}\), \(\theta_{base-Tamil} \rightarrow \vec{v}_{Tamil}\), and \(\theta_{base-Kannada} \rightarrow \vec{v}_{Kannada}\) for each respective language. These vectors are integrated using a TIES~\cite{yadav2023tiesmerging} merging technique to synthesize \(\theta_{merged}\). Subsequently, \(\theta_{merged}\) is edited in the same process as above to obtain \(\theta_{edit}\) each time adjusting its output specifically for inputs associated with the defined task and the language. %Mathematically, \(\theta_{merged}(x, lang)\) is configured to apply the appropriate model based on the language context of the input, represented as \(\theta_{merged}(langs) = \theta_{base} + \lambda.\vec{v}_{merged}\), where \(langs\) specifies targeted languages like Hindi, Tamil, or Kannada and \(\vec{v}_{merged}\) represents merged task vectors using TIES and \(\lambda\) is scaling hyperparameter. Following this, we engage in a precise editing process using edit descriptors \(<kn, a_{new}, a_{old}, lang>\) to refine \(\theta_{merged}\) into \(\theta_{edit}\), adjusting its output .
% For every edit with language $l$, the same knowledge in other languages ($\mathcal{L}\setminus l$ are tested on edited model $\theta{edit}^l$. This is conducted for four properties of editing to measure consistency of editing in other languages. For assessing the alignment between generated outputs and the ground truth, both the ground truth and the generated outputs are translated into English. We then verify if the ground truth is encapsulated within the generated output, indicating a successful edit. 

% \begin{equation}
% \theta_{edit}^l(kn^{l_i})
% \forall \text{lang} \in \text{Languages}
% \end{equation}

\section{Dataset}
For our experiments, we use the popular \ctr{}~\cite{meng2022locating} and \zsre{}~\cite{levy-etal-2017-zero} datasets. We uniformly sample $\sim$ 550 edit instances from each dataset. Each edit instance in these datasets includes the actual edit case, the reliability prompt, the generalization instances, the locality prompt and its answer, portability and its answer. Further we use google translator~\footnote{https://translate.google.com/} to translate each edit instance into seven other languages --  German (\textbf{De}), French (\textbf{Fr}), Italian (\textbf{It}), Spanish (\textbf{Es}), Hindi (\textbf{Hi}), Tamil (\textbf{Ta}) and Kannada (\textbf{Kn}). In both the datasets, the actual portability prompt is an interrogative sentence (i.e., in the form of question). However, when the question gets translated to other languages, the translated question becomes different from actual question format. For example, when the actual portability prompt in English ``To which language family does the official language of Sastamala  belong?'' is translated to French the new prompt becomes ``\`A quelle langue la famille appartient la langue officielle de Sastamala?''. However when this is back-translated to English the prompt means ``Which family language does the official language of Sastamala belong to?'' which is not the same as the original English prompt. We therefore employed GPT-4\footnote{openai.com/research/gpt-4, version: gpt-4-0125-preview} to convert question in the interrogative sentence into a task of sentence completion. Subsequently we translate this sentence completion form to other languages to obtain the corresponding portability prompt.\\
\noindent\textbf{Note to the choice of languages}: The Western languages that we choose are based on their cultural, economic and academic significance~\cite{Lobachev_2008}\footnote{https://preply.com/en/blog/most-important-languages/} and cover the Romance and the Germanic families. In addition, we include three Indic languages that have far lesser resources compared to their Western counterparts.

\section{Experimental setup}
\subsection{Selection of LLMs}

We use the following multilingual LLMs for our experiments:\\
\noindent \textbf{Mistral-7B-Instruct-v0.2}~(\mistral{})\footnote{huggingface.co/mistralai/Mistral-7B-Instruct-v0.2}: A multilingual causal language model~\cite{jiang2023mistral}, supporting diverse languages\footnote{https://encord.com/blog/mistral-large-explained/}.\\
\noindent \textbf{TowerInstruct-7B-v0.2}~(\tower{})\footnote{huggingface.co/Unbabel/TowerInstruct-7B-v0.2}: Based on LLaMA2~\cite{touvron2023llama}, supports multilinguality across 10 languages, including English, German, and Chinese.\\
\noindent \textbf{OpenHathi-7B-Hi-v0.1-Base}~(\hathi{})\footnote{huggingface.co/sarvamai/OpenHathi-7B-Hi-v0.1-Base}: Optimized for Indian languages like Hindi and Tamil using a GPT-3-like transformer with hybrid partitioned attention.\\
\noindent \textbf{Tamil-llama-7b-base-v0.1}~(\tamil{})\footnote{huggingface.co/abhinand/tamil-llama-7b-base-v0.1}: A bilingual Tamil-English model~\cite{balachandran2023tamilllamanewtamillanguage} using a 7B-parameter causal language framework.\\
\noindent \textbf{Kan-LLaMA-7B-SFT}~(\kan{})\footnote{huggingface.co/Tensoic/Kan-Llama-7B-SFT-v0.5}: Specialized in Kannada with a 49,420-token vocabulary, pre-trained on 600M tokens from CulturaX using low-rank adaptation.
More details on models are in Appendix~\ref{sec:modselection}.
\begin{table}[h]
\centering
\resizebox{0.48\textwidth}{!}{
\begin{tabular}{c|ccccc|cccc}
\hline
\multicolumn{1}{l|}{\multirow{3}{*}{\textbf{Languages}}} & \multicolumn{5}{c|}{\ctr}                                                                                                                                              & \multicolumn{4}{c}{\zsre}                                                                                                      \\ \cline{2-10} 
\multicolumn{1}{l|}{}                                    & \multicolumn{1}{c|}{\textbf{Models}}  & \multicolumn{2}{c|}{\tower}                                         & \multicolumn{2}{c|}{\mistral}                            & \multicolumn{2}{c|}{\tower}                                         & \multicolumn{2}{c}{\mistral}                             \\ \cline{2-10} 
\multicolumn{1}{l|}{}                                    & \multicolumn{1}{c|}{\textbf{Metrics}} & \multicolumn{1}{c|}{\textbf{RO}} & \multicolumn{1}{c|}{\textbf{ME}} & \multicolumn{1}{c|}{\textbf{RO}} & \textbf{ME}           & \multicolumn{1}{c|}{\textbf{RO}} & \multicolumn{1}{c|}{\textbf{ME}} & \multicolumn{1}{c|}{\textbf{RO}} & \textbf{ME}           \\ \hline
\multirow{4}{*}{\textbf{De}}                             & \multicolumn{1}{c|}{\textbf{Rel}}     & 0.83/0.96                        & 0.73/0.83                        & 0.83/0.96                        & 0.73/0.87             & 0.48/0.59                        & 0.25/0.30                        & 0.51/0.62                        & 0.38/0.47             \\
                                                         & \multicolumn{1}{c|}{\textbf{Gen}}     & 0.27/0.31                        & 0.19/0.22                        & 0.28/0.31                        & 0.19/0.22             & 0.33/0.39                        & 0.11/0.12                        & 0.35/0.45                        & 0.18/0.24             \\
                                                         & \multicolumn{1}{c|}{\textbf{Loc}}     & \colorbox{magenta!70}{0.22}/0.23                        & 0.19/0.22                        & 0.21/0.23                        & 0.24/0.27             & 0.00/0.01                        & 0.00/0.01                        & 0.01/0.02            & 0.01/0.03 \\
                                                         & \multicolumn{1}{c|}{\textbf{Port}}    & 0.01/0.01                        & 0.01/0.01                        & {0.03/0.04}                  & \textbf{\colorbox{magenta!70}{0.04}}/0.06    & 0.02/0.02                        & 0.00/0.00                        & \textbf{\colorbox{cyan!70}{0.08}}/0.10               & 0.02/0.04             \\ \hline
\multirow{4}{*}{\textbf{Es}}                             & \multicolumn{1}{c|}{\textbf{Rel}}     & 0.82/0.92                        & 0.70/0.80                        & 0.81/0.91                        & 0.78/0.86             & 0.44/0.59                        & 0.24/0.34                        & 0.49/0.61                        & 0.37/0.49             \\
                                                         & \multicolumn{1}{c|}{\textbf{Gen}}     & 0.33/0.37            & 0.23/0.27                        & 0.28/0.32                        & 0.22/0.27             & 0.30/0.40                        & 0.16/0.20                        & 0.35/0.45                        & 0.22/0.29             \\
                                                         & \multicolumn{1}{c|}{\textbf{Loc}}     & 0.21/0.22                        & 0.19/0.19                        & 0.25/0.27                        & \textbf{\colorbox{magenta!70}{0.27}}/0.29    & 0.00/0.01                        & 0.01/0.02            & 0.01/0.01            & \textbf{0.02}/0.02    \\
                                                         & \multicolumn{1}{c|}{\textbf{Port}}    & 0.00/0.00                        & 0.00/0.00                        & 0.03/0.03            & 0.03/0.04 & 0.02/0.02                        & 0.01/0.02                        & 0.03/0.07                        & 0.03/0.04             \\ \hline
\multirow{4}{*}{\textbf{It}}                             & \multicolumn{1}{c|}{\textbf{Rel}}     & \textbf{\colorbox{magenta!70}{0.87}}/0.93               & 0.74/0.78                        & \colorbox{magenta!70}{0.86}/0.91            & 0.80/0.88             & \underline{\colorbox{cyan!70}{0.54}}/0.62            & 0.25/0.29                        & \textbf{\colorbox{cyan!70}{0.58}}/0.65               & 0.42/0.50             \\
                                                         & \multicolumn{1}{c|}{\textbf{Gen}}     & \textbf{\colorbox{magenta!70}{0.35}}/0.38               & 0.25/0.26                        & 0.28/0.30                        & 0.24/0.27             & \colorbox{cyan!70}{0.35}/0.43                        & 0.16/0.20                        & \textbf{\colorbox{cyan!70}{0.42}}/0.48               & 0.25/0.31             \\
                                                         & \multicolumn{1}{c|}{\textbf{Loc}}     & 0.18/0.19                        & 0.20/0.20                        & 0.26/0.27            & \textbf{\colorbox{magenta!70}{0.27}}/0.28    & 0.00/0.00                        & 0.00/0.01                        & 0.00/0.02                        & 0.01/0.02 \\
                                                         & \multicolumn{1}{c|}{\textbf{Port}}    & \colorbox{magenta!70}{0.02}/0.02                        & \colorbox{magenta!70}{0.02}/0.03                        & 0.02/0.03                        & 0.03/0.03 & 0.01/0.02                        & 0.02/0.03                        & 0.07/0.08            & 0.01/0.03             \\ \hline
\multirow{4}{*}{\textbf{Fr}}                             & \multicolumn{1}{c|}{\textbf{Rel}}     & 0.83/0.90                        & 0.65/0.72                        & 0.83/0.89                        & 0.79/0.85             & 0.51/0.59                        & 0.27/0.35                        & 0.52/0.63                        & 0.40/0.50             \\
                                                         & \multicolumn{1}{c|}{\textbf{Gen}}     & 0.31/0.33                        & 0.22/0.24                        & \colorbox{magenta!70}{0.29}/0.30                        & 0.24/0.25             & 0.28/0.35                        & 0.14/0.17                        & 0.40/0.50            & 0.19/0.27             \\
                                                         & \multicolumn{1}{c|}{\textbf{Loc}}     & 0.21/0.22                        & 0.17/0.19                        & 0.20/0.22                        & 0.24/0.25             & 0.00/0.01                        & 0.00/0.02                        & 0.01/0.02            & 0.01/0.02 \\
                                                         & \multicolumn{1}{c|}{\textbf{Port}}    & 0.00/0.01                        & 0.00/0.00                        & 0.03/0.03            & 0.03/0.03 & \colorbox{cyan!70}{0.03}/0.05                        & \colorbox{cyan!70}{0.03}/0.03                        & 0.06/0.09                        & 0.04/0.06             \\ \hline
\end{tabular}
}
\caption{\footnotesize Comparison of reliability, generalization, locality, and portability scores across language models under \textit{Self edit - self inference} settings. The highest scores for individual metrics in ROME and MEMIT are highlighted in magenta for CounterFact and in cyan for ZSRE, with values shown as Exact Match/Partial Match.}
%\vspace{-0.3cm}
\label{tab:my-table1}
\end{table}
\begin{table}[h]
\centering
\resizebox{0.48\textwidth}{!}{
\begin{tabular}{c|ccccc|cccc}
\hline
\multicolumn{1}{l|}{\multirow{3}{*}{\textbf{Languages}}} & \multicolumn{5}{c|}{\ctr}                                                                                                                                              & \multicolumn{4}{c}{\zsre}                                                                                                      \\ \cline{2-10} 
\multicolumn{1}{l|}{}                                    & \multicolumn{1}{c|}{\textbf{Models}}  & \multicolumn{2}{c|}{\tower}                                         & \multicolumn{2}{c|}{\mistral}                            & \multicolumn{2}{c|}{\tower}                                         & \multicolumn{2}{c}{\mistral}                             \\ \cline{2-10} 
\multicolumn{1}{l|}{}                                    & \multicolumn{1}{c|}{\textbf{Metrics}} & \multicolumn{1}{c|}{\textbf{RO}} & \multicolumn{1}{c|}{\textbf{ME}} & \multicolumn{1}{c|}{\textbf{RO}} & \textbf{ME}           & \multicolumn{1}{c|}{\textbf{RO}} & \multicolumn{1}{c|}{\textbf{ME}} & \multicolumn{1}{c|}{\textbf{RO}} & \textbf{ME}           \\ \hline
\multirow{4}{*}{\textbf{De}}                             & \multicolumn{1}{c|}{\textbf{Rel}}     & 0.48/0.53                        & 0.40/0.46                        & 0.50/0.56                        & 0.54/0.61             & \colorbox{cyan!70}{0.24}/0.28                        & 0.10/0.14                        & 0.34/0.45                        & 0.14/0.18             \\
                                                         & \multicolumn{1}{c|}{\textbf{Gen}}     & 0.25/0.27                        & 0.13/0.17                        & 0.23/0.27                        & 0.22/0.23             & \colorbox{cyan!70}{0.18}/0.23                        & 0.12/0.14                        & 0.26/0.35                        & 0.14/0.16             \\
                                                         & \multicolumn{1}{c|}{\textbf{Loc}}     & 0.20/0.21                        & 0.19/0.22                        & 0.23/0.25                        & 0.26/0.28 & 0.00/0.01                        & 0.00/0.02                        & 0.01/0.02            & 0.01/0.03 \\
                                                         & \multicolumn{1}{c|}{\textbf{Port}}    & 0.00/0.00                        & 0.00/0.00                        & { 0.03/0.03}      & 0.03/0.04 & 0.02/0.02                        & 0.02/0.02                        & \underline{0.06}/0.07            & 0.02/0.03             \\ \hline
\multirow{4}{*}{\textbf{Es}}                             & \multicolumn{1}{c|}{\textbf{Rel}}     & \colorbox{magenta!70}{0.51}/0.56                        & 0.40/0.48                        & \textbf{\colorbox{magenta!70}{0.57}}/0.62               & 0.56/0.60 & \colorbox{cyan!70}{0.24}/0.29                        & 0.12/0.14                        & \textbf{\colorbox{cyan!70}{0.39}}/0.48               & 0.19/0.26             \\
                                                         & \multicolumn{1}{c|}{\textbf{Gen}}     & 0.26/0.29            & 0.18/0.22                        & 0.25/0.29                        & 0.21/0.26             & \colorbox{cyan!70}{0.18}/0.25                        & 0.09/0.11                        & \textbf{\colorbox{cyan!70}{0.33}}/0.41               & 0.14/0.21             \\
                                                         & \multicolumn{1}{c|}{\textbf{Loc}}     & 0.22/0.24                        & 0.17/0.17                        & 0.24/0.27                        & 0.25/0.27             & 0.00/0.01                        & 0.01/0.02            & 0.01/0.02            & \textbf{0.02}/0.02    \\
                                                         & \multicolumn{1}{c|}{\textbf{Port}}    & 0.00/0.00                        & 0.00/0.00                        & 0.03/0.03            & 0.03/0.04 & 0.02/0.03                        & 0.01/0.01                        & 0.04/0.06                        & 0.04/0.05             \\ \hline
\multirow{4}{*}{\textbf{It}}                             & \multicolumn{1}{c|}{\textbf{Rel}}     & 0.45/0.50                        & 0.35/0.40                        & 0.47/0.58                        & 0.44/0.49             & \colorbox{cyan!70}{0.24}/0.29                        & 0.12/0.14                        & 0.31/0.34                        & 0.23/0.27             \\
                                                         & \multicolumn{1}{c|}{\textbf{Gen}}     & 0.23/0.27                        & 0.19/0.20                        & 0.25/0.35                        & 0.21/0.23             & 0.17/0.22                        & 0.11/0.13                        & 0.26/0.32                        & 0.18/0.21             \\
                                                         & \multicolumn{1}{c|}{\textbf{Loc}}     & 0.20/0.21                        & 0.20/0.20                        & 0.24/0.36                        & \textbf{\colorbox{magenta!70}{0.28}}/0.29    & 0.00/0.00                        & 0.00/0.01                        & 0.00/0.02                        & \underline{0.01}/0.02 \\
                                                         & \multicolumn{1}{c|}{\textbf{Port}}    & 0.01/0.02                        & 0.01/0.02                        & 0.03/0.11            & 0.04/0.04    & 0.01/0.02                        & 0.02/0.02                        & \textbf{0.07}/0.08               & 0.01/0.01             \\ \hline
\multirow{4}{*}{\textbf{Fr}}                             & \multicolumn{1}{c|}{\textbf{Rel}}     & 0.50/0.53                        & 0.45/0.49                        & 0.49/0.55                        & 0.51/0.59             & 0.22/0.26                        & 0.12/0.17                        & 0.36/0.44            & 0.23/0.28             \\
                                                         & \multicolumn{1}{c|}{\textbf{Gen}}     & \textbf{\colorbox{magenta!70}{0.28}}/0.31               & 0.19/0.22                        & \textbf{\colorbox{magenta!70}{0.28}}/0.31               & 0.26/0.27 & 0.15/0.21                        & 0.08/0.10                        & 0.29/0.33            & 0.16/0.21             \\
                                                         & \multicolumn{1}{c|}{\textbf{Loc}}     & \colorbox{magenta!70}{0.23}/0.23                        & 0.19/0.21                        & 0.20/0.36                        & 0.25/0.26             & 0.00/0.01                        & 0.00/0.02                        & 0.01/0.03            & 0.01/0.02 \\
                                                         & \multicolumn{1}{c|}{\textbf{Port}}    & 0.01/0.01                        & 0.01/0.01                        & 0.01/0.12                        & 0.03/0.04 & 0.02/0.02                        & 0.02/0.02                        & 0.06/0.09            & 0.04/0.05             \\ \hline
\end{tabular}
}
\caption{\footnotesize Comparison of reliability, generalization, locality, and portability scores across language models under \textit{English edit - self inference} settings. The highest scores for individual metrics in ROME and MEMIT are highlighted in magenta for CounterFact and in cyan for ZSRE, with values shown as Exact Match/Partial Match.}
%\vspace{-0.3cm}
\label{tab:my-table2}
\end{table}
\begin{table}[h]
\centering
\resizebox{0.48\textwidth}{!}{
\begin{tabular}{c|ccccc|cccc}
\hline
\multirow{3}{*}{\textbf{\begin{tabular}[c]{@{}c@{}}Languages/\\ Models\end{tabular}}} & \multicolumn{1}{l}{}                  & \multicolumn{4}{c|}{\textit{self edit - self inference}}                                                                                 & \multicolumn{4}{c}{(\textit{English edit - self inference})}                                                                   \\ \cline{3-10} 
                                                                                      &                                       & \multicolumn{2}{c|}{\ctr}                                                     & \multicolumn{2}{c|}{\zsre}                               & \multicolumn{2}{c|}{\ctr}                                           & \multicolumn{2}{c}{\zsre}                                \\ \cline{2-10} 
                                                                                      & \multicolumn{1}{c|}{\textbf{Metrics}} & \multicolumn{1}{c|}{\textbf{RO}} & \multicolumn{1}{c|}{\textbf{ME}}           & \multicolumn{1}{c|}{\textbf{RO}} & \textbf{ME}           & \multicolumn{1}{c|}{\textbf{RO}} & \multicolumn{1}{c|}{\textbf{ME}} & \multicolumn{1}{c|}{\textbf{RO}} & \textbf{ME}           \\ \hline
\multirow{4}{*}{\textbf{\begin{tabular}[c]{@{}c@{}}Hi/\\ \hathi{}\end{tabular}}}      & \multicolumn{1}{c|}{\textbf{Rel}}     & 0.02/0.02                        & \multicolumn{1}{c|}{\underline{0.45}/0.60} & 0.03/0.06                        & \textbf{0.20}/0.33    & \textbf{0.56}/0.66               & \underline{0.02}/0.03            & \textbf{0.03}/0.03               & \textbf{0.03}/0.06    \\
                                                                                      & \multicolumn{1}{c|}{\textbf{Gen}}     & 0.00/0.00                        & \multicolumn{1}{c|}{\textbf{0.26}/0.33}    & 0.01/0.04                        & \textbf{0.19}/0.28    & \textbf{0.27}/0.34               & \underline{0.03}/0.03            & \underline{0.03}/0.03            & \textbf{0.04}/0.08    \\
                                                                                      & \multicolumn{1}{c|}{\textbf{Loc}}     & \textbf{0.31}/0.35               & \multicolumn{1}{c|}{0.02/0.03}             & \textbf{0.01}/0.01               & \underline{0.00}/0.01 & \textbf{0.26}/0.31               & \underline{0.03}/0.03            & \textbf{0.00}/0.00               & \textbf{0.00}/0.01    \\
                                                                                      & \multicolumn{1}{c|}{\textbf{Port}}    & \textbf{0.01}/0.01               & \multicolumn{1}{c|}{\textbf{0.01}/0.01}    & 0.00/0.00                        & \textbf{0.03}/0.03    & \textbf{0.02}/0.02               & \underline{0.00}/0.01            & \underline{0.00}/0.00            & \textbf{0.01}/0.01    \\ \hline
\multirow{4}{*}{\textbf{\begin{tabular}[c]{@{}c@{}}Ta/\\ \tamil{}\end{tabular}}}      & \multicolumn{1}{c|}{\textbf{Rel}}     & 0.12/0.15                        & \multicolumn{1}{c|}{\textbf{0.48}/0.59}    & 0.06/0.08                        & \underline{0.16}/0.21 & 0.00/0.00                        & 0.01/0.01                        & 0.00/0.00                        & \underline{0.01}/0.01 \\
                                                                                      & \multicolumn{1}{c|}{\textbf{Gen}}     & 0.03/0.04                        & \multicolumn{1}{c|}{\underline{0.21}/0.25} & 0.03/0.04                        & \underline{0.10}/0.14 & 0.00/0.00                        & 0.00/0.00                        & 0.00/0.00                        & 0.00/0.00             \\
                                                                                      & \multicolumn{1}{c|}{\textbf{Loc}}     & 0.01/0.01                        & \multicolumn{1}{c|}{0.01/0.01}             & \underline{0.00}/0.00            & \underline{0.00}/0.00 & 0.01/0.01                        & 0.01/0.02                        & \textbf{0.00}/0.00               & \textbf{0.00}/0.00    \\
                                                                                      & \multicolumn{1}{c|}{\textbf{Port}}    & \textbf{0.01}/0.01               & \multicolumn{1}{c|}{\textbf{0.01}/0.01}    & 0.00/0.00                        & \underline{0.01}/0.01 & \underline{0.00}/0.00            & \underline{0.00}/0.00            & \underline{0.00}/0.00            & \underline{0.00}/0.00 \\ \hline
\multirow{4}{*}{\textbf{\begin{tabular}[c]{@{}c@{}}Kn/\\ \kan{}\end{tabular}}}        & \multicolumn{1}{c|}{\textbf{Rel}}     & 0.21/0.26                        & \multicolumn{1}{c|}{0.14/0.18}             & 0.16/0.21                        & 0.05/0.07             & 0.01/0.01                        & 0.00/0.00                        & 0.00/0.01                        & 0.00/0.01             \\
                                                                                      & \multicolumn{1}{c|}{\textbf{Gen}}     & 0.07/0.08                        & \multicolumn{1}{c|}{0.04/0.05}             & 0.08/0.17                        & 0.05/0.05             & 0.00/0.01                        & 0.00/0.00                        & 0.00/0.00                        & 0.00/0.00             \\
                                                                                      & \multicolumn{1}{c|}{\textbf{Loc}}     & \underline{0.03}/0.04            & \multicolumn{1}{c|}{0.02/0.03}             & \underline{0.00}/0.00            & \underline{0.00}/0.00 & 0.02/0.02                        & \underline{0.03}/0.03            & \textbf{0.00}/0.00               & \textbf{0.00}/0.00    \\
                                                                                      & \multicolumn{1}{c|}{\textbf{Port}}    & \underline{0.00}/0.00            & \multicolumn{1}{c|}{\underline{0.00}/0.01} & 0.00/0.01                        & 0.00/0.00             & \underline{0.00}/0.00            & \underline{0.00}/0.00            & \underline{0.00}/0.00            & \underline{0.00}/0.00 \\ \hline
\end{tabular}
}
\caption{\footnotesize Comparison of scores in indic language models. Highest scores are in bold, second-highest underlined, with values shown as Exact Match/Partial Match.}
%\vspace{-0.5cm}
\label{tab:my-table3}
\end{table}
\subsection{Editing methods}
We use ROME (Rank-One Model Editing)~\cite{meng2022locating} and MEMIT (Mass Editing Memory in a Transformer)~\cite{meng2023massediting} which are the state-of-the-art editing schemes and particularly suitable for multilingual settings.\\
\noindent \textbf{Rank-One Model Editing (ROME)}: %Meng and colleagues (2022) introduced an advanced model editing method that leverages neuron interpretability to surpass other techniques in specificity and generalization. 
This method specifically alters the weights in the initial feed-forward layers of a pretrained model. It identifies factual associations through causal interventions, enabling precise and effective modifications.\\
\noindent \textbf{Mass Editing Memory in a Transformer (MEMIT)}: MEMIT advances ROME, by extending its capabilities. While ROME applied a rank-one modification to the MLP weights of a single layer to embed a memory directly into the model, MEMIT enhances this approach by adjusting the MLP weights across multiple critical layers to incorporate numerous memories.

\subsection{Evaluation metric}
We evaluate the edited models using two metrics: \\
\noindent\textit{\textbf{Exact match}}: Here accuracy is determined by checking if the ground truth is present in the model's output. Outputs containing the exact expected response are classified as correct, while others are deemed incorrect, providing a binary measure of performance.\\
\noindent\textit{\textbf{Partial match}}: The Levenshtein ratio~\cite{Levenshtein1965BinaryCC} measures textual similarity, calculated as the Levenshtein distance divided by the maximum text length. Outputs surpassing an 80\% ratio but not containing the ground truth as a substring are considered accurate, allowing for minor acceptable deviations.

\section{Results}

\begin{table*}[!t]
\centering
\resizebox{1.0\textwidth}{!}{
\begin{tabular}{cccccccccc|cccccccc}
\hline
\multicolumn{2}{c|}{\textbf{Dataset}} &
  \multicolumn{8}{c|}{\ctr{}} &
  \multicolumn{8}{c}{\zsre{}} \\ \hline
\multicolumn{2}{c}{\textbf{Inferencing language}} &
  \multicolumn{2}{c}{\textbf{En}} &
  \multicolumn{2}{c}{\textbf{Hi}} &
  \multicolumn{2}{c}{\textbf{Ta}} &
  \multicolumn{2}{c|}{\textbf{Kn}} &
  \multicolumn{2}{c}{\textbf{En}} &
  \multicolumn{2}{c}{\textbf{Hi}} &
  \multicolumn{2}{c}{\textbf{Ta}} &
  \multicolumn{2}{c}{\textbf{Kn}} \\ \hline
\textbf{Editing language} &
  \textbf{Properties} &
  \textbf{ROME} &
  \textbf{MEMIT} &
  \textbf{ROME} &
  \textbf{MEMIT} &
  \textbf{ROME} &
  \textbf{MEMIT} &
  \textbf{ROME} &
  \textbf{MEMIT} &
  \textbf{ROME} &
  \textbf{MEMIT} &
  \textbf{ROME} &
  \textbf{MEMIT} &
  \textbf{ROME} &
  \textbf{MEMIT} &
  \textbf{ROME} &
  \textbf{MEMIT} \\ \hline
\multirow{4}{*}{\textbf{En}} &
  Rel &
  \underline{0.73}/0.75 &
  \textbf{0.95}/0.95 &
  0.00/0.00 &
  0.01/0.01 &
  0.00/0.00 &
  0.01/0.01 &
  0.00/0.01 &
  0.00/0.01 &
  \underline{0.29}/0.33 &
  \textbf{0.59}/0.59 &
  0.01/0.02 &
  0.02/0.02 &
  0.00/0.00 &
  0.00/0.00 &
  0.00/0.02 &
  0.00/0.00 \\
 &
  Gen &
  \underline{0.35}/0.35 &
  \textbf{0.64}/0.64 &
  0.01/0.01 &
  0.02/0.02 &
  0.01/0.01 &
  0.01/0.02 &
  0.00/0.01 &
  0.00/0.01 &
  \underline{0.29}/0.31 &
  \textbf{0.52}/0.54 &
  0.01/0.02 &
  0.00/0.00 &
  0.01/0.01 &
  0.00/0.00 &
  0.00/0.03 &
  0.00/0.00 \\
 &
  Loc &
  0.33/0.33 &
  0.27/0.27 &
  0.01/0.01 &
  0.01/0.01 &
  0.02/0.02 &
  0.03/0.03 &
  0.11/0.11 &
  0.12/0.12 &
  0.00/0.00 &
  0.00/0.00 &
  0.00/0.00 &
  0.00/0.00 &
  \underline{0.01}/0.01 &
  0.00/0.04 &
  \underline{0.01}/0.02 &
  \textbf{0.02}/0.04 \\
 &
  Port &
  \underline{0.00}/0.00 &
  \underline{0.00}/0.01 &
  \underline{0.00}/0.01 &
  \underline{0.00}/0.01 &
  \underline{0.00}/0.00 &
  \underline{0.00}/0.00 &
  \underline{0.00}/0.00 &
  \underline{0.00}/0.00 &
  \underline{0.03}/0.04 &
  0.02/0.04 &
  0.00/0.01 &
  0.00/0.00 &
  0.00/0.01 &
  0.00/0.00 &
  0.00/0.01 &
  0.00/0.00 \\ \hline
\multirow{4}{*}{\textbf{Hi}} &
  Rel &
  0.00/0.01 &
  0.01/0.01 &
  0.01/0.03 &
  0.07/0.09 &
  0.00/0.00 &
  0.01/0.01 &
  0.00/0.01 &
  0.00/0.01 &
  0.00/0.00 &
  0.00/0.00 &
  0.01/0.03 &
  0.05/0.05 &
  0.00/0.00 &
  0.00/0.00 &
  0.00/0.02 &
  0.00/0.01 \\
 &
  Gen &
  0.00/0.00 &
  0.01/0.01 &
  0.02/0.03 &
  0.03/0.04 &
  0.00/0.00 &
  0.01/0.01 &
  0.00/0.01 &
  0.00/0.01 &
  0.00/0.00 &
  0.01/0.01 &
  0.01/0.03 &
  0.02/0.03 &
  0.01/0.02 &
  0.01/0.02 &
  0.00/0.03 &
  0.00/0.02 \\
 &
  Loc &
  \underline{0.35}/0.35 &
  \underline{0.35}/0.36 &
  0.01/0.01 &
  0.01/0.01 &
  0.03/0.03 &
  0.03/0.03 &
  0.12/0.12 &
  0.13/0.13 &
  0.00/0.00 &
  0.00/0.00 &
  0.00/0.00 &
  0.00/0.00 &
  \underline{0.01}/0.01 &
  \underline{0.01}/0.01 &
  \underline{0.01}/0.01 &
  \underline{0.01}/0.01 \\
 &
  Port &
  \underline{0.00}/0.00 &
  \underline{0.00}/0.00 &
  \textbf{0.01}/0.01 &
  \underline{0.00}/0.00 &
  \underline{0.00}/0.00 &
  \underline{0.00}/0.00 &
  \underline{0.00}/0.00 &
  \underline{0.00}/0.00 &
  \textbf{0.07}/0.08 &
  0.00/0.00 &
  0.00/0.01 &
  0.00/0.01 &
  0.00/0.01 &
  0.00/0.01 &
  0.00/0.01 &
  0.00/0.01 \\ \hline
\multirow{4}{*}{\textbf{Ta}} &
  Rel &
  0.00/0.01 &
  0.00/0.01 &
  0.00/0.00 &
  0.00/0.00 &
  0.00/0.01 &
  0.01/0.01 &
  0.00/0.01 &
  0.00/0.01 &
  0.00/0.00 &
  0.01/0.01 &
  0.00/0.00 &
  0.01/0.01 &
  0.00/0.02 &
  0.01/0.03 &
  0.00/0.01 &
  0.00/0.01 \\
 &
  Gen &
  0.00/0.00 &
  0.00/0.00 &
  0.00/0.00 &
  0.00/0.00 &
  0.01/0.01 &
  0.00/0.00 &
  0.00/0.01 &
  0.00/0.01 &
  0.00/0.00 &
  0.01/0.01 &
  0.00/0.00 &
  0.01/0.01 &
  0.01/0.01 &
  0.02/0.03 &
  0.00/0.02 &
  0.00/0.02 \\
 &
  Loc &
  \textbf{0.36}/0.36 &
  0.33/0.34 &
  0.01/0.01 &
  0.02/0.02 &
  0.02/0.02 &
  0.02/0.02 &
  0.11/0.11 &
  0.11/0.11 &
  0.00/0.00 &
  0.00/0.00 &
  0.00/0.00 &
  0.00/0.00 &
  \underline{0.01}/0.03 &
  \underline{0.01}/0.02 &
  \underline{0.01}/0.02 &
  \underline{0.01}/0.02 \\
 &
  Port &
  \underline{0.00}/0.00 &
  \underline{0.00}/0.00 &
  \underline{0.00}/0.01 &
  \underline{0.00}/0.01 &
  \underline{0.00}/0.00 &
  \underline{0.00}/0.00 &
  \underline{0.00}/0.00 &
  \underline{0.00}/0.00 &
  0.00/0.00 &
  0.00/0.00 &
  0.00/0.00 &
  0.00/0.00 &
  \textbf{0.01}/0.01 &
  0.00/0.01 &
  0.00/0.01 &
  0.00/0.01 \\ \hline
\multirow{4}{*}{\textbf{Kn}} &
  Rel &
  0.00/0.01 &
  0.00/0.01 &
  0.00/0.00 &
  0.00/0.00 &
  0.00/0.00 &
  0.00/0.00 &
  0.00/0.01 &
  0.00/0.01 &
  0.00/0.00 &
  0.00/0.00 &
  0.00/0.01 &
  0.00/0.00 &
  0.00/0.02 &
  0.00/0.00 &
  0.03/0.03 &
  0.00/0.03 \\
 &
  Gen &
  0.00/0.00 &
  0.00/0.00 &
  0.00/0.00 &
  0.00/0.00 &
  0.00/0.01 &
  0.00/0.00 &
  0.00/0.01 &
  0.00/0.01 &
  0.00/0.00 &
  0.00/0.00 &
  0.00/0.01 &
  0.00/0.00 &
  0.01/0.03 &
  0.01/0.02 &
  0.01/0.03 &
  0.00/0.04 \\
 &
  Loc &
  \underline{0.35}/0.35 &
  0.34/0.34 &
  0.01/0.01 &
  0.02/0.02 &
  0.03/0.03 &
  0.03/0.03 &
  0.12/0.12 &
  0.12/0.12 &
  0.00/0.00 &
  0.00/0.00 &
  \underline{0.01}/0.01 &
  0.00/0.00 &
  0.00/0.01 &
  0.00/0.00 &
  \underline{0.01}/0.01 &
  0.00/0.00 \\
 &
  Port &
  \underline{0.00}/0.00 &
  \underline{0.00}/0.00 &
  \underline{0.00}/0.01 &
  \underline{0.00}/0.01 &
  \underline{0.00}/0.00 &
  \underline{0.00}/0.00 &
  \underline{0.00}/0.00 &
  \underline{0.00}/0.00 &
  0.00/0.00 &
  0.00/0.00 &
  0.00/0.01 &
  0.00/0.00 &
  0.00/0.01 &
  0.00/0.00 &
  0.00/0.01 &
  0.00/0.01 \\ \hline
\end{tabular}
}
\caption{\footnotesize Comparison of scores across the merged model for three Indic languages, evaluated using the \ctr{} and \zsre{} datasets for each language and others. Highest scores are in bold, and second-highest are underlined. Values represent Exact Match/Partial Match results.}
\label{tab:merged}
%\vspace{-0.4cm}
\end{table*}
\subsection{\textit{Self edit - self inference} perspective}
In this setup we perform the edit in a particular language (say German) and obtain the generated output from the model in the same language (i.e., German itself).\\
\noindent \ctr{} \textbf{dataset}: In our evaluations of the model performance for the \ctr{} dataset, we observe marked variations across different languages and metrics in Table~\ref{tab:my-table1}, illustrating significant challenges in multilingual adaptability and contextual understanding. For instance, German language tests show that models like \tower{} and \mistral{} achieve good reliability scores (ROME at 0.83 and MEMIT at 0.73 for \tower{}; the same scores are at 0.83 and 0.73 respectively for \mistral{}). These scores illustrate good model performance in understanding the contextual nuances of German. However, generalization and locality score are less impressive (\tower{} at 0.27 and 0.22 on ROME for generalization and locality respectively), indicating difficulties in applying the learned information across broader contexts and different locales within the German language. Similar patterns are observed in Spanish and Italian. In Spanish, \tower{} reaches a reliability score of 0.82 for ROME and 0.70 for MEMIT; for \mistral{} the reliability scores are 0.81 for ROME and 0.78 for MEMIT, suggesting decent grasp of Spanish contexts. However, the generalization scores remain below 0.35 for ROME and locality scores do not exceed 0.29 for MEMIT for any model. Despite \tower{} showing a relatively high reliability in Italian with a ROME at 0.87 and MEMIT at 0.74, the generalization and locality scores remain low (highest being 0.35 on ROME and 0.28 on MEMIT for \mistral{}). In case of the three Indic languages the discrepancies become even more pronounced (See Table~\ref{tab:my-table3}). \hathi{}, for example, shows a drastic drop in Hindi, with a ROME reliability of just 0.02 and a MEMIT of 0.45, indicating almost no comprehension of the language nuances. \tamil{} and \kan{} also display low scores across all properties. The highest reliability achieved is 0.21 for ROME for \kan{} and 0.48 for MEMIT in case of \tamil{}, which highlights the limitations in these language models. Portability scores are consistently low across all languages, models, and metrics, demonstrating a significant gap in model training as it fails to  effectively account for diverse linguistic structures and cultural contexts.

\begin{table*}[h]
\centering
\resizebox{1.0\textwidth}{!}{
\begin{tabular}{l|l|l}
\hline
\multicolumn{1}{c|}{\textbf{Category}} & \multicolumn{1}{c|}{\textbf{Examples}}                                                                                                                                                                                                                                                                                                                                                                                                                                                                                                                                                                                                                                                                                                                                                                                                                                          & \multicolumn{1}{c}{\textbf{Possible solution}}         \\ \hline
\textbf{Lexical ambiguity}             & \begin{tabular}[c]{@{}l@{}}English: `Fair' can mean a carnival, treating someone right, or having light skin and/or hair\\ French: `Livre' can refer to a book or to the weight measure pound.\\ \if{0}German: “Bank” might mean a financial institution or a bench.\\ Italian: `Lingua' means both language and tongue.\\ Spanish: “Doble” can be a double or a fold.\\ \fi\end{tabular}                                                                                                                                                                                                                                                                                                                                                                                                                                                                                       & Context-aware models                                   \\ \hline
\textbf{Syntactic ambiguity}           & \begin{tabular}[c]{@{}l@{}}English: ``Visiting relatives can be boring.'' (Ambiguous: Visiting them, or the relatives who visit, can be boring.)\\ \if{o}French: "Ils pêchent des poissons." (They fish for fish. Ambiguous due to lack of context about who or what is being fished.)\\ \fi German: ``Er sah den Mann mit dem Fernglas.'' (He saw the man with the binoculars. Ambiguous: Who has the binoculars?)\\ Italian: ``Ho visto l'uomo con il binocolo.'' (I saw the man with the binocular. Ambiguous similar to German.)\\ \if{0}Spanish: "Juan vio al hombre con el telescopio." (Juan saw the man with the telescope. Ambiguous: Who has the telescope?) \fi\end{tabular}                                                                                                                                                                                         & Better parsing                                         \\ \hline
\textbf{Semantic ambiguity}            & \begin{tabular}[c]{@{}l@{}}\if{0}English: ``In which city famous for its role in the industrial revolution did May Wright Sewall pass away?''\\ (Ambiguous: industrial revolution of which country or is it the first industrial revolution?)\\ \fi French: ``Mexx, ça a commencé en'' (Mexx, that was started in. Ambiguous: started means founded or\\ started in a particular region)\\ Spanish: ``Spike Hughes se origina de'' (Spike Hughes originates from. Ambiguous: originates from a place or\\ from a particular family)\\ \if{0}Italian: "In quale città è avvenuto l'attentato suicida al ristorante Sbarro secondo la risposta modificata?"\\ (In which city did the suicide bombing of the Sbarro restaurant take place according to the altered\\ answer?; Ambiguous: no relation of altered answer i.e. Manhattan with Sbarro suicide bombing)\fi\end{tabular} & Incorporation of additional semantic cues              \\ \hline
\textbf{Cultural ambiguity}            & \begin{tabular}[c]{@{}l@{}}English: ``Arrow of Time/The Cycle of Time'' (Is an album of Peter Michael Hamel. But it could also mean the flow of time)\\ French: ``Ce n'est pas ma tasse de thé.'' (It's not my cup of tea. Ambiguous without understanding the idiom.)\\ \if{0}German: "Das ist nicht mein Bier." (That's not my beer. Similar to the French, but with a cultural twist.)\\ \fi Italian: ``In bocca al lupo.'' (In the wolf's mouth, means good luck. Could be confusing without cultural context.)\\ \if{0}Spanish: "Estar en las nubes." (To be in the clouds, meaning to be daydreaming. Might be taken literally.)\fi\end{tabular}                                                                                                                                                                                                                          & Deeper multi-cultural context                          \\ \hline
\textbf{Translation errors}            & \begin{tabular}[c]{@{}l@{}}English: ``In which country's capital city would you most likely\\ hear Faithless' original language spoken?'' translated into French and back to English becomes ``In which\\ country's capital would you most likely hear the original language of the original spoken''\end{tabular}                                                                                                                                                                                                                                                                                                                                                                                                                                                                                                                                                              & Reinterpretation of the translation in target language \\ \hline
\textbf{NER errors}                    & \begin{tabular}[c]{@{}l@{}}English: ``The Little Match Girl'' could be a literary fairy tale.\\ Spanish: `Rio' can mean a river or refer to the city Rio de Janeiro.\end{tabular}                                                                                                                                                                                                                                                                                                                                                                                                                                                                                                                                                                                                                                                                                               & Integration of knowledge graphs                        \\ \hline
\textbf{Idioms}                        & \begin{tabular}[c]{@{}l@{}}German: ``Der Blick von unten'' (Literally: Seeing things from a low physical position. Meaning: Considering\\ a situation from a marginalized or disadvantaged perspective.)\end{tabular}                                                                                                                                                                                                                                                                                                                                                                                                                                                                                                                                                                                                                                                           & Maintain exception lists                               \\ \hline
\textbf{Phonetic/orthographic errors}  & \begin{tabular}[c]{@{}l@{}}English: `Their' vs. `There' vs. `They’re'\\ Spanish: `Vino' (came) vs. `Vino' (wine)\end{tabular}                                                                                                                                                                                                                                                                                                                                                                                                                                                                                                                                                                                                                                                                                                                                                   & Context-sensitive correction of word forms             \\ \hline
\textbf{Morphological errors}          & \begin{tabular}[c]{@{}l@{}}German: The misuse of gender-specific articles "der" (masculine), "die" (feminine), "das" (neuter) can lead to confusion\\ Italian: Confusion between "mangiato" (eaten) and "mangiando" (eating) can change the temporal context of a sentence.\end{tabular}                                                                                                                                                                                                                                                                                                                                                                                                                                                                                                                                                                                        & Integration of specialised morphological rules         \\ \hline
\textbf{Pragmatic errors}              & French: Using `tu' (informal you) instead of `vous' (formal or plural you) in a formal context can be seen as rude or too casual.                                                                                                                                                                                                                                                                                                                                                                                                                                                                                                                                                                                                                                                                                                                                               & Understanding cultural norms                           \\ \hline
\end{tabular}
}
\caption{Categorization of multilingual knowledge editing errors, including lexical, syntactic, semantic, cultural, and contextual ambiguities, with examples from English, French, German, Italian, and Spanish, highlighting challenges in cross-lingual consistency and accuracy.}
%\vspace{-0.5cm}
\label{tab:error_analysis}
\end{table*}
\noindent\zsre{} \textbf{dataset}: In case of \zsre{} dataset (see Table~\ref{tab:my-table1}) German shows moderate performance in reliability with scores like 0.48 on ROME and 0.25 on MEMIT for \tower{}. The generalization (0.33 for ROME) and locality scores ($\sim 0$) are also very poor. These results indicate substantial deficiencies in capturing language-specific details and generalizing learned information. Spanish fares slightly better in reliability, achieving up to 0.49 on ROME with \tower{} and \mistral{}, but like German, faces challenges in generalization and locality, with the best generality reaching only 0.35 and locality remaining near zero. Italian (It) generally scores higher in reliability, particularly with \mistral{} reaching 0.58 on ROME, though it too struggles with generality and locality. French exhibits a similar trend, with reliability scores reaching up to 0.52 for ROME with \mistral{} and both generalization and locality scores remaining low. Performance markedly drops for the three Indic languages (See Table~\ref{tab:my-table3}). For instance, Hindi's highest reliability is just 0.03 for ROME, while Tamil and Kannada only achieve maximum reliability scores of 0.06 and 0.16  respectively for ROME. Across all languages, portability scores are low, reflecting limited adaptability and the challenge of transferring learned capabilities from one linguistic context to another.%, with scores like 0.02 for German and 0.03 for Spanish in TowerInstruct, suggesting that even languages like Italian and French that exhibit slightly better portability still face significant hurdles in generalizing across languages.

\subsection{\textit{English edit - self inference} perspective}
In this setup we perform the edit in a English and obtain the generated output from the model in other languages (e.g., German, Italian etc.).

\noindent\ctr{} \textbf{dataset}: In German, the reliability scores for models such as \tower{} and \mistral{} suggest moderate effectiveness, with ROME around 0.48 and MEMIT around 0.40 (see Table~\ref{tab:my-table2}). However, their generalization and locality scores reveal limitations in the models' ability to generalize and localize content effectively with scores not exceeding 0.25 and 0.26 respectively. For Spanish, there is a noticeable improvement in reliability, with ROME scores for \mistral{} reaching 0.57, and a slight improvement in generalization and locality metrics compared to German. Italian and French show similar trends, with reliability scores peaking at 0.47 for \mistral{} in Italian and 0.49 in French; the generalization and locality scores are still lower. For Tamil and Kannada the reliability are exceptionally low (See Table~\ref{tab:my-table3}). In fact, in case of Tamil this score is 0 for ROME and 0.01 for MEMIT. Comparatively for Hindi the reliability scores are quite good with 0.56 for ROME. However the portability and generalization scores are again very poor. \\ 
\begin{theo}[]
\footnotesize
{\color{red}\small\faHandPointRight} Models like \tower{} and \mistral{} excel in context-specific reliability but falter in generalization and locality.\\
{\color{red}\small\faHandPointRight} Indic languages exhibit larger gaps, reflecting limited linguistic diversity in training.\\ 
{\color{red}\small\faHandPointRight} Cross-lingual edits expose critical weaknesses, with performance dropping across linguistic boundaries, and model merging fails to enhance reliability, locality, or generalization on either dataset.
\end{theo}
\noindent\zsre{} \textbf{dataset}: For languages such as German and Spanish, the models display moderate reliability with \mistral{}, achieving ROME scores up to 0.34 and 0.39 respectively, and MEMIT scores of 0.14 and 0.19 respectively (see Table~\ref{tab:my-table2}). However, the scores significantly drop for locality and portability, showing that while the models can identify relevant relationships, they struggle to generalize and adapt to the specific linguistic nuances of these languages. The trends are similar in Italian and French, where reliability scores are moderate while locality and generalization scores are poor. %for example, Mistral’s performance in Italian indicates a ROME relevance up to 0.34, yet both locality and generality scores are considerably lower, underscoring difficulties in adapting the models to effectively process and understand language-specific nuances. 
Further, for the Indic languages, the score are exceedingly low for all the properties indicating the stark gap in performance highly resource scarce languages.%; for instance, the OpenHathi model in Hindi shows minimal relevance and very low generality and locality, while Tamil and Kannada, evaluated through Tamil-Llama and Kan-Llama respectively, demonstrate virtually no effective model performance, with scores nearing zero in all assessed metrics.
%\vspace{-0.2cm}

\subsection{\textit{Merged} model perspective}
% The Table~\ref{tab:merged} presents performance metrics for the merged model, showcasing how it handles edits across various languages, with columns representing the inferencing language and rows indicating the respective editing language. When both editing and inferencing are done in English the reliability scores for the \ctr{} dataset are quite high with ROME and MEMIT respectively reaching 0.73 and 0.95. 
% However, the performance sharply declines when editing in English and inferencing is in Hindi, Tamil, and Kannada with scores nearly zero, highlighting a stark limitation in the model's cross-lingual capabilities. This trend is consistently observed across both datasets. Editing in languages like Hindi, Tamil, or Kannada results in uniformly poor outcomes across all four properties regardless of the inferencing language, which indicates profound inadequacies in the model’s ability to generalize and adapt across linguistic barriers. This pattern emphasizes the critical need for advancing multilingual model adaptability. The current findings suggest that while the model operates effectively within the confines of the same linguistic environment, its performance deteriorates dramatically across linguistic boundaries, especially from lesser-resourced languages. Such insights advocate for a significant enhancement in training approaches, aiming to foster robust multilingual support and ensure that models are truly multilingual in functionality, proficiently managing edits and inferencing across a diverse linguistic landscape.
Table~\ref{tab:merged} presents performance metrics for the merged model, with columns representing inferencing languages and rows indicating editing languages. Editing and inferencing in English yield high reliability scores on the \ctr{} dataset (ROME: 0.73, MEMIT: 0.95). However, performance drops to near zero when editing in English and inferencing in Hindi, Tamil, or Kannada, exposing the model's cross-lingual limitations. Editing in Hindi, Tamil, or Kannada consistently results in poor outcomes across all properties, regardless of the inferencing language. This highlights the model's inability to generalize across linguistic barriers and underscores the need for improved multilingual adaptability. The findings reveal that while the model performs well within the same linguistic environment, its performance deteriorates significantly across lesser-resourced languages, necessitating enhanced training approaches for robust multilingual support.
%\vspace{-0.3cm}
\section{Error analysis}
%To enhance our comprehension of the intricacies involved in quality translation and post-editing processes, we have meticulously analyzed a selection of handcrafted samples. Our examination reveals that inaccuracies may stem from various stages—either within the translation itself, the editing phase, or a combination of both. 
In Table~\ref{tab:error_analysis} we show the different types of linguistic errors encountered during the translation and editing process. The errors are categorised based on the different types of ambiguities and sheds light on how future models should strengthened by carefully harnessing techniques to tackle these errors. More details are available in Appendix~\ref{sec:erroranalysis}.

\section{Discussion}
Here we discuss two important questions -- \textit{How do multilingual LLMs handle cross-lingual knowledge edits?} and \textit{What steps can industry practitioners take to address cross-lingual disparities?} %We examine the core issues in cross-lingual knowledge editing and propose practical solutions for real-world deployments.

\begin{mdframed}[roundcorner=6pt,backgroundcolor=gray!10]
\subsection*{\textit{How do multilingual LLMs handle cross-lingual knowledge edits?}}
Modern \emph{LLMs} often fail to propagate factual updates consistently across languages. While languages like English, French, and German benefit from extensive corpora \cite{Xu2024ASO}, those like Hindi, Tamil, and Kannada suffer from data scarcity, causing unstable knowledge transfer \cite{qi-etal-2023-cross}. Further, editing methods \textbf{ROME} and \textbf{MEMIT} encounter problems with highly agglutinative or morphologically rich languages.
\paragraph{Key observations}
\begin{compactitem}
\item \textbf{Data scarcity}: Inadequate corpora produce sparse embeddings, disrupting the model's ability to adapt newly introduced facts \cite{10.1145/3511095.3531277}.
\item \textbf{Architectural bias}: LLM pipelines typically prioritize English, overlooking morphological idiosyncrasies in languages like Tamil or Kannada.
\item \textbf{Complex linguistic features}: Idiomatic expressions and cultural references can invalidate edits that were accurate in English \cite{beniwal-etal-2024-cross}; merging specialized models can exacerbate divergences if representations are misaligned \cite{yadav2023tiesmerging}.
\end{compactitem}
\end{mdframed}
\begin{mdframed}[roundcorner=6pt,backgroundcolor=gray!5]
\subsection*{\textit{What steps can industry practitioners take to address cross-lingual disparities?}}
A holistic approach is needed to ensure consistent, multi-lingual fact-editing. Below are five key strategies:

\begin{compactitem}
\item \textbf{Expand low-resource corpora}:\\ \emph{\underline{Rationale}}: Larger, more representative datasets address embedding sparsity; \\ \emph{\underline{Implementation}}: Generate crowdsourced/synthetic data \cite{hazra-etal-2024-sowing}.
\item \textbf{Continuous model editing}:\\ \emph{\underline{Rationale}}: Iterative edits balance new knowledge with existing facts\footnote{https://www.microsoft.com/en-us/research/blog/lifelong-model-editing-in-large-language-models-balancing-low-cost-targeted-edits-and-catastrophic-forgetting/}; primarily important for industries dealing with finance, healthcare, and law (e.g., updating a multilingual LLM to reflect new data privacy laws (GDPR, CPRA) in different regions without retraining from scratch).\\ \emph{\underline{Case study}}: Microsoft’s lifelong editing merges local patches with broader retraining \cite{decao2021editing}.
\item \textbf{Alignment-focused architectures}:\\ \emph{\underline{Rationale}}: Combine morphological analysis, advanced NER, \& cross-lingual parameter sharing;\\ \emph{\underline{Benefit}}: Stable knowledge propagation in structurally diverse languages \cite{wang2023crosslingual}.
\item \textbf{Dedicated edit modules}:\\ \emph{\underline{Rationale}}: Log each update \& validate in all languages to avoid accidental overwrites;\\ \emph{\underline{Implementation}}: Use an ``edit ledger'' in attention layers \cite{hase2023does}.
\item \textbf{Rigorous multilingual testing:}\\ \emph{\underline{Rationale:}} Systematic checks prevent bias \& misinformation from creeping in;\\ \emph{\underline{Tools}}: Curated test suites for reliability, cultural fitness, and domain-specific accuracy \cite{hazra-etal-2024-sowing}.
\end{compactitem}
\end{mdframed}

\section{Conclusion}

In this study, we investigated the impact of knowledge editing across different languages based on the \ctr{} and \zsre{} datasets along with their translations. Our extensive experiments employing a variety of knowledge editing techniques on an array of multilingual LLMs resulted in various crucial observations. We discovered that variations in language-specific model architecture significantly affect the success of knowledge edits, that current editing methods often fail to seamlessly transfer alterations from one language to another, and that modifications made in one language might unexpectedly alter model behavior in another language. %These results highlight the intricate challenges and inherent limitations of current methodologies in managing and deploying knowledge across diverse linguistic environments within LLMs. Moreover, our findings underscore the need for ongoing research aimed at improving the robustness and reliability of LLMs in multilingual contexts, which is essential for creating more equitable and universally applicable AI systems in the increasingly interconnected world. 
This study lays the groundwork for future innovations that could lead to more sophisticated and linguistically inclusive AI technologies.

\section{Limitations}
Despite the promising results, our study has several limitations. The variability in performance across different languages highlights the inherent challenges in achieving true multilingual consistency, with models exhibiting substantial difficulties in generalizing and localizing edits, particularly in low-resourced languages such as Hindi, Tamil, and Kannada. This discrepancy indicates a need for more inclusive and representative training datasets that encompass a wider range of linguistic and cultural contexts. Additionally, our focus on decoder-only models limits the generalizability of our findings to other types of language models, such as encoder-decoder architectures. The relatively low portability scores across all languages further indicate that current models struggle to transfer learned knowledge effectively from one linguistic context to another, especially in cross-lingual edits where modifications in one language often fail to translate accurately into another. Moreover, the merging of models, while showing some promise, does not consistently improve reliability, locality, or generalization metrics, suggesting that further research is needed to optimize these approaches.

\section{Ethical consideration}

Our research raises ethical concerns regarding linguistic equity and cultural sensitivity. Disparities in model performance could reinforce existing linguistic inequities, limiting access to AI technologies for speakers of low-resourced languages. Future model development must include diverse languages and dialects to promote equity. Additionally, errors related to cultural ambiguity and idiomatic expressions can lead to misinterpretations or offensive content, necessitating robust evaluation frameworks to ensure cultural sensitivity. Privacy and security risks are also significant, as models may inadvertently reveal sensitive information during knowledge editing processes. Researchers must prioritize user privacy and implement stringent data protection measures to prevent misuse of personal data, ensuring AI technologies are effective and equitable for all users.

\section{Potential risk}
LLMs can be used for harmful content generation and misinformation spread. The prompts used and generated in this work can be misused to generate harmful content.

\bibliography{custom}

\appendix

\section{Model selection}
\label{sec:modselection}
\noindent \textbf{Mistral-7B-Instruct-v0.2}\footnote{https://huggingface.co/mistralai/Mistral-7B-Instruct-v0.2}: The model was developed by~\cite{jiang2023mistral} and supports multilinguality\footnote{https://encord.com/blog/mistral-large-explained/}. It is designed around the causal language modeling framework. We shall refer to this model as \mistral{}.\\ 
\noindent \textbf{TowerInstruct-7B-v0.2}\footnote{https://huggingface.co/Unbabel/TowerInstruct-7B-v0.2}: This model~\cite{tower_llm_2024} has been developed on top of LLaMA2~\cite{touvron2023llama} architecture and supports multilinguality including English, German, French, Spanish, Chinese, Portuguese, Italian, Russian, Korean, and Dutch. We shall refer to this model as \tower{}.\\
\noindent \textbf{OpenHathi-7B-Hi-v0.1-Base}\footnote{https://huggingface.co/sarvamai/OpenHathi-7B-Hi-v0.1-Base}: The model is designed to optimize multilingual interactions with a special focus on Indian languages. It uses a transformer-based architecture similar to GPT-3 but introduces hybrid partitioned attention to efficiently manage computational resources and enhance responsiveness across languages like Hindi, Tamil, and Bengali. We shall refer to this model as \hathi{}.\\
\noindent \textbf{Tamil-llama-7b-base-v0.1}\footnote{https://huggingface.co/abhinand/tamil-llama-7b-base-v0.1}: This is a sophisticated model~\cite{balachandran2023tamilllamanewtamillanguage} developed specifically for bilingual tasks in Tamil and English, leveraging a 7 billion parameter causal language modeling  framework. We shall refer to this model as \tamil{}.\\
\noindent \textbf{Kan-LLaMA-7B-SFT}\footnote{https://huggingface.co/Tensoic/Kan-Llama-7B-SFT-v0.5}: This model is tailored for efficient Kannada text processing with an expanded 49,420-token vocabulary, enhancing its language handling capabilities. Pre-trained on 600 million Kannada tokens from the CulturaX dataset, it employs a low-rank adaptation technique to minimize computational costs while preserving the model’s integrity. We shall refer to this model as \kan{}.

\section{Error analysis}
\label{sec:erroranalysis}
\noindent \textbf{Lexical ambiguity} Lexical ambiguity occurs when a word has multiple meanings, leading to confusion without context. For instance, the English word "crane" can refer to a bird or construction equipment, a distinction crucial for accurate knowledge representation. %Multilingual systems must utilize contextual clues to disambiguate such terms, a task that becomes increasingly complex across languages with diverse lexical scopes and cultural nuances.

\noindent \textbf{Syntactic ambiguity} Syntactic ambiguity arises from sentence structures that can be interpreted in multiple ways. An example is the English sentence "Visiting relatives can be boring," which could imply either the act of visiting relatives is boring or that the relatives being visited are boring. Resolving these ambiguities requires advanced parsing techniques and an understanding of the specific language's syntax to ensure accurate interpretation.

\noindent \textbf{Semantic ambiguity errors} Semantic ambiguity pertains to the uncertainty of meaning within a sentence or phrase. For example, "He gave her a ring" could mean a telephone call or presenting a piece of jewelry. Multilingual systems need to discern the intended meaning based on semantic cues and the broader context, a challenging task given the subtlety of cues and cultural specificities in language use.

\noindent \textbf{Cultural and contextual errors} These errors occur when language processing fails to account for cultural idioms or context-specific meanings. Phrases like "Piece of cake" in English, meaning something easy, can be misunderstood if taken literally or translated directly into another language without considering idiomatic expressions. Handling these requires deep cultural knowledge and contextual understanding beyond linguistic analysis.

\noindent \textbf{Translation errors} Translation errors emerge when converting text from one language to another, often leading to loss of meaning or inaccuracies. These can be particularly problematic in knowledge editing, where precision is paramount. For example, translating idiomatic expressions or culturally specific terms often requires not just a direct translation but a reinterpretation in the target language.

\noindent \textbf{Named entity recognition (NER) errors} NER errors involve the incorrect identification or classification of proper nouns in text. For instance, distinguishing between "Rio" as a river or the city of Rio de Janeiro in Spanish requires contextual analysis. Accurate NER is essential for knowledge databases to correctly link information to entities, demanding sophisticated language models that can navigate these nuances.

\noindent \textbf{Idiomatic expression errors} Errors in understanding or translating idiomatic expressions can significantly alter the intended meaning. For example, the Italian idiom "Tra il dire e il fare c'è di mezzo il mare" illustrates the difference between saying and doing, a concept that might be lost if translated literally. Addressing these requires an in-depth understanding of both the source and target languages' idioms.

\noindent \textbf{Phonetic and orthographic errors} These errors occur with words that sound similar (homophones) or are spelt similarly (homographs) but have different meanings. For instance, "their," "there," and "they’re" in English. Multilingual systems must accurately identify and apply the correct form based on context, a challenging task that often requires human-like understanding of language.

\noindent \textbf{Morphological errors} Morphological errors refer to the misuse of word forms, affecting the grammatical structure and potentially changing the meaning of sentences. German's gender-specific articles—der, die, das—offer a prime example, where incorrect usage can confuse readers and misrepresent information. Overcoming these demands a robust grasp of linguistic rules and the flexibility to apply them in diverse contexts.

\noindent \textbf{Pragmatic errors} Pragmatic errors involve the misuse or misunderstanding of language in social context, such as politeness or formality levels. An example is the inappropriate use of "tu" (informal) and "vous" (formal or plural) in French, which can significantly affect the tone and perceived respectfulness of an interaction. Addressing these requires sensitivity to cultural norms and the social dynamics of language, highlighting the complexity of human communication and the challenges in replicating these nuances in AI systems.

\section{Hyperparameters}

We adopt all essential parameter values from the ROME and MEMIT study for all the LLMs. The details of these hyperparameters are provided in Table~\ref{tab:hyper}. %Due to technical limitations, we were unable to conduct the MEMIT experiments on Gemma using the default settings.

% \begin{table}[h]
% \centering
% %\small
% \resizebox{.38\textwidth}{!}{
% \begin{tabular}{|l|}
% \hline
% \multicolumn{1}{|c|}{\textbf{Hyperparameter Values}}                                                                                                                                                                                                                                                                                                                                                                                                                                                                                                                                                                \\ \hline
% \begin{tabular}[c]{@{}l@{}}layers: {[}5{]}\\ fact\_token: "subject\_last"\\ v\_num\_grad\_steps: 25\\ v\_lr: 5e-1\\ v\_loss\_layer: 31\\ v\_weight\_decay: 1e-3\\ clamp\_norm\_factor: 4\\ kl\_factor: 0.0625\\ mom2\_adjustment: false\\ context\_template\_length\_params: {[}{[}5, 10{]}, {[}10, 10{]}{]}\\ rewrite\_module\_tmp: "model.layers.\{\}.mlp.down\_proj"\\ layer\_module\_tmp: "model.layers.\{\}"\\ mlp\_module\_tmp: "model.layers.\{\}.mlp"\\ attn\_module\_tmp: "model.layers.\{\}.self\_attn"\\ ln\_f\_module: "model.norm"\\ lm\_head\_module: "lm\_head"\\ model\_parallel: true\end{tabular} \\ \hline
% \end{tabular}
% }
% \caption{Hyper parameter values (Most of the default values extend from ROME and MEMIT setup).}
% \label{tab:hyper}
% \end{table}

\begin{table}[h]
\centering
%\small
\resizebox{.38\textwidth}{!}{
\begin{tabular}{|l|l|}
\hline
\multicolumn{2}{|c|}{\textbf{Hyperparameter values}} \\ \hline
layers & {[}5{]} \\
fact\_token & subject\_last \\
v\_num\_grad\_steps & 25 \\
v\_lr & 5e-1 \\
v\_loss\_layer & 31 \\
v\_weight\_decay & 1e-3 \\
clamp\_norm\_factor & 4 \\
kl\_factor & 0.0625 \\
mom2\_adjustment & false \\
context\_template\_length\_params & {[}{[}5, 10{]}, {[}10, 10{]}{]} \\
rewrite\_module\_tmp & model.layers.\{\}.mlp.down\_proj \\
layer\_module\_tmp & model.layers.\{\} \\
mlp\_module\_tmp & model.layers.\{\}.mlp \\
attn\_module\_tmp & model.layers.\{\}.self\_attn \\
ln\_f\_module & model.norm \\
lm\_head\_module & lm\_head \\
model\_parallel & true \\
\hline
\end{tabular}
}
\caption{Hyperparameter values (most of the default values extend from ROME and MEMIT setup).}
\label{tab:hyper}
\end{table}

\if{0}
\section{Results from another model}
\noindent \textbf{Gemma-7b-it}\footnote{https://huggingface.co/google/gemma-7b-it}: Gemma-7b-it~\cite{gemmateam2024gemma}, a new suite of models from Google. As a family of lightweight, cutting-edge open models, Gemma excels in text-to-text tasks through its decoder-only architecture. Gemma come in both pre-trained and instruction-tuned variants, making them highly adaptable for diverse text generation needs like question answering, summarization, and complex reasoning. Available in both 2 and 7-billion parameter configurations, Gemma stands out by employing an expansive token set of 256k unique tokens, dwarfing the conventional 50k. This extensive range of tokens allows for a notably broader embeddings space, offering valuable insights into the impact of increased linguistic diversity on model performance.

\section{Extended Results}
\label{sec:sppendixresults}

  \begin{figure*}[h]
  \centering
    \includegraphics[width=0.8\textwidth]{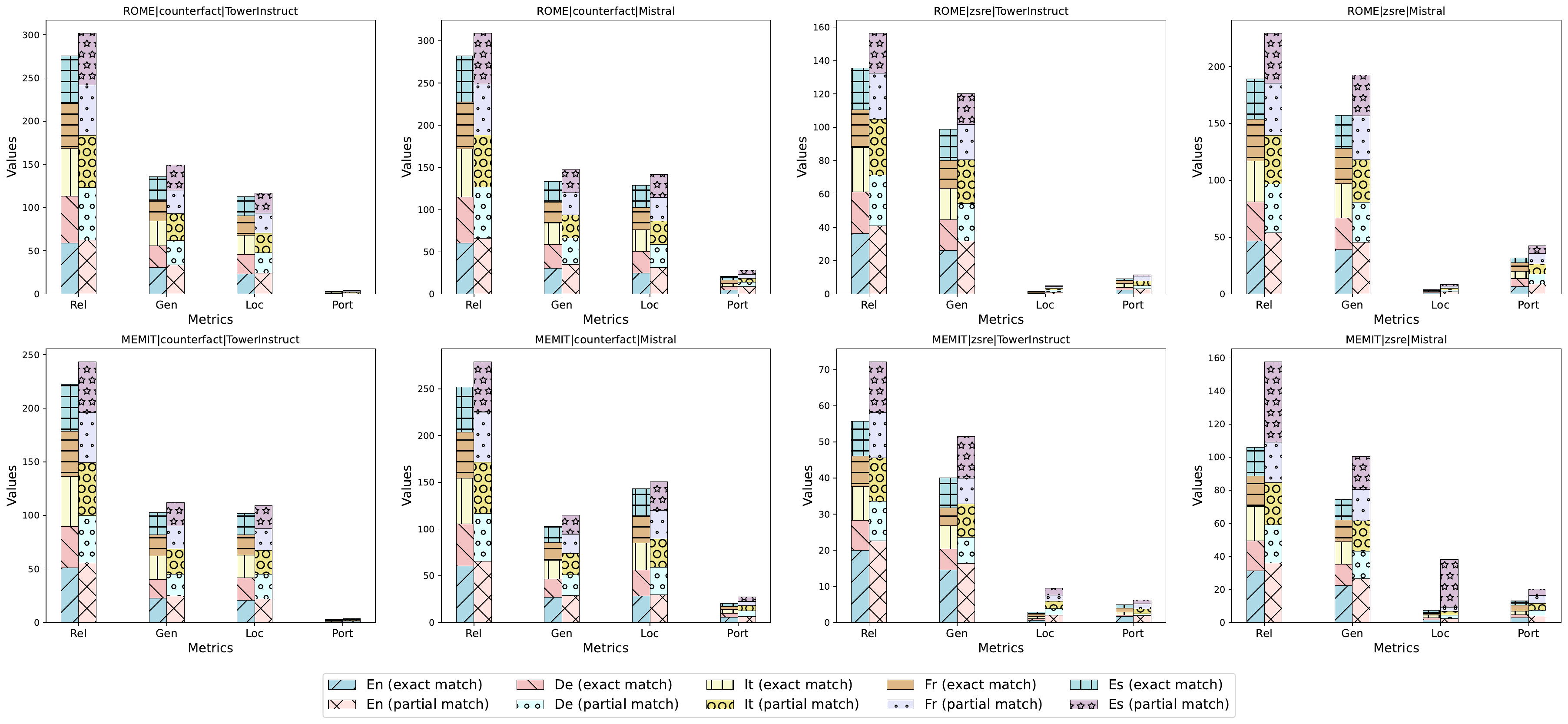} %[width=\linewidth]
    \caption{Each metric on the x-axis is represented by two bars: the left bar indicates an exact match, while the right bar indicates a partial match. For each bar, the divisions along the y-axis reflect the average values of the metric, aggregated across romance and germanic languages evaluated. These subdivisions are color-coded to denote the editing language, as specified in the legend. }\label{fig:enhCon}
    %\vspace*{-0.6cm}
  \end{figure*}

\subsection{Language perspective}

\subsubsection{Counterfact}
In case of Counterfact dataset, significant disparities is observed in edited model performance across different languages. When edit done with ~\emph{en}, tested with ~\emph{En} consistently showed high relevance scores across all models, with Mistral achieving nearly perfect relevance at 0.994 and TowerInstruct at 0.996. However, performance in ~\emph{De}, ~\emph{it}, ~\emph{Fr}, and ~\emph{Es} was notably lower, particularly in generality (in between $\sim$0.28-0.35 for Mistral) and locality (0.32-0.36 for Mistral) metrics, indicating challenges in generalization and nuanced information processing in non-English contexts. The portability scores were modest across the board, underscoring a pronounced need for enhanced multilingual model adaptability. %This analysis stresses the importance of improving models to perform equitably across a broader linguistic spectrum, not just in understanding but in processing and adapting to diverse linguistic structures.

When the edit is conducted with ~\emph{De}, German (De) shows robust relevance scores, particularly with TowerInstruct (0.828) and Mistral (0.834), indicating strong contextual understanding. However, other languages like Italian (It), French (Fr), and Spanish (Es) exhibit lower scores, reflecting challenges in language-specific processing. Gemma-7b underperforms significantly in all languages except German, where it achieves a moderate relevance of 0.625. The models generally show weaker performance in Gen and Loc metrics, suggesting difficulties in generalization and localized understanding across linguistic variations. The Port scores are particularly low, highlighting a critical need for improved model adaptability to diverse language structures.

% After editing with ~\emph{it} language, it demonstrate the highest relevance, particularly with TowerInstruct, achieving a Rel score of 0.871. This indicates a robust ability of the models to interpret Italian context accurately, likely due to focused training. However, the performance in other languages showed variability; for instance, relevance scores were noticeably lower in German and Spanish, suggesting difficulties in model transferability across linguistic boundaries. Generality scores were moderate across languages but showed the models' inconsistent ability to apply learned concepts from Italian to other languages, with notable drops in performance in French and Spanish. Locality scores were generally low across all languages, indicating a widespread challenge in pinpointing specific linguistic features effectively. Portability scores highlighted significant limitations in adapting Italian-trained models to other linguistic frameworks, particularly in non-Romance languages. These results underscore the necessity for models to undergo more comprehensive and nuanced multilingual training to ensure equitable performance across diverse linguistic contexts.

After editing the model with ~\emph{it} language, the edited model achieved the highest reliability score with TowerInstruct (0.871). However, the reliability scores in other languages were lower, with ~\emph{En} at 0.535, ~\emph{De} at 0.398, ~\emph{Fr} at 0.490, and ~\emph{Es} at 0.487, reflecting the challenge of extending training efficiencies beyond Italian. In case of generalization, Gemma-7b performed best with a score of 0.680 in ~\emph{it} but showed a decline in other languages, achieving 0.520 in both ~\emph{De} and ~\emph{Fr}. %This inconsistency suggests limitations in the models' ability to apply Italian-based learnings universally. 
Locality scores were uniformly low across all languages, indicating a persistent difficulty in identifying language-specific information, with scores never exceeding 0.257 in ~\emph{it} and dropping to zero in ~\emph{Fr} and ~\emph{De} for Gemma-7b.
%Portability, measuring the adaptability of Italian-trained models to other languages, also revealed challenges. 
The highest portability score was seen in ~\emph{it} with Mistral and TowerInstruct at 0.092, the scores were significantly lower in other languages, notably dropping to 0.040 in ~\emph{Es} for Gemma-7b. 

In case of edit with ~\emph{Fr} language, ~\emph{it} language achieved the highest scores, with the TowerInstruct model where it reached 0.454, compared to model's performance in other languages like ~\emph{En} (0.530), ~\emph{De} (0.383), ~\emph{Fr} (0.827), and ~\emph{Es} (0.458). This high score in ~\emph{Fr} for TowerInstruct, however, suggests that certain models can still effectively align with training data even in non-primary languages. In case generality (Gen) and locality (Loc), the scores were universally lower across all models and languages, indicating a struggle in generalizing the ~\emph{Fr} editing. For example, Gemma-7b's generality score in Italian stood at 0.360, comparable to its performance in ~\emph{it}, but dipped significantly in ~\emph{en} (0.320) and Spanish (0.360). Locality scores also pointed to difficulties in identifying language-specific nuances, with TowerInstruct showing a modestly better understanding in Italian (0.189) and French (0.214), yet still remaining low. 
%Portability (Port) scores highlighted ongoing challenges, with all models showing limited adaptability to languages other than Italian, demonstrating the critical need for enhanced training approaches to foster better multilingual support and robustness across diverse linguistic environments.

After editing with ~\emph{Es}, ~\emph{it} consistently demonstrated superior reliability score in TowerInstruct of 0.451, which starkly contrasted with lower scores in other languages, such as 0.391 in ~\emph{de} and 0.555 in ~\emph{en}. However, ~\emph{es} exhibited notably high relevance scores, with TowerInstruct achieving 0.822 and Mistral 0.812, indicating these models' effective adaptation to Spanish linguistic features. Generality and locality metrics, which measure a model’s ability to generalize training and identify language-specific information, respectively, showed universally lower scores across all languages, highlighting challenges in cross-lingual applicability. For instance, Gemma-7b's generality in Spanish reached 0.667, surpassing its score in Italian (0.500). Portability results were generally modest but slightly better in Spanish, where TowerInstruct scored 0.097. These findings emphasize the need for enhanced multilingual training strategies to improve overall model robustness and linguistic versatility.

\subsubsection{ZSRE}
  \begin{figure*}[h]
  \centering
    \includegraphics[width=0.8\textwidth]{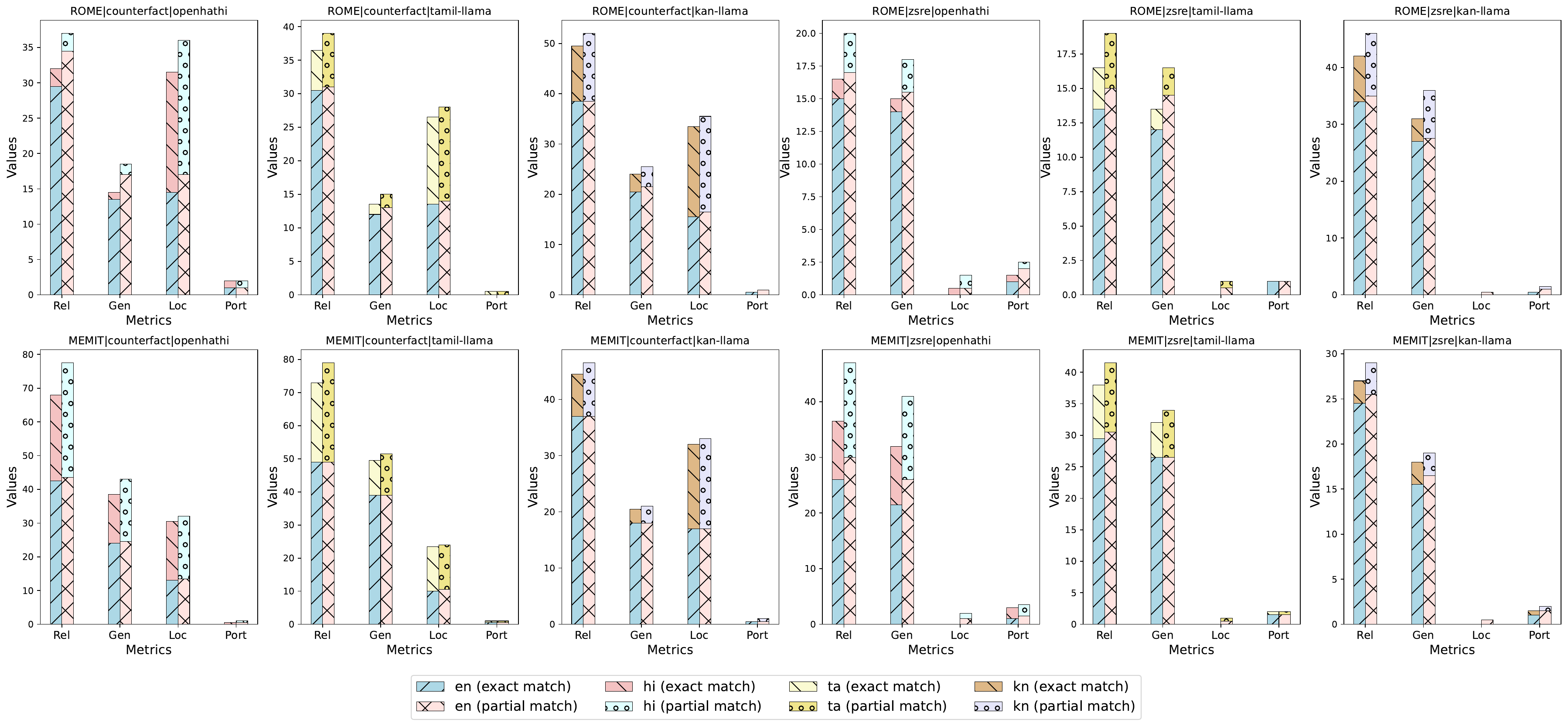} %[width=\linewidth]
    \caption{Each metric on the x-axis is represented by two bars: the left bar indicates an exact match, while the right bar indicates a partial match. For each bar, the divisions along the y-axis reflect the average values of the metric, aggregated across all indic languages evaluated. These subdivisions are color-coded to denote the editing language, as specified in the legend. }\label{fig:indicCon}
    %\vspace*{-0.6cm}
  \end{figure*}
After editing with ~\emph{en} language, the reliability score for mistral model in ~\emph{en} was remarkably high at 0.929. However, this contrasts sharply with its performance in other languages such as ~\emph{de} (0.344) and ~\emph{it} (0.312), suggesting a significant drop in model effectiveness when transitioning from ~\emph{en}. Similarly, the TowerInstruct model showed a strong performance in ~\emph{en} with a relevance score of 0.875, yet scores in other languages like ~\emph{de} (0.236) and ~\emph{fr} (0.221) were markedly lower, highlighting the challenges in maintaining model performance across linguistic boundaries.
In case of generalization (Gen) and locality (Loc), the scores also emphasizes the disparity. While the Mistral model displayed a solid generality in ~\emph{English} with a score of 0.811, its scores in languages such as ~\emph{de} and ~\emph{it} were only around 0.260. This trend of decreased performance is echoed in the locality scores, where Mistral exhibited almost no ability to identify language-specific nuances in ~\emph{it} and ~\emph{fr}. TowerInstruct’s portability (Port) score for ~\emph{en} was 0.125, which, although not very high, still outperforms its ~\emph{de} and ~\emph{fr} counterparts, suggesting a somewhat better but still limited ability to adapt training across languages. %These results underline the need for models to enhance their cross-lingual applicability to ensure consistency and effectiveness across diverse linguistic datasets.

After editing with ~\emph{De}, the TowerInstruct model exhibited significant variations in reliability (Rel) scores, achieving its highest in ~\emph{it} (0.480) but only 0.157 in ~\emph{de}, indicating a substantial challenge in adapting to ~\emph{de} compared to other languages. Similarly, the Mistral model displayed relatively better relevance in ~\emph{de} at 0.382, but this still fell short compared to its performance in ~\emph{it} (0.513), suggesting a consistent trend of models performing better in Romance languages. Further examination into generalization (Gen) and locality (Loc) metrics highlights these disparities even more. For instance, generalization scores for Mistral in ~\emph{de} stood at 0.349, closely aligned with its score in ~\emph{it}, yet locality scores were nearly zero across the board, showing a significant deficiency in capturing language-specific details. Portability (Port) scores also reflect limited adaptability, with Mistral scoring only 0.046 in ~\emph{de} compared to a slightly better performance in ~\emph{it} (0.039), underscoring the need for model training approaches that better address and bridge these linguistic gaps to enhance overall performance and applicability across diverse linguistic datasets.

After editing with ~\emph{it} language, the TowerInstruct model exhibited a disparity in reliability (Rel) scores, achieving a high of 0.537 in ~\emph{Fr} but only 0.185 in ~\emph{it}, underscoring a significant challenge in adapting to ~\emph{it} compared to other Romanic languages. Similarly, the Mistral model demonstrated better reliability in ~\emph{fr} (0.575) than in ~\emph{it} (0.288), further indicating that models tend to align more effectively with training data in certain languages over others. In terms of generality (Gen) and locality (Loc), the scores further emphasize these challenges. The generality score for TowerInstruct in Italian was notably lower (0.346) compared to its performance in French (0.418), suggesting difficulties in generalizing the Italian-based training effectively across languages. Locality scores were uniformly low, indicating a universal struggle among the models to identify language-specific nuances, with the highest locality score for Italian being just 0.033 in Gemma-7b. This pattern is mirrored in the portability (Port) scores, where even the highest scores are relatively modest (e.g., TowerInstruct's 0.107 in Italian), reflecting a constrained ability of models to adapt their Italian training across different linguistic contexts efficiently. These findings highlight a critical need for enhanced training methodologies that better accommodate the linguistic complexities inherent in Italian and other similar languages.

After editing with ~\emph{fr} language, the TowerInstruct model demonstrated a stronger performance in Spanish with a relevance score of 0.507 and a generality score of 0.281, compared to its performance in French (Rel: 0.197, Gen: 0.281) and Italian (Rel: 0.153, Gen: 0.167). This indicates a more robust alignment with Spanish linguistic features. Conversely, the Mistral model exhibited its highest relevance in Spanish (0.517) but struggled in German (0.205) and Italian (0.159), further underscoring the varying model efficiencies across languages. These findings highlight significant challenges in model training, where improvements are needed to enhance language-specific understanding and adaptability, ensuring models perform consistently well across a diverse linguistic spectrum.

After editing with ~\emph{es} language, For instance, the TowerInstruct model achieved a high relevance (Rel) score of 0.443 for Spanish, significantly surpassing its scores in other languages such as English (0.240) and German (0.232). This trend suggests a stronger model alignment with Spanish linguistic properties. In generality (Gen), the TowerInstruct again highlighted its proficiency in Spanish with a score of 0.300, contrasted with lower scores in Italian (0.210) and French (0.182). The locality (Loc) scores were generally low across all languages, indicating a challenge in models' ability to identify and adapt to language-specific nuances. Furthermore, the portability alternative (Port) score for TowerInstruct in Spanish was the highest at 0.050, demonstrating better adaptability when compared to other languages like German (0.039) and Italian (0.034). These metrics underscore the imperative for ongoing optimization in models to ensure consistent performance across a diverse linguistic spectrum.\fi

\section{Worked-out Example}
For instance, a model’s recognition of ``\textit{Dent Island Light, located in: Belgium}'' \textbf{(Post Edit)} (see Figure~\ref{fig:enhCon} should be consistent, irrespective of the language employed. Such consistency is crucial for ensuring a uniform user experience across different languages, thereby democratizing access to information and technology.

  \begin{figure*}
  \centering
    \includegraphics[scale=0.40]{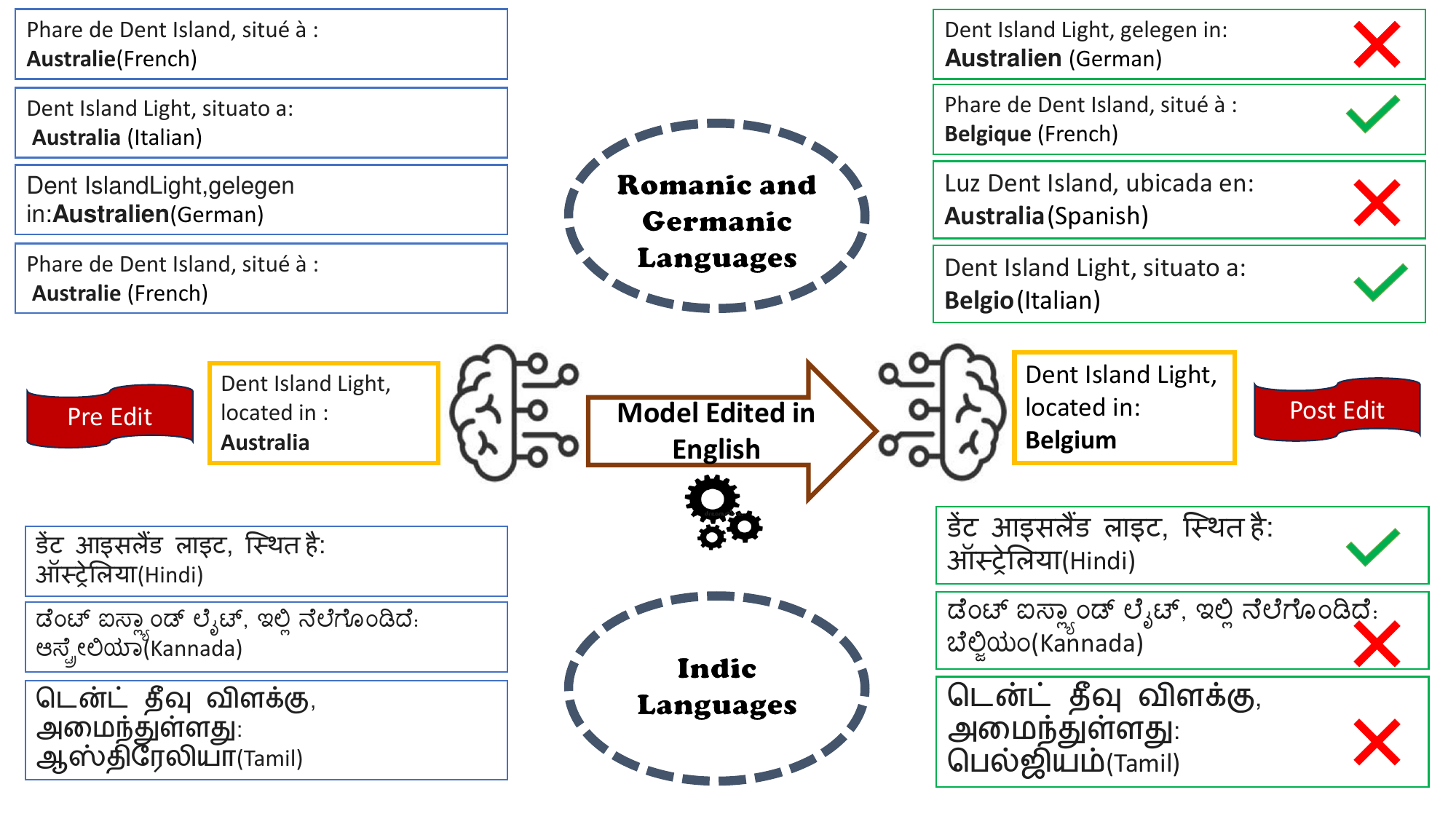} %[width=\linewidth]
    \caption{\label{fig:enhCon}Edited knowledge conflict across various languages for TowerInstruct.}
    %\vspace*{-0.6cm}
  \end{figure*}
  
\section{Exact vs partial match}
We showcase plot correlations in Figures~\ref{fig:enhCon} and~\ref{fig:indicCon}.
\begin{figure*}[h]
  \centering
    \includegraphics[width=1.0\textwidth]{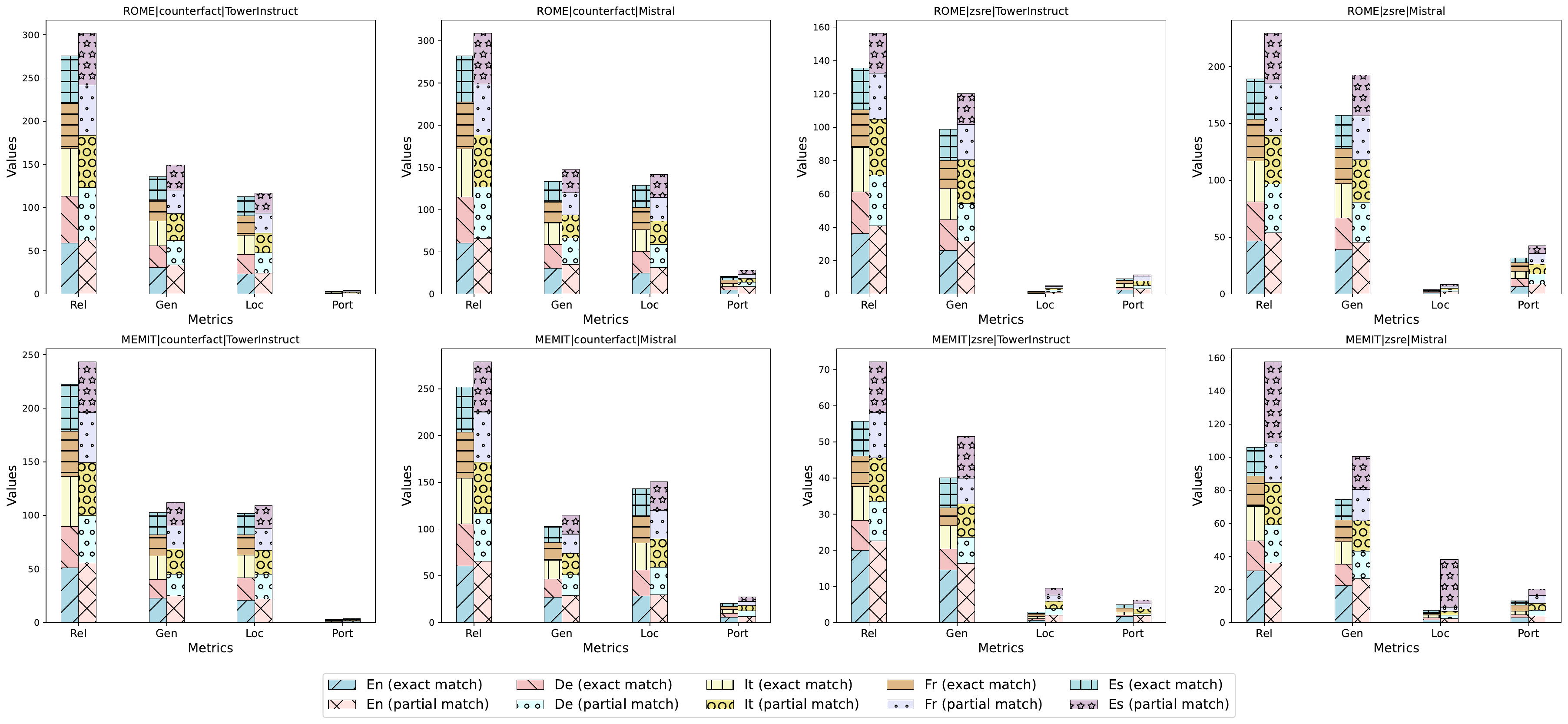} %[width=\linewidth]
    \caption{Each metric on the $x$-axis is represented by two bars: the left bar indicates an exact match, while the right bar indicates a partial match. For each bar, the divisions along the $y$-axis reflect the average values of the metric, aggregated across Romance and Germanic languages evaluated. These subdivisions are color-coded to denote the editing language, as specified in the legend. }\label{fig:enhCon}
    %\vspace*{-0.6cm}
  \end{figure*}

  \begin{figure*}[h]
  \centering
    \includegraphics[width=1.0\textwidth]{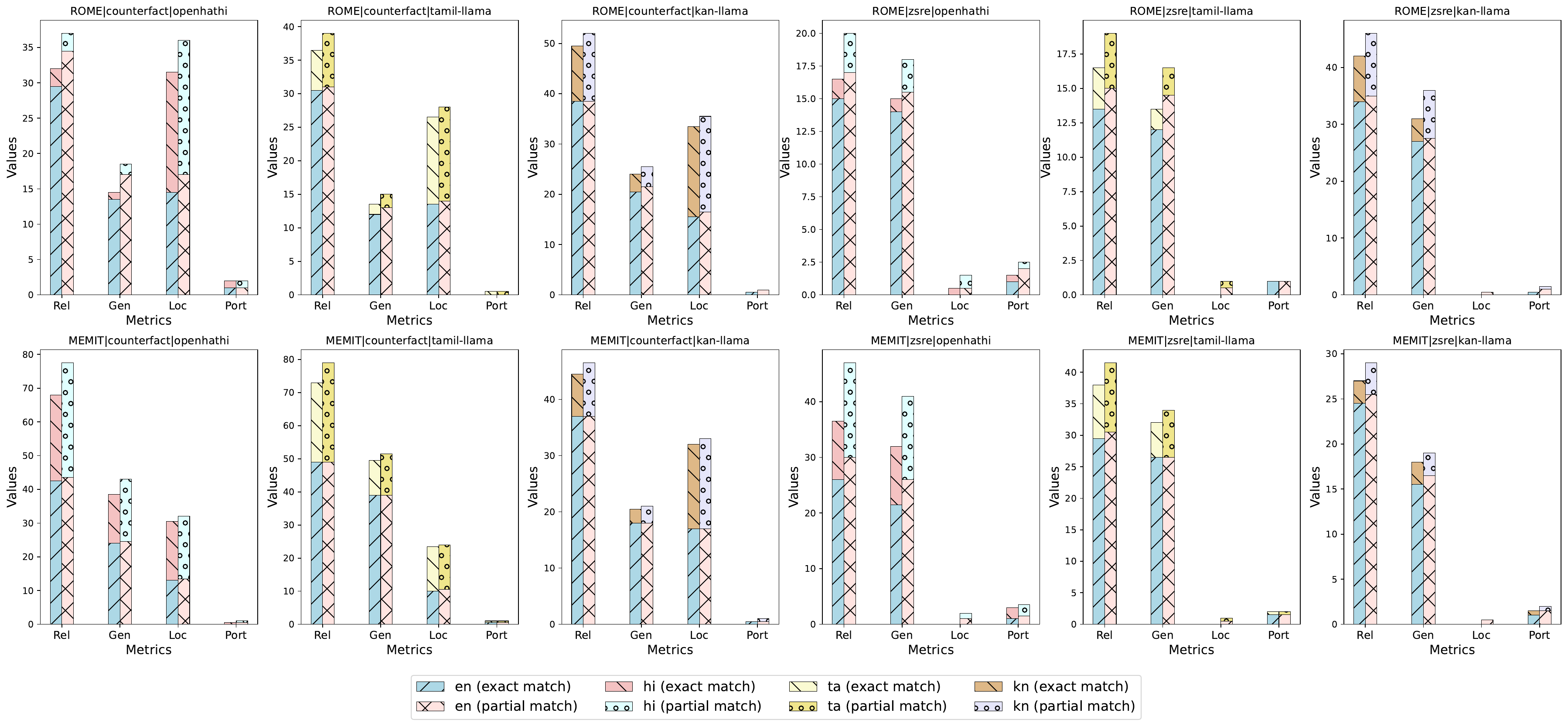} %[width=\linewidth]
    \caption{Each metric on the $x$-axis is represented by two bars: the left bar indicates an exact match, while the right bar indicates a partial match. For each bar, the divisions along the $y$-axis reflect the average values of the metric, aggregated across all Indic languages evaluated. These subdivisions are color-coded to denote the editing language, as specified in the legend. }\label{fig:indicCon}
    %\vspace*{-0.6cm}
  \end{figure*}
%\am{Plot correlations}
\section{Romance and Germanic languages}

\subsection{Language perspective}

\subsubsection{\ctr{}}
In case of \ctr{} dataset, significant disparities are observed in edited model performance across different languages. Edits done with \textbf{En} and tested on \textbf{En} consistently showed high reliability scores across all models, with \mistral{} achieving nearly perfect reliability at 0.994 and \tower{} at 0.996 (for ROME). However, performances while testing with \textbf{De}, \textbf{It}, \textbf{Fr}, and \textbf{Es} were notably lower, particularly in generalisation (in between $\sim$0.21-0.28 for \mistral{}) and locality (0.20-0.28 for \mistral{}) metrics, indicating challenges in generalization and nuanced information processing in non-English contexts. The portability scores were modest across the board, underscoring a pronounced need for enhanced multilingual model adaptability. %This analysis stresses the importance of improving models to perform equitably across a broader linguistic spectrum, not just in understanding but in processing and adapting to diverse linguistic structures.

When the edit is conducted with \textbf{De} and tested on \textbf{De} reliability scores for \tower{} (0.828) and \mistral{} (0.834) (for ROME) are reasonably high indicating strong contextual understanding. However, testing with other languages like \textbf{It}, \textbf{Fr}, and \textbf{Es} exhibit lower scores, reflecting challenges in language-specific processing. %Gemma-7b underperforms significantly in all languages except German, where it achieves a moderate relevance of 0.625. The models generally show weaker performance in Gen and Loc metrics, suggesting difficulties in generalization and localized understanding across linguistic variations. The Port scores are particularly low, highlighting a critical need for improved model adaptability to diverse language structures.

% After editing with ~\emph{it} language, it demonstrate the highest relevance, particularly with TowerInstruct, achieving a Rel score of 0.871. This indicates a robust ability of the models to interpret Italian context accurately, likely due to focused training. However, the performance in other languages showed variability; for instance, relevance scores were noticeably lower in German and Spanish, suggesting difficulties in model transferability across linguistic boundaries. Generality scores were moderate across languages but showed the models' inconsistent ability to apply learned concepts from Italian to other languages, with notable drops in performance in French and Spanish. Locality scores were generally low across all languages, indicating a widespread challenge in pinpointing specific linguistic features effectively. Portability scores highlighted significant limitations in adapting Italian-trained models to other linguistic frameworks, particularly in non-Romance languages. These results underscore the necessity for models to undergo more comprehensive and nuanced multilingual training to ensure equitable performance across diverse linguistic contexts.

After editing the model with \textbf{It} the edited model achieved the highest reliability score with \tower{} for test language \textbf{It} (0.871) (for ROME). However, the reliability scores for other test languages were lower, with \textbf{En} at 0.535, \textbf{De} at 0.398, \textbf{Fr} at 0.490, and \textbf{Es} at 0.488, reflecting the challenge of extending training efficiencies beyond Italian. %In case of generalization, Gemma-7b performed best with a score of 0.680 in ~\emph{it} but showed a decline in other languages, achieving 0.520 in both ~\emph{De} and ~\emph{Fr}. %This inconsistency suggests limitations in the models' ability to apply Italian-based learnings universally. 
%Locality scores were uniformly low across all languages, indicating a persistent difficulty in identifying language-specific information, with scores never exceeding 0.257 in ~\emph{it} and dropping to zero in ~\emph{Fr} and ~\emph{De} for Gemma-7b.
%Portability, measuring the adaptability of Italian-trained models to other languages, also revealed challenges. 
The highest portability score was seen in \textbf{It} with \mistral{} and \tower{} at 0.095 (for ROME), the scores were significantly lower in other languages. %notably dropping to 0.040 in ~\emph{Es} for Gemma-7b. 
%These results highlight the critical need for enhancing multilingual training protocols to improve models' robustness and their adaptability across diverse linguistic environments, ensuring consistency and accuracy irrespective of the language context.

In case of edit with \textbf{Fr}, test language \textbf{Fr} achieved the highest scores (0.832), with \tower{} where it reached 0.454, compared to model's performance in other languages like \textbf{En} (0.519), \textbf{De} (0.417), \textbf{It} (0.509), and \textbf{Es} (0.511). This high score in \textbf{Fr} for \tower{}, however, suggests that certain models can still effectively align with training data even in non-primary languages. In case generality and locality, the scores were universally lower across all models and languages, indicating a struggle in generalizing the \textbf{Fr} editing. %For example, Gemma-7b's generality score in Italian stood at 0.360, comparable to its performance in ~\emph{it}, but dipped significantly in ~\emph{en} (0.320) and Spanish (0.360). 
Locality scores also pointed to difficulties in identifying language-specific nuances, with \tower{} showing a modestly better understanding in \textbf{It} (0.189) and \textbf{Fr} (0.214), yet still remaining low. 
%Portability (Port) scores highlighted ongoing challenges, with all models showing limited adaptability to languages other than Italian, demonstrating the critical need for enhanced training approaches to foster better multilingual support and robustness across diverse linguistic environments.

After editing with \textbf{Es}, \textbf{En} (0.555) consistently demonstrated superior reliability score for \tower{}, compared to other languages such as \textbf{De} (0.391) and \textbf{It} (0.451) (excluding \textbf{Es}). However, \textbf{Es} exhibited notably high reliability scores, with \tower{} achieving 0.822 and \mistral{} 0.812, indicating these models' effective adaptation to Spanish linguistic features. Generality and locality metrics, which measure a model’s ability to generalize training and identify language-specific information, respectively, showed universally lower scores across all languages, highlighting challenges in cross-lingual applicability. %For instance, Gemma-7b's generality in Spanish reached 0.667, surpassing its score in Italian (0.500). Portability results were generally modest but slightly better in Spanish, where TowerInstruct scored 0.097. These findings emphasize the need for enhanced multilingual training strategies to improve overall model robustness and linguistic versatility.

\subsubsection{\zsre{}}
After editing with \textbf{En} language, the reliability score for \mistral{} model in \textbf{En} was remarkably high at 0.929. However, this contrasts sharply with its performance in other languages such as \textbf{De} (0.344) and \textbf{It} (0.312), suggesting a significant drop in model effectiveness when transitioning from ~\textbf{En}. Similarly, the \tower{} model showed a strong performance when the test langauage was \textbf{En} with a relevance score of 0.875, yet scores in other languages like \textbf{De} (0.236) and \textbf{Fr} (0.221) were markedly lower, highlighting the challenges in maintaining model performance across linguistic boundaries (for ROME).
In case of generalization and locality, the scores also emphasize the disparity. While \mistral{} displayed a good generality in \textbf{Eng} (0.812), its scores in languages such as \textbf{De} and \textbf{It} were only around 0.260. This trend of decreased performance is echoed in the locality scores, where \mistral{} exhibited almost no ability to identify language-specific nuances in \textbf{It} and \textbf{Fr}. \tower{}’s portability score for \textbf{En} was 0.097, which, although not very high, still outperforms its \textbf{De} and \textbf{Fr} counterparts, suggesting a somewhat better but still limited ability to adapt training across languages (for ROME). %These results underline the need for models to enhance their cross-lingual applicability to ensure consistency and effectiveness across diverse linguistic datasets.

After editing with \textbf{De}, the \tower{} model exhibited significant variations in reliability scores, achieving its highest in \textbf{De} (0.480) but only 0.157 in \textbf{En}, indicating a substantial challenge in adapting to \textbf{De} compared to other languages. Similarly, \mistral{} displayed relatively better relevance in \textbf{De} at 0.513, but this still fell short compared to its performance in \textbf{It} (0.257), suggesting a consistent trend of models performing better in Romance languages. Further examination of generalization and locality metrics highlights these disparities even more. For instance, generalization scores for \mistral{} in \textbf{De} stood at 0.349, yet locality scores were nearly zero across the board, showing a significant deficiency in capturing language-specific details. Portability scores also reflect limited adaptability, with \mistral{} scoring only 0.079 for \textbf{De} compared to a slightly better performance in \textbf{It} (0.066), underscoring the need for model training approaches that better address and bridge these linguistic gaps to enhance overall performance and applicability across diverse linguistic datasets (for ROME).

After editing with \textbf{It}, \tower{} model exhibited a disparity in reliability scores, achieving a high value of 0.537 in \textbf{It} but only 0.185 in \textbf{De}, underscoring a significant challenge in adapting to \textbf{De} compared to other Romance languages. Similarly, \mistral{} demonstrated better reliability in \textbf{It} (0.575), further indicating that models tend to align more effectively with training data in certain languages over others. In terms of generality and locality, the scores further emphasize these challenges. %The generality score for TowerInstruct in Italian was notably lower (0.346) compared to its performance in French (0.418), suggesting difficulties in generalizing the Italian-based training effectively across languages. Locality scores were uniformly low, indicating a universal struggle among the models to identify language-specific nuances, with the highest locality score for Italian being just 0.033 in Gemma-7b. This pattern is mirrored in the portability (Port) scores, where even the highest scores are relatively modest (e.g., TowerInstruct's 0.107 in Italian), reflecting a constrained ability of models to adapt their Italian training across different linguistic contexts efficiently. These findings highlight a critical need for enhanced training methodologies that better accommodate the linguistic complexities inherent in Italian and other similar languages.

After editing with \textbf{Fr}, the \tower{} demonstrated a stronger performance in \textbf{Fr} with a reliability score of 0.507 and a generality score of 0.281, compared to its performance in \textbf{Es} (Rel: 0.138, Gen: 0.113) and \textbf{It} (Rel: 0.197, Gen: 0.167). This indicates a more robust alignment with \textbf{Fr} linguistic features. On the other hand, \mistral{} also exhibited its highest reliability in \textbf{Fr} (0.517) but struggled in \textbf{De} (0.298) and \textbf{It} (0.272), further underscoring the varying model efficiencies across languages. These findings highlight significant challenges in model training, where improvements are needed to enhance language-specific understanding and adaptability, ensuring that models perform consistently well across a diverse linguistic spectrum.

After editing with \textbf{Es}, \tower{} achieved a high reliability score of 0.443 for \textbf{Es}, significantly surpassing its scores in other languages such as \textbf{En} (0.232) and \textbf{De} (0.148). This trend suggests a stronger model alignment with the linguistic properties of \textbf{Es}. In generality, \tower{} highlights better performance in \textbf{Es} with a score of 0.305, contrasted with lower scores in \textbf{It} (0.202) and \textbf{Fr} (0.182). The locality scores were generally low across all languages.

\begin{table*}[h]
\centering
\resizebox{1.0\textwidth}{!}{
\begin{tabular}{|cc|l|lllll|lllll|lllll|}
\hline
\multicolumn{2}{|c|}{\multirow{2}{*}{\textbf{\begin{tabular}[c]{@{}c@{}}Datasets/\\ Languages\end{tabular}}}} & \multicolumn{1}{c|}{\multirow{2}{*}{\textbf{Score}}} & \multicolumn{5}{c|}{\textbf{Mistral}}                                                                                                                                                                                                                                  & \multicolumn{5}{c|}{\textbf{TowerInstruct}}                                                                                                             \\ \cline{4-13} 
\multicolumn{2}{|c|}{}                                                                                                  & \multicolumn{1}{c|}{}                                & \multicolumn{1}{l|}{\textbf{En}} & \multicolumn{1}{l|}{\textbf{De}} & \multicolumn{1}{l|}{\textbf{It}} & \multicolumn{1}{l|}{\textbf{Fr}} & \textbf{Es} & \multicolumn{1}{l|}{\textbf{En}} & \multicolumn{1}{l|}{\textbf{De}} & \multicolumn{1}{l|}{\textbf{It}} & \multicolumn{1}{l|}{\textbf{Fr}} & \textbf{Es} \\ \hline
\multicolumn{1}{|c|}{\multirow{30}{*}{\textbf{CounterFact}}}               & \multirow{6}{*}{\textbf{En}}               & \textbf{Rel}                                        & \multicolumn{1}{l|}{0.994/0.994}  & \multicolumn{1}{l|}{0.498/0.560}  & \multicolumn{1}{l|}{0.469/0.578}  & \multicolumn{1}{l|}{0.487/0.548}& 0.571/0.617  & \multicolumn{1}{l|}{0.996/0.996}  & \multicolumn{1}{l|}{0.482/0.529}  & \multicolumn{1}{l|}{0.455/0.500}  & \multicolumn{1}{l|}{0.498/0.527}  & \multicolumn{1}{l|}{0.511/0.562}   \\ \cline{3-13} 
\multicolumn{1}{|c|}{}                                                     &                                            & \textbf{Gen}                                       & \multicolumn{1}{l|}{0.512/0.529}  & \multicolumn{1}{l|}{0.233/0.269}  & \multicolumn{1}{l|}{0.246/0.346}  & \multicolumn{1}{l|}{0.279/0.305}& 0.252/0.294    & \multicolumn{1}{l|}{0.522/0.538}  & \multicolumn{1}{l|}{0.245/0.273}  & \multicolumn{1}{l|}{0.231/0.267}  & \multicolumn{1}{l|}{0.280/0.309}  & \multicolumn{1}{l|}{0.256/0.291}       \\ \cline{3-13} 
\multicolumn{1}{|c|}{}                                                     &                                            & \textbf{Loc}                                        & \multicolumn{1}{l|}{0.327/0.338}  & \multicolumn{1}{l|}{0.227/0.250}  & \multicolumn{1}{l|}{0.240/0.358}  & \multicolumn{1}{l|}{0.200/0.362}& 0.244/0.265   & \multicolumn{1}{l|}{0.307/0.315}  & \multicolumn{1}{l|}{0.196/0.207}  & \multicolumn{1}{l|}{0.204/0.209}  & \multicolumn{1}{l|}{0.225/0.235}  & \multicolumn{1}{l|}{0.224/0.238}      \\ \cline{3-13} 
%\multicolumn{1}{|c|}{}                                                     &                                            & \textbf{Port}                                        & \multicolumn{1}{l|}{0.133}       & \multicolumn{1}{l|}{0.114}       & \multicolumn{1}{l|}{0.095}       & \multicolumn{1}{l|}{0.106}       & 0.106       & \multicolumn{1}{l|}{0.063}       & \multicolumn{1}{l|}{0.037}       & \multicolumn{1}{l|}{0.037}       & \multicolumn{1}{l|}{0.037}       & 0.046       & \multicolumn{1}{l|}{0.005}       & \multicolumn{1}{l|}{0.000}       & \multicolumn{1}{l|}{0.011}       & \multicolumn{1}{l|}{0.005}       & 0.002       \\ \cline{3-18} 
\multicolumn{1}{|c|}{}                                                     &                                            & \textbf{Port}                                   & \multicolumn{1}{l|}{0.133/0.144}  & \multicolumn{1}{l|}{0.029/0.033}  & \multicolumn{1}{l|}{0.027/0.111}  & \multicolumn{1}{l|}{0.013/0.119}& 0.027/0.035   & \multicolumn{1}{l|}{0.005/0.013}  & \multicolumn{1}{l|}{0.000/0.004}  & \multicolumn{1}{l|}{0.011/0.018}  & \multicolumn{1}{l|}{0.005/0.005}  & \multicolumn{1}{l|}{0.002/0.004}      \\ \cline{3-13} 
%\multicolumn{1}{|c|}{}                                                     &      %                                      & \textbf{Port\_Best}                                  & \multicolumn{1}{l|}{0.258}       & \multicolumn{1}{l|}{0.214}       & \multicolumn{1}{l|}{0.181}       & \multicolumn{1}{l|}{0.213}       & 0.198       & \multicolumn{1}{l|}{0.081}       & \multicolumn{1}{l|}{0.050}       & \multicolumn{1}{l|}{0.054}       & \multicolumn{1}{l|}{0.054}       & 0.062       & \multicolumn{1}{l|}{0.065}       & \multicolumn{1}{l|}{0.122}       & \multicolumn{1}{l|}{0.073}       & \multicolumn{1}{l|}{0.104}       & 0.098       \\ 
\cline{2-13} 
\multicolumn{1}{|c|}{}                                                     & \multirow{6}{*}{\textbf{De}}               & \textbf{Rel}                                       & \multicolumn{1}{l|}{0.558/0.591}  & \multicolumn{1}{l|}{0.834/0.961}  & \multicolumn{1}{l|}{0.471/0.506}  & \multicolumn{1}{l|}{0.423/0.471}& 0.446/0.500   & \multicolumn{1}{l|}{0.589/0.614}  & \multicolumn{1}{l|}{0.828/0.959}  & \multicolumn{1}{l|}{0.431/0.489}  & \multicolumn{1}{l|}{0.439/0.481}  & \multicolumn{1}{l|}{0.429/0.497}     \\ \cline{3-13} 
\multicolumn{1}{|c|}{}                                                     &                                            & \textbf{Gen}                                         & \multicolumn{1}{l|}{0.355/0.394}  & \multicolumn{1}{l|}{0.284/0.313}  & \multicolumn{1}{l|}{0.266/0.303}  & \multicolumn{1}{l|}{0.255/0.286}& 0.245/0.282  & \multicolumn{1}{l|}{0.322/0.345}  & \multicolumn{1}{l|}{0.271/0.314}  & \multicolumn{1}{l|}{0.211/0.246}  & \multicolumn{1}{l|}{0.224/0.246}  & \multicolumn{1}{l|}{0.224/0.255}    \\ \cline{3-13} 
\multicolumn{1}{|c|}{}                                                     &                                            & \textbf{Loc}                                         & \multicolumn{1}{l|}{0.365/0.376}  & \multicolumn{1}{l|}{0.208/0.228}  & \multicolumn{1}{l|}{0.251/0.264}  & \multicolumn{1}{l|}{0.193/0.207}& 0.263/0.280  & \multicolumn{1}{l|}{0.287/0.292}  & \multicolumn{1}{l|}{0.222/0.232}  & \multicolumn{1}{l|}{0.212/0.216}  & \multicolumn{1}{l|}{0.214/0.224}  & \multicolumn{1}{l|}{0.211/0.224}     \\ \cline{3-13} 
%\multicolumn{1}{|c|}{}                                                     &      %                                      & \textbf{Port}                                        & \multicolumn{1}{l|}{0.029}       & \multicolumn{1}{l|}{0.029}       & \multicolumn{1}{l|}{0.031}       & \multicolumn{1}{l|}{0.020}       & 0.022       & \multicolumn{1}{l|}{0.042}       & \multicolumn{1}{l|}{0.000}       & \multicolumn{1}{l|}{0.000}       & \multicolumn{1}{l|}{0.000}       & 0.042       & \multicolumn{1}{l|}{0.004}       & \multicolumn{1}{l|}{0.008}       & \multicolumn{1}{l|}{0.004}       & \multicolumn{1}{l|}{0.006}       & 0.000       \\ \cline{3-18} 
\multicolumn{1}{|c|}{}                                                     &                                            & \textbf{Port}                                   & \multicolumn{1}{l|}{0.114/0.133}  & \multicolumn{1}{l|}{0.029/0.039}  & \multicolumn{1}{l|}{0.025/0.027}  & \multicolumn{1}{l|}{0.023/0.023}& 0.033/0.037  & \multicolumn{1}{l|}{0.004/0.014}  & \multicolumn{1}{l|}{0.008/0.008}  & \multicolumn{1}{l|}{0.004/0.006}  & \multicolumn{1}{l|}{0.006/0.012}  & \multicolumn{1}{l|}{0.000/0.002}       \\ \cline{3-13} 
%\multicolumn{1}{|c|}{}                                                     &      %                                      & \textbf{Port\_Best}                                  & \multicolumn{1}{l|}{0.154}       & \multicolumn{1}{l|}{0.139}       & \multicolumn{1}{l|}{0.138}       & \multicolumn{1}{l|}{0.133}       & 0.133       & \multicolumn{1}{l|}{0.042}       & \multicolumn{1}{l|}{0.000}       & \multicolumn{1}{l|}{0.042}       & \multicolumn{1}{l|}{0.000}       & 0.042       & \multicolumn{1}{l|}{0.039}       & \multicolumn{1}{l|}{0.109}       & \multicolumn{1}{l|}{0.060}       & \multicolumn{1}{l|}{0.094}       & 0.082       \\ 
\cline{2-13} 
\multicolumn{1}{|c|}{}                                                     & \multirow{6}{*}{\textbf{It}}               & \textbf{Rel}                                        & \multicolumn{1}{l|}{0.541/0.578}  & \multicolumn{1}{l|}{0.422/0.477}  & \multicolumn{1}{l|}{0.860/0.914}  & \multicolumn{1}{l|}{0.502/0.542}& 0.519/0.582   & \multicolumn{1}{l|}{0.535/0.564}  & \multicolumn{1}{l|}{0.398/0.450}  & \multicolumn{1}{l|}{0.871/0.932}  & \multicolumn{1}{l|}{0.490/0.535}  & \multicolumn{1}{l|}{0.488/0.556}  \\ \cline{3-13} 
\multicolumn{1}{|c|}{}                                                     &                                            & \textbf{Gen}                                         & \multicolumn{1}{l|}{0.319/0.346}  & \multicolumn{1}{l|}{0.202/0.218}  & \multicolumn{1}{l|}{0.278/0.296}  & \multicolumn{1}{l|}{0.235/0.239}& 0.235/0.267    & \multicolumn{1}{l|}{0.330/0.349}  & \multicolumn{1}{l|}{0.226/0.253}  & \multicolumn{1}{l|}{0.346/0.376}  & \multicolumn{1}{l|}{0.263/0.290}  & \multicolumn{1}{l|}{0.268/0.311}     \\ \cline{3-13} 
\multicolumn{1}{|c|}{}                                                     &                                            & \textbf{Loc}                                        & \multicolumn{1}{l|}{0.350/0.358}  & \multicolumn{1}{l|}{0.230/0.251}  & \multicolumn{1}{l|}{0.257/0.270}  & \multicolumn{1}{l|}{0.210/0.264}& 0.253/0.265   & \multicolumn{1}{l|}{0.293/0.301}  & \multicolumn{1}{l|}{0.199/0.205}  & \multicolumn{1}{l|}{0.185/0.189}  & \multicolumn{1}{l|}{0.214/0.222}  & \multicolumn{1}{l|}{0.203/0.216}      \\ \cline{3-13} 
%\multicolumn{1}{|c|}{}                                                     &                                            & \textbf{Port}                                        & \multicolumn{1}{l|}{0.027}       & \multicolumn{1}{l|}{0.025}       & \multicolumn{1}{l|}{0.021}       & \multicolumn{1}{l|}{0.022}       & 0.035       & \multicolumn{1}{l|}{0.080}       & \multicolumn{1}{l|}{0.080}       & \multicolumn{1}{l|}{0.080}       & \multicolumn{1}{l|}{0.080}       & 0.080       & \multicolumn{1}{l|}{0.008}       & \multicolumn{1}{l|}{0.004}       & \multicolumn{1}{l|}{0.019}       & \multicolumn{1}{l|}{0.010}       & 0.006       \\ \cline{3-18} 
\multicolumn{1}{|c|}{}                                                     &                                            & \textbf{Port}                                   & \multicolumn{1}{l|}{0.095/0.111}  & \multicolumn{1}{l|}{0.031/0.045}  & \multicolumn{1}{l|}{0.021/0.031}  & \multicolumn{1}{l|}{0.012/0.023}& 0.019/0.031   & \multicolumn{1}{l|}{0.008/0.010}  & \multicolumn{1}{l|}{0.004/0.004}  & \multicolumn{1}{l|}{0.019/0.021}  & \multicolumn{1}{l|}{0.010/0.012}  & \multicolumn{1}{l|}{0.006/0.006}    \\ \cline{3-13} 
%\multicolumn{1}{|c|}{}                                                     &                                            & \textbf{Port\_Best}                                  & \multicolumn{1}{l|}{0.108}       & \multicolumn{1}{l|}{0.095}       & \multicolumn{1}{l|}{0.086}       & \multicolumn{1}{l|}{0.104}       & 0.110       & \multicolumn{1}{l|}{0.120}       & \multicolumn{1}{l|}{0.080}       & \multicolumn{1}{l|}{0.080}       & \multicolumn{1}{l|}{0.080}       & 0.080       & \multicolumn{1}{l|}{0.066}       & \multicolumn{1}{l|}{0.106}       & \multicolumn{1}{l|}{0.075}       & \multicolumn{1}{l|}{0.085}       & 0.102       \\ 
\cline{2-13} 
\multicolumn{1}{|c|}{}                                                     & \multirow{6}{*}{\textbf{Fr}}               & \textbf{Rel}                                        & \multicolumn{1}{l|}{0.519/0.548}  & \multicolumn{1}{l|}{0.417/0.485}  & \multicolumn{1}{l|}{0.509/0.542}  & \multicolumn{1}{l|}{0.832/0.890}& 0.511/0.566   & \multicolumn{1}{l|}{0.530/0.550}  & \multicolumn{1}{l|}{0.383/0.440}  & \multicolumn{1}{l|}{0.454/0.501}  & \multicolumn{1}{l|}{0.827/0.898}  & \multicolumn{1}{l|}{0.458/0.506}  \\ \cline{3-13} 
\multicolumn{1}{|c|}{}                                                     &                                            & \textbf{Gen}                                       & \multicolumn{1}{l|}{0.282/0.305}  & \multicolumn{1}{l|}{0.190/0.215}  & \multicolumn{1}{l|}{0.219/0.239}  & \multicolumn{1}{l|}{0.294/0.297}& 0.252/0.268   & \multicolumn{1}{l|}{0.281/0.297}  & \multicolumn{1}{l|}{0.200/0.222}  & \multicolumn{1}{l|}{0.208/0.230}  & \multicolumn{1}{l|}{0.308/0.330}  & \multicolumn{1}{l|}{0.234/0.281}  \\ \cline{3-13} 
\multicolumn{1}{|c|}{}                                                     &                                            & \textbf{Loc}                                        & \multicolumn{1}{l|}{0.350/0.362}  & \multicolumn{1}{l|}{0.243/0.256}  & \multicolumn{1}{l|}{0.249/0.264}  & \multicolumn{1}{l|}{0.204/0.217}& 0.276/0.294   & \multicolumn{1}{l|}{0.303/0.316}  & \multicolumn{1}{l|}{0.204/0.214}  & \multicolumn{1}{l|}{0.189/0.198}  & \multicolumn{1}{l|}{0.214/0.220}  & \multicolumn{1}{l|}{0.224/0.208}      \\ \cline{3-13} 
%\multicolumn{1}{|c|}{}                                                     &                                            & \textbf{Port}                                        & \multicolumn{1}{l|}{0.013}       & \multicolumn{1}{l|}{0.023}       & \multicolumn{1}{l|}{0.012}       & \multicolumn{1}{l|}{0.029}       & 0.023       & \multicolumn{1}{l|}{0.080}       & \multicolumn{1}{l|}{0.000}       & \multicolumn{1}{l|}{0.080}       & \multicolumn{1}{l|}{0.120}       & 0.080       & \multicolumn{1}{l|}{0.006}       & \multicolumn{1}{l|}{0.010}       & \multicolumn{1}{l|}{0.010}       & \multicolumn{1}{l|}{0.004}       & 0.002       \\ \cline{3-18} 
\multicolumn{1}{|c|}{}                                                     &                                            & \textbf{Port}                                   & \multicolumn{1}{l|}{0.106/0.119}  & \multicolumn{1}{l|}{0.020/0.025}  & \multicolumn{1}{l|}{0.022/0.023}  & \multicolumn{1}{l|}{0.029/0.033}& 0.023/0.029  & \multicolumn{1}{l|}{0.006/0.018}  & \multicolumn{1}{l|}{0.010/0.016}  & \multicolumn{1}{l|}{0.010/0.012}  & \multicolumn{1}{l|}{0.004/0.006}  & \multicolumn{1}{l|}{0.002/0.008}   \\ \cline{3-13} 
%\multicolumn{1}{|c|}{}                                                     &                                            & \textbf{Port\_Best}                                  & \multicolumn{1}{l|}{0.117}       & \multicolumn{1}{l|}{0.120}       & \multicolumn{1}{l|}{0.088}       & \multicolumn{1}{l|}{0.129}       & 0.114       & \multicolumn{1}{l|}{0.240}       & \multicolumn{1}{l|}{0.040}       & \multicolumn{1}{l|}{0.080}       & \multicolumn{1}{l|}{0.120}       & 0.080       & \multicolumn{1}{l|}{0.049}       & \multicolumn{1}{l|}{0.098}       & \multicolumn{1}{l|}{0.075}       & \multicolumn{1}{l|}{0.100}       & 0.108       \\ 
\cline{2-13} 
\multicolumn{1}{|c|}{}                                                     & \multirow{6}{*}{\textbf{Es}}               & \textbf{Rel}                                        & \multicolumn{1}{l|}{0.528/0.548}  & \multicolumn{1}{l|}{0.409/0.458}  & \multicolumn{1}{l|}{0.483/0.542}  & \multicolumn{1}{l|}{0.489/0.544}& 0.812/0.908   & \multicolumn{1}{l|}{0.555/0.581}  & \multicolumn{1}{l|}{0.391/0.429}  & \multicolumn{1}{l|}{0.451/0.516}  & \multicolumn{1}{l|}{0.466/0.554}  & \multicolumn{1}{l|}{0.822/0.921}       \\ \cline{3-13} 
\multicolumn{1}{|c|}{}                                                     &                                            & \textbf{Gen}                                        & \multicolumn{1}{l|}{0.297/0.321}  & \multicolumn{1}{l|}{0.194/0.217}  & \multicolumn{1}{l|}{0.241/0.272}  & \multicolumn{1}{l|}{0.231/0.252}& 0.280/0.315   & \multicolumn{1}{l|}{0.318/0.340}  & \multicolumn{1}{l|}{0.184/0.219}  & \multicolumn{1}{l|}{0.233/0.251}  & \multicolumn{1}{l|}{0.265/0.263}  & \multicolumn{1}{l|}{0.330/0.372}   \\ \cline{3-13} 
\multicolumn{1}{|c|}{}                                                     &                                            & \textbf{Loc}                                         & \multicolumn{1}{l|}{0.346/0.358}  & \multicolumn{1}{l|}{0.235/0.250}  & \multicolumn{1}{l|}{0.249/0.262}  & \multicolumn{1}{l|}{0.209/0.223}& 0.254/0.268  & \multicolumn{1}{l|}{0.294/0.300}  & \multicolumn{1}{l|}{0.211/0.217}  & \multicolumn{1}{l|}{0.186/0.188}  & \multicolumn{1}{l|}{0.200/0.238}  & \multicolumn{1}{l|}{0.211/0.223}       \\ \cline{3-13} 
%\multicolumn{1}{|c|}{}                                                     &                                            & \textbf{Port}                                        & \multicolumn{1}{l|}{0.027}       & \multicolumn{1}{l|}{0.033}       & \multicolumn{1}{l|}{0.019}       & \multicolumn{1}{l|}{0.023}       & 0.029       & \multicolumn{1}{l|}{0.042}       & \multicolumn{1}{l|}{0.042}       & \multicolumn{1}{l|}{0.000}       & \multicolumn{1}{l|}{0.042}       & 0.083       & \multicolumn{1}{l|}{0.008}       & \multicolumn{1}{l|}{0.002}       & \multicolumn{1}{l|}{0.008}       & \multicolumn{1}{l|}{0.008}       & 0.000       \\ \cline{3-18} 
\multicolumn{1}{|c|}{}                                                     &                                            & \textbf{Port}                                   & \multicolumn{1}{l|}{0.106/0.123}  & \multicolumn{1}{l|}{0.022/0.023}  & \multicolumn{1}{l|}{0.035/0.037}  & \multicolumn{1}{l|}{0.023/0.025}& 0.029/0.033    & \multicolumn{1}{l|}{0.008/0.014}  & \multicolumn{1}{l|}{0.002/0.002}  & \multicolumn{1}{l|}{0.008/0.014}  & \multicolumn{1}{l|}{0.010/0.020}  & \multicolumn{1}{l|}{0.000/0.002}       \\ \cline{1-13} 
%\multicolumn{1}{|c|}{}                                                     &                                            & \textbf{Port\_Best}                                  & \multicolumn{1}{l|}{0.146}       & \multicolumn{1}{l|}{0.135}       & \multicolumn{1}{l|}{0.125}       & \multicolumn{1}{l|}{0.127}       & 0.145       & \multicolumn{1}{l|}{0.083}       & \multicolumn{1}{l|}{0.083}       & \multicolumn{1}{l|}{0.042}       & \multicolumn{1}{l|}{0.083}       & 0.083       & \multicolumn{1}{l|}{0.045}       & \multicolumn{1}{l|}{0.085}       & \multicolumn{1}{l|}{0.073}       & \multicolumn{1}{l|}{0.091}       & 0.097       \\ \hline
\end{tabular}
}
\caption{Comparison of reliability (Rel), generalization (Gen), locality (Loc), and portability (Port) scores for multiple language models evaluated using the CounterFact dataset and the ROME editing method. The second column indicates the language in which each model was edited.}
\label{tab:counter_rome}
% \end{adjustbox}
\end{table*}

% \usepackage{multirow}
\begin{table*}[h]
\centering
\resizebox{1.0\textwidth}{!}{
\begin{tabular}{|cc|l|lllll|lllll|lllll|}
\hline
\multicolumn{2}{|c|}{\multirow{2}{*}{\textbf{\begin{tabular}[c]{@{}c@{}}Datasets/\\ Languages\end{tabular}}}} & \multicolumn{1}{c|}{\multirow{2}{*}{\textbf{Score}}} & \multicolumn{5}{c|}{\textbf{Mistral}}                                                                                                                                                                                                                                     & \multicolumn{5}{c|}{\textbf{TowerInstruct}}                                                                                                             \\ \cline{4-13} 
\multicolumn{2}{|c|}{}                                                                                                  & \multicolumn{1}{c|}{}                                & \multicolumn{1}{l|}{\textbf{En}} & \multicolumn{1}{l|}{\textbf{De}} & \multicolumn{1}{l|}{\textbf{It}} & \multicolumn{1}{l|}{\textbf{Fr}} & \textbf{Es} & \multicolumn{1}{l|}{\textbf{En}} & \multicolumn{1}{l|}{\textbf{De}} & \multicolumn{1}{l|}{\textbf{It}} & \multicolumn{1}{l|}{\textbf{Fr}} & \textbf{Es} \\ \hline
\multicolumn{1}{|c|}{\multirow{30}{*}{\textbf{ZSRE}}}                   & \multirow{6}{*}{\textbf{En}}                  & \textbf{Rel}                                         & \multicolumn{1}{l|}{0.929/0.981}  & \multicolumn{1}{l|}{0.344/0.448}  & \multicolumn{1}{l|}{0.312/0.344}  & \multicolumn{1}{l|}{0.364/0.442}& 0.390/0.481    & \multicolumn{1}{l|}{0.875/0.928}  & \multicolumn{1}{l|}{0.236/0.279}  & \multicolumn{1}{l|}{0.240/0.293}  & \multicolumn{1}{l|}{0.221/0.260}  & \multicolumn{1}{l|}{0.240/0.288}      \\ \cline{3-13} 
\multicolumn{1}{|c|}{}                                                  &                                               & \textbf{Gen}                                         & \multicolumn{1}{l|}{0.812/0.851}  & \multicolumn{1}{l|}{0.260/0.351}  & \multicolumn{1}{l|}{0.260/0.325}  & \multicolumn{1}{l|}{0.292/0.331}& 0.331/0.409    & \multicolumn{1}{l|}{0.620/0.683}  & \multicolumn{1}{l|}{0.183/0.226}  & \multicolumn{1}{l|}{0.168/0.216}  & \multicolumn{1}{l|}{0.149/0.207}  & \multicolumn{1}{l|}{0.183/0.255}       \\ \cline{3-13} 
\multicolumn{1}{|c|}{}                                                  &                                               & \textbf{Loc}                                         & \multicolumn{1}{l|}{0.000/0.006}  & \multicolumn{1}{l|}{0.013/0.019}  & \multicolumn{1}{l|}{0.000/0.019}  & \multicolumn{1}{l|}{0.013/0.026}& 0.013/0.019    & \multicolumn{1}{l|}{0.010/0.019}  & \multicolumn{1}{l|}{0.000/0.010}  & \multicolumn{1}{l|}{0.000/0.005}  & \multicolumn{1}{l|}{0.000/0.014}  & \multicolumn{1}{l|}{0.005/0.010}           \\ \cline{3-13}                                                            \multicolumn{1}{|c|}{}                                                  &                                               & \textbf{Port}                                       &\multicolumn{1}{l|}{0.097/0.136}  & \multicolumn{1}{l|}{0.065/0.071}  & \multicolumn{1}{l|}{0.071/0.078}  & \multicolumn{1}{l|}{0.058/0.091}& 0.039/0.058   & \multicolumn{1}{l|}{0.053/0.062}  & \multicolumn{1}{l|}{0.019/0.019}  & \multicolumn{1}{l|}{0.010/0.024}  & \multicolumn{1}{l|}{0.019/0.019}  & \multicolumn{1}{l|}{0.019/0.034}       \\ \cline{3-13} 
% \multicolumn{1}{|c|}{}                                                  &                                               & \textbf{Port\_Alt}                                   & \multicolumn{1}{l|}{0.188}       & \multicolumn{1}{l|}{0.039}       & \multicolumn{1}{l|}{0.078}       & \multicolumn{1}{l|}{0.065}       & 0.084       & \multicolumn{1}{l|}{0.032}       & \multicolumn{1}{l|}{0.026}       & \multicolumn{1}{l|}{0.013}       & \multicolumn{1}{l|}{0.039}       & 0.019       & \multicolumn{1}{l|}{0.125}       & \multicolumn{1}{l|}{0.048}       & \multicolumn{1}{l|}{0.039}       & \multicolumn{1}{l|}{0.043}       & 0.063       \\ \cline{3-18} 
% \multicolumn{1}{|c|}{}                                                  &                                               & \textbf{Port\_Best}                                  & \multicolumn{1}{l|}{0.214}       & \multicolumn{1}{l|}{0.091}       & \multicolumn{1}{l|}{0.104}       & \multicolumn{1}{l|}{0.091}       & 0.104       & \multicolumn{1}{l|}{0.052}       & \multicolumn{1}{l|}{0.045}       & \multicolumn{1}{l|}{0.026}       & \multicolumn{1}{l|}{0.058}       & 0.032       & \multicolumn{1}{l|}{0.154}       & \multicolumn{1}{l|}{0.058}       & \multicolumn{1}{l|}{0.048}       & \multicolumn{1}{l|}{0.063}       & 0.077       \\ 
\cline{2-13} 
\multicolumn{1}{|c|}{}                                                  & \multirow{6}{*}{\textbf{De}}                  & \textbf{Rel}                                        & \multicolumn{1}{l|}{0.382/0.474}  & \multicolumn{1}{l|}{0.513/0.625}  & \multicolumn{1}{l|}{0.257/0.336}  & \multicolumn{1}{l|}{0.289/0.349}& 0.270/0.355   & \multicolumn{1}{l|}{0.157/0.216}  & \multicolumn{1}{l|}{0.480/0.593}  & \multicolumn{1}{l|}{0.221/0.260}  & \multicolumn{1}{l|}{0.211/0.240}  & \multicolumn{1}{l|}{0.176/0.211}       \\ \cline{3-13} 
\multicolumn{1}{|c|}{}                                                  &                                               & \textbf{Gen}                                         & \multicolumn{1}{l|}{0.342/0.428}  & \multicolumn{1}{l|}{0.349/0.454}  & \multicolumn{1}{l|}{0.237/0.309}  & \multicolumn{1}{l|}{0.237/0.289}& 0.217/0.289   & \multicolumn{1}{l|}{0.152/0.196}  & \multicolumn{1}{l|}{0.333/0.387}  & \multicolumn{1}{l|}{0.162/0.201}  & \multicolumn{1}{l|}{0.142/0.172}  & \multicolumn{1}{l|}{0.132/0.167}       \\ \cline{3-13} 
\multicolumn{1}{|c|}{}                                                  &                                               & \textbf{Loc}                                        & \multicolumn{1}{l|}{0.000/0.007}  & \multicolumn{1}{l|}{0.013/0.020}  & \multicolumn{1}{l|}{0.000/0.013}  & \multicolumn{1}{l|}{0.013/0.020}& 0.013/0.020   & \multicolumn{1}{l|}{0.010/0.020}  & \multicolumn{1}{l|}{0.000/0.010}  & \multicolumn{1}{l|}{0.000/0.005}  & \multicolumn{1}{l|}{0.000/0.015}  & \multicolumn{1}{l|}{0.005/0.010}       \\ \cline{3-13} 
\multicolumn{1}{|c|}{}                                                  &                                               & \textbf{Port}                                        & \multicolumn{1}{l|}{0.079/0.092}  & \multicolumn{1}{l|}{0.079/0.099}  & \multicolumn{1}{l|}{0.066/0.079}  & \multicolumn{1}{l|}{0.072/0.099}& 0.053/0.086   & \multicolumn{1}{l|}{0.010/0.020}  & \multicolumn{1}{l|}{0.025/0.025}  & \multicolumn{1}{l|}{0.010/0.015}  & \multicolumn{1}{l|}{0.020/0.020}  & \multicolumn{1}{l|}{0.010/0.015}      \\ \cline{3-13} 
% \multicolumn{1}{|c|}{}                                                  &                                               & \textbf{Port\_Alt}                                   & \multicolumn{1}{l|}{0.125}       & \multicolumn{1}{l|}{0.046}       & \multicolumn{1}{l|}{0.039}       & \multicolumn{1}{l|}{0.053}       & 0.072       & \multicolumn{1}{l|}{0.046}       & \multicolumn{1}{l|}{0.026}       & \multicolumn{1}{l|}{0.026}       & \multicolumn{1}{l|}{0.026}       & 0.020       & \multicolumn{1}{l|}{0.108}       & \multicolumn{1}{l|}{0.064}       & \multicolumn{1}{l|}{0.025}       & \multicolumn{1}{l|}{0.025}       & 0.050       \\ \cline{3-18} 
% \multicolumn{1}{|c|}{}                                                  &                                               & \textbf{Port\_Best}                                  & \multicolumn{1}{l|}{0.168}       & \multicolumn{1}{l|}{0.092}       & \multicolumn{1}{l|}{0.079}       & \multicolumn{1}{l|}{0.099}       & 0.099       & \multicolumn{1}{l|}{0.053}       & \multicolumn{1}{l|}{0.026}       & \multicolumn{1}{l|}{0.026}       & \multicolumn{1}{l|}{0.026}       & 0.026       & \multicolumn{1}{l|}{0.118}       & \multicolumn{1}{l|}{0.083}       & \multicolumn{1}{l|}{0.029}       & \multicolumn{1}{l|}{0.039}       & 0.058       \\ 
\cline{2-13} 
\multicolumn{1}{|c|}{}                                                  & \multirow{6}{*}{\textbf{It}}                  & \textbf{Rel}                                         & \multicolumn{1}{l|}{0.314/0.386}  & \multicolumn{1}{l|}{0.288/0.340}  & \multicolumn{1}{l|}{0.575/0.654}  & \multicolumn{1}{l|}{0.333/0.399}& 0.281/0.366    & \multicolumn{1}{l|}{0.176/0.224}  & \multicolumn{1}{l|}{0.185/0.215}  & \multicolumn{1}{l|}{0.537/0.624}  & \multicolumn{1}{l|}{0.210/0.268}  & \multicolumn{1}{l|}{0.229/0.340}       \\ \cline{3-13} 
\multicolumn{1}{|c|}{}                                                  &                                               & \textbf{Gen}                                          & \multicolumn{1}{l|}{0.340/0.405}  & \multicolumn{1}{l|}{0.242/0.281}  & \multicolumn{1}{l|}{0.418/0.484}  & \multicolumn{1}{l|}{0.294/0.373}& 0.222/0.327    & \multicolumn{1}{l|}{0.161/0.200}  & \multicolumn{1}{l|}{0.137/0.185}  & \multicolumn{1}{l|}{0.346/0.429}  & \multicolumn{1}{l|}{0.180/0.239}  & \multicolumn{1}{l|}{0.122/0.271}       \\ \cline{3-13} 
\multicolumn{1}{|c|}{}                                                  &                                               & \textbf{Loc}                                         & \multicolumn{1}{l|}{0.000/0.007}  & \multicolumn{1}{l|}{0.013/0.020}  & \multicolumn{1}{l|}{0.000/0.020}  & \multicolumn{1}{l|}{0.013/0.020}& 0.013/0.020   & \multicolumn{1}{l|}{0.010/0.015}  & \multicolumn{1}{l|}{0.000/0.010}  & \multicolumn{1}{l|}{0.000/0.005}  & \multicolumn{1}{l|}{0.000/0.015}  & \multicolumn{1}{l|}{0.005/0.005}    \\ \cline{3-13} 
\multicolumn{1}{|c|}{}                                                  &                                               & \textbf{Port}                                        & \multicolumn{1}{l|}{0.059/0.085}  & \multicolumn{1}{l|}{0.072/0.078}  & \multicolumn{1}{l|}{0.072/0.085}  & \multicolumn{1}{l|}{0.078/0.105}& 0.039/0.072   & \multicolumn{1}{l|}{0.029/0.029}  & \multicolumn{1}{l|}{0.029/0.029}  & \multicolumn{1}{l|}{0.015/0.020}  & \multicolumn{1}{l|}{0.029/0.034}  & \multicolumn{1}{l|}{0.020/0.030}       \\ \cline{3-13} 
% \multicolumn{1}{|c|}{}                                                  &                                               & \textbf{Port\_Alt}                                   & \multicolumn{1}{l|}{0.111}       & \multicolumn{1}{l|}{0.039}       & \multicolumn{1}{l|}{0.085}       & \multicolumn{1}{l|}{0.065}       & 0.072       & \multicolumn{1}{l|}{0.092}       & \multicolumn{1}{l|}{0.020}       & \multicolumn{1}{l|}{0.013}       & \multicolumn{1}{l|}{0.026}       & 0.033       & \multicolumn{1}{l|}{0.107}       & \multicolumn{1}{l|}{0.063}       & \multicolumn{1}{l|}{0.044}       & \multicolumn{1}{l|}{0.059}       & 0.059       \\ \cline{3-18} 
% \multicolumn{1}{|c|}{}                                                  &                                               & \textbf{Port\_Best}                                  & \multicolumn{1}{l|}{0.131}       & \multicolumn{1}{l|}{0.098}       & \multicolumn{1}{l|}{0.111}       & \multicolumn{1}{l|}{0.104}       & 0.098       & \multicolumn{1}{l|}{0.105}       & \multicolumn{1}{l|}{0.026}       & \multicolumn{1}{l|}{0.026}       & \multicolumn{1}{l|}{0.026}       & 0.039       & \multicolumn{1}{l|}{0.132}       & \multicolumn{1}{l|}{0.088}       & \multicolumn{1}{l|}{0.049}       & \multicolumn{1}{l|}{0.073}       & 0.068       \\ 
\cline{2-13} 
\multicolumn{1}{|c|}{}                                                  & \multirow{6}{*}{\textbf{Fr}}                  & \textbf{Rel}                                        & \multicolumn{1}{l|}{0.424/0.477}  & \multicolumn{1}{l|}{0.298/0.344}  & \multicolumn{1}{l|}{0.272/0.391}  & \multicolumn{1}{l|}{0.517/0.629}& 0.331/0.444   & \multicolumn{1}{l|}{0.143/0.177}  & \multicolumn{1}{l|}{0.153/0.187}  & \multicolumn{1}{l|}{0.197/0.256}  & \multicolumn{1}{l|}{0.507/0.591}  & \multicolumn{1}{l|}{0.138/0.167}       \\ \cline{3-13} 
\multicolumn{1}{|c|}{}                                                  &                                               & \textbf{Gen}                                         & \multicolumn{1}{l|}{0.371/0.424}  & \multicolumn{1}{l|}{0.285/0.325}  & \multicolumn{1}{l|}{0.245/0.325}  & \multicolumn{1}{l|}{0.404/0.503}& 0.245/0.351   & \multicolumn{1}{l|}{0.138/0.177}  & \multicolumn{1}{l|}{0.133/0.167}  & \multicolumn{1}{l|}{0.167/0.192}  & \multicolumn{1}{l|}{0.281/0.350}  & \multicolumn{1}{l|}{0.113/0.163}     \\ \cline{3-13} 
\multicolumn{1}{|c|}{}                                                  &                                               & \textbf{Loc}                                        & \multicolumn{1}{l|}{0.000/0.007}  & \multicolumn{1}{l|}{0.013/0.020}  & \multicolumn{1}{l|}{0.000/0.020}  & \multicolumn{1}{l|}{0.013/0.020}& 0.013/0.020   & \multicolumn{1}{l|}{0.010/0.020}  & \multicolumn{1}{l|}{0.000/0.010}  & \multicolumn{1}{l|}{0.000/0.005}  & \multicolumn{1}{l|}{0.005/0.015}  & \multicolumn{1}{l|}{0.005/0.010}       \\ \cline{3-13} 
\multicolumn{1}{|c|}{}                                                  &                                               & \textbf{Port}                                        & \multicolumn{1}{l|}{0.132/0.159}  & \multicolumn{1}{l|}{0.066/0.066}  & \multicolumn{1}{l|}{0.073/0.086}  & \multicolumn{1}{l|}{0.060/0.093}& 0.040/0.060   & \multicolumn{1}{l|}{0.015/0.025}  & \multicolumn{1}{l|}{0.025/0.025}  & \multicolumn{1}{l|}{0.010/0.020}  & \multicolumn{1}{l|}{0.034/0.054}  & \multicolumn{1}{l|}{0.005/0.020}        \\ \cline{3-13} 
% \multicolumn{1}{|c|}{}                                                  &                                               & \textbf{Port\_Alt}                                   & \multicolumn{1}{l|}{0.152}       & \multicolumn{1}{l|}{0.040}       & \multicolumn{1}{l|}{0.066}       & \multicolumn{1}{l|}{0.066}       & 0.073       & \multicolumn{1}{l|}{0.026}       & \multicolumn{1}{l|}{0.013}       & \multicolumn{1}{l|}{0.020}       & \multicolumn{1}{l|}{0.013}       & 0.013       & \multicolumn{1}{l|}{0.084}       & \multicolumn{1}{l|}{0.049}       & \multicolumn{1}{l|}{0.020}       & \multicolumn{1}{l|}{0.039}       & 0.025       \\ 
\cline{3-13} 
% \multicolumn{1}{|c|}{}                                                  &                                               & \textbf{Port\_Best}                                  & \multicolumn{1}{l|}{0.225}       & \multicolumn{1}{l|}{0.099}       & \multicolumn{1}{l|}{0.099}       & \multicolumn{1}{l|}{0.106}       & 0.093       & \multicolumn{1}{l|}{0.040}       & \multicolumn{1}{l|}{0.026}       & \multicolumn{1}{l|}{0.026}       & \multicolumn{1}{l|}{0.033}       & 0.026       & \multicolumn{1}{l|}{0.094}       & \multicolumn{1}{l|}{0.069}       & \multicolumn{1}{l|}{0.030}       & \multicolumn{1}{l|}{0.064}       & 0.030       \\ 
\cline{2-13} 
\multicolumn{1}{|c|}{}                                                  & \multirow{6}{*}{\textbf{Es}}                  & \textbf{Rel}                                        & \multicolumn{1}{l|}{0.367/0.440}  & \multicolumn{1}{l|}{0.260/0.320}  & \multicolumn{1}{l|}{0.360/0.433}  & \multicolumn{1}{l|}{0.307/0.400}& 0.487/0.607   & \multicolumn{1}{l|}{0.232/0.232}  & \multicolumn{1}{l|}{0.148/0.158}  & \multicolumn{1}{l|}{0.241/0.340}  & \multicolumn{1}{l|}{0.182/0.236}  & \multicolumn{1}{l|}{0.443/0.591}       \\ \cline{3-13} 
\multicolumn{1}{|c|}{}                                                  &                                               & \textbf{Gen}                                         & \multicolumn{1}{l|}{0.287/0.367}  & \multicolumn{1}{l|}{0.227/0.280}  & \multicolumn{1}{l|}{0.247/0.313}  & \multicolumn{1}{l|}{0.333/0.387}& 0.353/0.453  & \multicolumn{1}{l|}{0.153/0.177}  & \multicolumn{1}{l|}{0.094/0.118}  & \multicolumn{1}{l|}{0.202/0.271}  & \multicolumn{1}{l|}{0.182/0.241}  & \multicolumn{1}{l|}{0.305/0.404}\\ \cline{3-13} 
\multicolumn{1}{|c|}{}                                                  &                                               & \textbf{Loc}                                         & \multicolumn{1}{l|}{0.000/0.007}  & \multicolumn{1}{l|}{0.013/0.020}  & \multicolumn{1}{l|}{0.000/0.020}  & \multicolumn{1}{l|}{0.013/0.020}& 0.007/0.013   & \multicolumn{1}{l|}{0.010/0.010}  & \multicolumn{1}{l|}{0.000/0.005}  & \multicolumn{1}{l|}{0.000/0.005}  & \multicolumn{1}{l|}{0.000/0.010}  & \multicolumn{1}{l|}{0.005/0.010}       \\ \cline{3-13} 
\multicolumn{1}{|c|}{}                                                  &                                               & \textbf{Port}                                        & \multicolumn{1}{l|}{0.060/0.080}  & \multicolumn{1}{l|}{0.040/0.060}  & \multicolumn{1}{l|}{0.033/0.060}  & \multicolumn{1}{l|}{0.047/0.080}& 0.033/0.067    & \multicolumn{1}{l|}{0.000/0.000}  & \multicolumn{1}{l|}{0.010/0.010}  & \multicolumn{1}{l|}{0.015/0.030}  & \multicolumn{1}{l|}{0.010/0.020}  & \multicolumn{1}{l|}{0.020/0.020}      \\ \cline{3-13} 
% \multicolumn{1}{|c|}{}                                                  &                                               & \textbf{Port\_Alt}                                   & \multicolumn{1}{l|}{0.113}       & \multicolumn{1}{l|}{0.033}       & \multicolumn{1}{l|}{0.060}       & \multicolumn{1}{l|}{0.053}       & 0.067       & \multicolumn{1}{l|}{0.027}       & \multicolumn{1}{l|}{0.013}       & \multicolumn{1}{l|}{0.007}       & \multicolumn{1}{l|}{0.007}       & 0.000       & \multicolumn{1}{l|}{0.103}       & \multicolumn{1}{l|}{0.039}       & \multicolumn{1}{l|}{0.034}       & \multicolumn{1}{l|}{0.034}       & 0.050       \\ 
\cline{1-13} 
% \multicolumn{1}{|c|}{}                                                  &                                               & \textbf{Port\_Best}                                  & \multicolumn{1}{l|}{0.127}       & \multicolumn{1}{l|}{0.060}       & \multicolumn{1}{l|}{0.080}       & \multicolumn{1}{l|}{0.073}       & 0.080       & \multicolumn{1}{l|}{0.047}       & \multicolumn{1}{l|}{0.020}       & \multicolumn{1}{l|}{0.007}       & \multicolumn{1}{l|}{0.020}       & 0.000       & \multicolumn{1}{l|}{0.103}       & \multicolumn{1}{l|}{0.049}       & \multicolumn{1}{l|}{0.049}       & \multicolumn{1}{l|}{0.034}       & 0.060       \\ \hline
\end{tabular}
}
\caption{Comparison of reliability (Rel), generalization (Gen), locality (Loc), and portability (Port) scores for multiple language models evaluated using the ZsRE dataset and the ROME editing method. The second column indicates the language in which each model was edited.}
\label{tab:rome_zsre}
\end{table*}

\begin{table*}[h]
\centering
\resizebox{1.0\textwidth}{!}{
\begin{tabular}{|cc|l|lllll|lllll|}
\hline
\multicolumn{2}{|c|}{\multirow{2}{*}{\textbf{\begin{tabular}[c]{@{}c@{}}Datasets/\\ Languages\end{tabular}}}} & \multicolumn{1}{c|}{\multirow{2}{*}{\textbf{Score}}} & \multicolumn{5}{c|}{\textbf{Mistral}}                                                                                                                                        & \multicolumn{5}{c|}{\textbf{TowerInstruct}}                                                                                                                                  \\ \cline{4-13} 
\multicolumn{2}{|c|}{}                                                                                                  & \multicolumn{1}{c|}{}                                & \multicolumn{1}{c|}{\textbf{En}} & \multicolumn{1}{c|}{\textbf{De}} & \multicolumn{1}{c|}{\textbf{It}} & \multicolumn{1}{c|}{\textbf{Fr}} & \multicolumn{1}{c|}{\textbf{Es}} & \multicolumn{1}{c|}{\textbf{En}} & \multicolumn{1}{c|}{\textbf{De}} & \multicolumn{1}{c|}{\textbf{It}} & \multicolumn{1}{c|}{\textbf{Fr}} & \multicolumn{1}{c|}{\textbf{Es}} \\ \hline
\multicolumn{1}{|c|}{\multirow{30}{*}{\textbf{CounterFact}}}               & \multirow{6}{*}{\textbf{En}}               & \textbf{Rel}                                        & \multicolumn{1}{l|}{0.988/0.988}  & \multicolumn{1}{l|}{0.537/0.606}  & \multicolumn{1}{l|}{0.438/0.494}  & \multicolumn{1}{l|}{0.506/0.588}& 0.562/0.600  & \multicolumn{1}{l|}{0.954/0.963}  & \multicolumn{1}{l|}{0.404/0.459}  & \multicolumn{1}{l|}{0.349/0.404}  & \multicolumn{1}{l|}{0.450/0.486}& 0.404/0.477                            \\ \cline{3-13} 
\multicolumn{1}{|c|}{}                                                     &                                            & \textbf{Gen}                                         & \multicolumn{1}{l|}{0.444/0.456}  & \multicolumn{1}{l|}{0.219/0.225}  & \multicolumn{1}{l|}{0.212/0.225}  & \multicolumn{1}{l|}{0.263/0.269}& 0.212/0.263  & \multicolumn{1}{l|}{0.431/0.431}  & \multicolumn{1}{l|}{0.128/0.174}  & \multicolumn{1}{l|}{0.193/0.202}  & \multicolumn{1}{l|}{0.193/0.220}& 0.183/0.220                           \\ \cline{3-13} 
\multicolumn{1}{|c|}{}                                                     &                                            & \textbf{Loc}                                         & \multicolumn{1}{l|}{0.381/0.388}  & \multicolumn{1}{l|}{0.256/0.281}  & \multicolumn{1}{l|}{0.275/0.287}  & \multicolumn{1}{l|}{0.250/0.263}& 0.250/0.269  & \multicolumn{1}{l|}{0.275/0.294}  & \multicolumn{1}{l|}{0.193/0.220}  & \multicolumn{1}{l|}{0.202/0.202}  & \multicolumn{1}{l|}{0.193/0.211}& 0.165/0.165                            \\ \cline{3-13} 
\multicolumn{1}{|c|}{}                                                     &                                            & \textbf{Port}                                        & \multicolumn{1}{l|}{0.156/0.188}  & \multicolumn{1}{l|}{0.025/0.037}  & \multicolumn{1}{l|}{0.037/0.037}  & \multicolumn{1}{l|}{0.031/0.037}& 0.025/0.037  & \multicolumn{1}{l|}{0.000/0.000}  & \multicolumn{1}{l|}{0.000/0.000}  & \multicolumn{1}{l|}{0.009/0.018}  & \multicolumn{1}{l|}{0.009/0.009}& 0.000/0.000                     \\ \cline{3-13} 
% \multicolumn{1}{|c|}{}                                                     &                                            & \textbf{Port\_Alt}                                   & \multicolumn{1}{l|}{0.237}       & \multicolumn{1}{l|}{0.156}       & \multicolumn{1}{l|}{0.112}       & \multicolumn{1}{l|}{0.118}       & 0.162                            & \multicolumn{1}{l|}{0.055}       & \multicolumn{1}{l|}{0.119}       & \multicolumn{1}{l|}{0.073}       & \multicolumn{1}{l|}{0.137}       & 0.091                            \\ 
\cline{3-13} 
% \multicolumn{1}{|c|}{}                                                     &                                            & \textbf{Port\_Best}                                  & \multicolumn{1}{l|}{0.281}       & \multicolumn{1}{l|}{0.175}       & \multicolumn{1}{l|}{0.131}       & \multicolumn{1}{l|}{0.131}       & 0.181                            & \multicolumn{1}{l|}{0.055}       & \multicolumn{1}{l|}{0.119}       & \multicolumn{1}{l|}{0.073}       & \multicolumn{1}{l|}{0.137}       & 0.091                            \\ 
\cline{2-13} 
\multicolumn{1}{|c|}{}                                                     & \multirow{6}{*}{\textbf{De}}               & \textbf{Rel}                                         & \multicolumn{1}{l|}{0.439/0.484}  & \multicolumn{1}{l|}{0.726/0.866}  & \multicolumn{1}{l|}{0.376/0.420}  & \multicolumn{1}{l|}{0.350/0.369}& 0.363/0.414  & \multicolumn{1}{l|}{0.355/0.391}  & \multicolumn{1}{l|}{0.727/0.827}  & \multicolumn{1}{l|}{0.282/0.380}  & \multicolumn{1}{l|}{0.309/0.309}& 0.255/0.300                           \\ \cline{3-13} 
\multicolumn{1}{|c|}{}                                                     &                                            & \textbf{Gen}                                         & \multicolumn{1}{l|}{0.242/0.280}  & \multicolumn{1}{l|}{0.191/0.223}  & \multicolumn{1}{l|}{0.185/0.191}  & \multicolumn{1}{l|}{0.185/0.217}& 0.178/0.210  & \multicolumn{1}{l|}{0.227/0.236}  & \multicolumn{1}{l|}{0.191/0.218}  & \multicolumn{1}{l|}{0.136/0.176}  & \multicolumn{1}{l|}{0.182/0.209}& 0.145/0.164                         \\ \cline{3-13} 
\multicolumn{1}{|c|}{}                                                     &                                            & \textbf{Loc}                                         & \multicolumn{1}{l|}{0.376/0.389}  & \multicolumn{1}{l|}{0.242/0.268}  & \multicolumn{1}{l|}{0.280/0.293}  & \multicolumn{1}{l|}{0.229/0.242}& 0.274/0.280  & \multicolumn{1}{l|}{0.264/0.282}  & \multicolumn{1}{l|}{0.191/0.218}  & \multicolumn{1}{l|}{0.200/0.231}  & \multicolumn{1}{l|}{0.209/0.227}& 0.200/0.200                            \\ \cline{3-13} 
\multicolumn{1}{|c|}{}                                                     &                                            & \textbf{Port}                                       & \multicolumn{1}{l|}{0.108/0.134}  & \multicolumn{1}{l|}{0.045/0.064}  & \multicolumn{1}{l|}{0.025/0.025}  & \multicolumn{1}{l|}{0.013/0.025}& 0.032/0.051  & \multicolumn{1}{l|}{0.000/0.000}  & \multicolumn{1}{l|}{0.009/0.009}  & \multicolumn{1}{l|}{0.009/0.009}  & \multicolumn{1}{l|}{0.009/0.009}& 0.000/0.000                           \\ \cline{3-13} 
% \multicolumn{1}{|c|}{}                                                     &                                            & \textbf{Port\_Alt}                                   & \multicolumn{1}{l|}{0.165}       & \multicolumn{1}{l|}{0.140}       & \multicolumn{1}{l|}{0.095}       & \multicolumn{1}{l|}{0.114}       & 0.140                            & \multicolumn{1}{l|}{0.090}       & \multicolumn{1}{l|}{0.145}       & \multicolumn{1}{l|}{0.090}       & \multicolumn{1}{l|}{0.127}       & 0.118                            \\ 
\cline{3-13} 
% \multicolumn{1}{|c|}{}                                                     &                                            & \textbf{Port\_Best}                                  & \multicolumn{1}{l|}{0.203}       & \multicolumn{1}{l|}{0.159}       & \multicolumn{1}{l|}{0.101}       & \multicolumn{1}{l|}{0.114}       & 0.165                            & \multicolumn{1}{l|}{0.09}        & \multicolumn{1}{l|}{0.154}       & \multicolumn{1}{l|}{0.090}       & \multicolumn{1}{l|}{0.127}       & 0.118                            \\ 
\cline{2-13} 
\multicolumn{1}{|c|}{}                                                     & \multirow{6}{*}{\textbf{It}}               & \textbf{Rel}                                           & \multicolumn{1}{l|}{0.372/0.404}  & \multicolumn{1}{l|}{0.353/0.410}  & \multicolumn{1}{l|}{0.801/0.878}  & \multicolumn{1}{l|}{0.455/0.526}& 0.449/0.526  & \multicolumn{1}{l|}{0.407/0.444}  & \multicolumn{1}{l|}{0.361/0.380}  & \multicolumn{1}{l|}{0.741/0.778}  & \multicolumn{1}{l|}{0.389/0.417}& 0.426/0.454                           \\ \cline{3-13} 
\multicolumn{1}{|c|}{}                                                     &                                            & \textbf{Gen}                                        & \multicolumn{1}{l|}{0.256/0.263}  & \multicolumn{1}{l|}{0.141/0.167}  & \multicolumn{1}{l|}{0.237/0.269}  & \multicolumn{1}{l|}{0.192/0.231}& 0.179/0.212  & \multicolumn{1}{l|}{0.315/0.315}  & \multicolumn{1}{l|}{0.139/0.176}  & \multicolumn{1}{l|}{0.250/0.259}  & \multicolumn{1}{l|}{0.204/0.213}& 0.185/0.213                            \\ \cline{3-13} 
\multicolumn{1}{|c|}{}                                                     &                                            & \textbf{Loc}                                         & \multicolumn{1}{l|}{0.385/0.397}  & \multicolumn{1}{l|}{0.263/0.288}  & \multicolumn{1}{l|}{0.269/0.282}  & \multicolumn{1}{l|}{0.250/0.263}& 0.276/0.282  & \multicolumn{1}{l|}{0.269/0.287}  & \multicolumn{1}{l|}{0.204/0.231}  & \multicolumn{1}{l|}{0.204/0.204}  & \multicolumn{1}{l|}{0.194/0.213}& 0.176/0.176                            \\ \cline{3-13} 
\multicolumn{1}{|c|}{}                                                     &                                            & \textbf{Port}                                        & \multicolumn{1}{l|}{0.122/0.147}  & \multicolumn{1}{l|}{0.013/0.032}  & \multicolumn{1}{l|}{0.026/0.026}  & \multicolumn{1}{l|}{0.013/0.019}& 0.019/0.045  & \multicolumn{1}{l|}{0.009/0.009}  & \multicolumn{1}{l|}{0.009/0.009}  & \multicolumn{1}{l|}{0.019/0.028}  & \multicolumn{1}{l|}{0.009/0.009}& 0.000/0.000                           \\ \cline{3-13} 
% \multicolumn{1}{|c|}{}                                                     &                                            & \textbf{Port\_Alt}                                   & \multicolumn{1}{l|}{0.173}       & \multicolumn{1}{l|}{0.089}       & \multicolumn{1}{l|}{0.076}       & \multicolumn{1}{l|}{0.083}       & 0.121                            & \multicolumn{1}{l|}{0.090}       & \multicolumn{1}{l|}{0.145}       & \multicolumn{1}{l|}{0.090}       & \multicolumn{1}{l|}{0.127}       & 0.118                            \\ 
\cline{3-13} 
% \multicolumn{1}{|c|}{}                                                     &                                            & \textbf{Port\_Best}                                  & \multicolumn{1}{l|}{0.217}       & \multicolumn{1}{l|}{0.096}       & \multicolumn{1}{l|}{0.096}       & \multicolumn{1}{l|}{0.089}       & 0.134                            & \multicolumn{1}{l|}{0.090}       & \multicolumn{1}{l|}{0.154}       & \multicolumn{1}{l|}{0.090}       & \multicolumn{1}{l|}{0.127}       & 0.118                            \\ 
\cline{2-13} 
\multicolumn{1}{|c|}{}                                                     & \multirow{6}{*}{\textbf{Fr}}               & \textbf{Rel}                                       & \multicolumn{1}{l|}{0.439/0.459}  & \multicolumn{1}{l|}{0.395/0.471}  & \multicolumn{1}{l|}{0.401/0.433}  & \multicolumn{1}{l|}{0.790/0.847}& 0.446/0.478  & \multicolumn{1}{l|}{0.468/0.477}  & \multicolumn{1}{l|}{0.330/0.385}  & \multicolumn{1}{l|}{0.330/0.376}  & \multicolumn{1}{l|}{0.651/0.716}& 0.330/0.367                            \\ \cline{3-13} 
\multicolumn{1}{|c|}{}                                                     &                                            & \textbf{Gen}                                        & \multicolumn{1}{l|}{0.229/0.268}  & \multicolumn{1}{l|}{0.153/0.166}  & \multicolumn{1}{l|}{0.159/0.172}  & \multicolumn{1}{l|}{0.236/0.255}& 0.153/0.172  & \multicolumn{1}{l|}{0.294/0.312}  & \multicolumn{1}{l|}{0.128/0.147}  & \multicolumn{1}{l|}{0.183/0.183}  & \multicolumn{1}{l|}{0.220/0.239}& 0.174/0.193                            \\ \cline{3-13} 
\multicolumn{1}{|c|}{}                                                     &                                            & \textbf{Loc}                                         & \multicolumn{1}{l|}{0.389/0.401}  & \multicolumn{1}{l|}{0.268/0.293}  & \multicolumn{1}{l|}{0.280/0.293}  & \multicolumn{1}{l|}{0.242/0.255}& 0.274/0.280  & \multicolumn{1}{l|}{0.248/0.266}  & \multicolumn{1}{l|}{0.183/0.211}  & \multicolumn{1}{l|}{0.183/0.183}  & \multicolumn{1}{l|}{0.174/0.193}& 0.174/0.174                           \\ \cline{3-13} 
\multicolumn{1}{|c|}{}                                                     &                                            & \textbf{Port}                                        & \multicolumn{1}{l|}{0.089/0.115}  & \multicolumn{1}{l|}{0.019/0.032}  & \multicolumn{1}{l|}{0.019/0.019}  & \multicolumn{1}{l|}{0.025/0.032}& 0.013/0.025  & \multicolumn{1}{l|}{0.000/0.000}  & \multicolumn{1}{l|}{0.009/0.009}  & \multicolumn{1}{l|}{0.009/0.018}  & \multicolumn{1}{l|}{0.000/0.000}& 0.000/0.000                           \\ \cline{3-13} 
% \multicolumn{1}{|c|}{}                                                     &                                            & \textbf{Port\_Alt}                                   & \multicolumn{1}{l|}{0.159}       & \multicolumn{1}{l|}{0.127}       & \multicolumn{1}{l|}{0.089}       & \multicolumn{1}{l|}{0.108}       & 0.133                            & \multicolumn{1}{l|}{0.036}       & \multicolumn{1}{l|}{0.091}       & \multicolumn{1}{l|}{0.064}       & \multicolumn{1}{l|}{0.110}       & 0.110                            \\ 
\cline{3-13} 
% \multicolumn{1}{|c|}{}                                                     &                                            & \textbf{Port\_Best}                                  & \multicolumn{1}{l|}{0.203}       & \multicolumn{1}{l|}{0.146}       & \multicolumn{1}{l|}{0.108}       & \multicolumn{1}{l|}{0.127}       & 0.146                            & \multicolumn{1}{l|}{0.036}       & \multicolumn{1}{l|}{0.100}       & \multicolumn{1}{l|}{0.073}       & \multicolumn{1}{l|}{0.110}       & 0.110                            \\ 
\cline{2-13} 
\multicolumn{1}{|c|}{}                                                     & \multirow{6}{*}{\textbf{Es}}               & \textbf{Rel}                                       & \multicolumn{1}{l|}{0.433/0.465}  & \multicolumn{1}{l|}{0.338/0.382}  & \multicolumn{1}{l|}{0.401/0.452}  & \multicolumn{1}{l|}{0.471/0.522}& 0.777/0.860  & \multicolumn{1}{l|}{0.435/0.463}  & \multicolumn{1}{l|}{0.306/0.324}  & \multicolumn{1}{l|}{0.370/0.398}  & \multicolumn{1}{l|}{0.380/0.398}& 0.704/0.796                          \\ \cline{3-13} 
\multicolumn{1}{|c|}{}                                                     &                                            & \textbf{Gen}                                         & \multicolumn{1}{l|}{0.210/0.229}  & \multicolumn{1}{l|}{0.127/0.159}  & \multicolumn{1}{l|}{0.121/0.134}  & \multicolumn{1}{l|}{0.185/0.217}& 0.223/0.274  & \multicolumn{1}{l|}{0.241/0.250}  & \multicolumn{1}{l|}{0.148/0.157}  & \multicolumn{1}{l|}{0.194/0.204}  & \multicolumn{1}{l|}{0.213/0.213}& 0.231/0.269                          \\ \cline{3-13} 
\multicolumn{1}{|c|}{}                                                     &                                            & \textbf{Loc}                                         & \multicolumn{1}{l|}{0.395/0.408}  & \multicolumn{1}{l|}{0.274/0.306}  & \multicolumn{1}{l|}{0.268/0.287}  & \multicolumn{1}{l|}{0.242/0.255}& 0.274/0.287  & \multicolumn{1}{l|}{0.259/0.278}  & \multicolumn{1}{l|}{0.194/0.222}  & \multicolumn{1}{l|}{0.185/0.185}  & \multicolumn{1}{l|}{0.176/0.194}& 0.185/0.185                            \\ \cline{3-13} 
\multicolumn{1}{|c|}{}                                                     &                                            & \textbf{Port}                                        & \multicolumn{1}{l|}{0.108/0.134}  & \multicolumn{1}{l|}{0.025/0.051}  & \multicolumn{1}{l|}{0.006/0.006}  & \multicolumn{1}{l|}{0.013/0.013}& 0.025/0.045  & \multicolumn{1}{l|}{0.009/0.009}  & \multicolumn{1}{l|}{0.000/0.009}  & \multicolumn{1}{l|}{0.009/0.019}  & \multicolumn{1}{l|}{0.019/0.019}& 0.000/0.000                           \\ \cline{3-13} 
% \multicolumn{1}{|c|}{}                                                     &                                            & \textbf{Port\_Alt}                                   & \multicolumn{1}{l|}{0.146}       & \multicolumn{1}{l|}{0.089}       & \multicolumn{1}{l|}{0.095}       & \multicolumn{1}{l|}{0.082}       & 0.165                            & \multicolumn{1}{l|}{0.055}       & \multicolumn{1}{l|}{0.129}       & \multicolumn{1}{l|}{0.046}       & \multicolumn{1}{l|}{0.092}       & 0.111                            \\ 
\cline{3-13} 
% \multicolumn{1}{|c|}{}                                                     &                                            & \textbf{Port\_Best}                                  & \multicolumn{1}{l|}{0.191}       & \multicolumn{1}{l|}{0.101}       & \multicolumn{1}{l|}{0.095}       & \multicolumn{1}{l|}{0.089}       & 0.178                            & \multicolumn{1}{l|}{0.064}       & \multicolumn{1}{l|}{0.129}       & \multicolumn{1}{l|}{0.055}       & \multicolumn{1}{l|}{0.101}       & 0.111                            \\ 
\hline
\end{tabular}
}
\caption{Comparison of reliability (Rel), generalization (Gen), locality (Loc), and portability (Port) scores for multiple language models evaluated using the CounterFact dataset and the MEMIT editing method. The second column indicates the language in which each model was edited.}
\label{tab:counter_MEMIT}
\end{table*}

% \usepackage{multirow}

\begin{table*}[h]
\centering
\resizebox{1.00\textwidth}{!}{
\begin{tabular}{|cc|l|lllll|lllll|}
\hline
\multicolumn{2}{|c|}{\multirow{2}{*}{\textbf{\begin{tabular}[c]{@{}c@{}}Datasets/\\ Languages\end{tabular}}}} & \multicolumn{1}{c|}{\multirow{2}{*}{\textbf{Score}}} & \multicolumn{5}{c|}{\textbf{Mistral}}                                                                                                                                        & \multicolumn{5}{c|}{\textbf{TowerInstruct}}                                                                                                                                  \\ \cline{4-13} 
\multicolumn{2}{|c|}{}                                                                                                  & \multicolumn{1}{c|}{}                                & \multicolumn{1}{c|}{\textbf{En}} & \multicolumn{1}{c|}{\textbf{De}} & \multicolumn{1}{c|}{\textbf{It}} & \multicolumn{1}{c|}{\textbf{Fr}} & \multicolumn{1}{c|}{\textbf{Es}} & \multicolumn{1}{c|}{\textbf{En}} & \multicolumn{1}{c|}{\textbf{De}} & \multicolumn{1}{c|}{\textbf{It}} & \multicolumn{1}{c|}{\textbf{Fr}} & \multicolumn{1}{c|}{\textbf{Es}} \\ \hline
\multicolumn{1}{|c|}{\multirow{30}{*}{\textbf{ZSRE}}}                   & \multirow{6}{*}{\textbf{En}}                  & \textbf{Rel}                                        & \multicolumn{1}{l|}{0.786/0.812}  & \multicolumn{1}{l|}{0.136/0.182}  & \multicolumn{1}{l|}{0.227/0.266}  & \multicolumn{1}{l|}{0.227/0.279}& 0.188/0.260  & \multicolumn{1}{l|}{0.528/0.538}  & \multicolumn{1}{l|}{0.104/0.142}  & \multicolumn{1}{l|}{0.123/0.142}  & \multicolumn{1}{l|}{0.123/0.170}& 0.123/0.142                         \\ \cline{3-13} 
\multicolumn{1}{|c|}{}                                                  &                                               & \textbf{Gen}                                        & \multicolumn{1}{l|}{0.513/0.545}  & \multicolumn{1}{l|}{0.136/0.162}  & \multicolumn{1}{l|}{0.175/0.208}  & \multicolumn{1}{l|}{0.156/0.208}& 0.136/0.208  & \multicolumn{1}{l|}{0.321/0.330}  & \multicolumn{1}{l|}{0.123/0.142}  & \multicolumn{1}{l|}{0.113/0.132}  & \multicolumn{1}{l|}{0.075/0.104}& 0.094/0.113                         \\ \cline{3-13} 
\multicolumn{1}{|c|}{}                                                  &                                               & \textbf{Loc}                                        & \multicolumn{1}{l|}{0.019/0.026}  & \multicolumn{1}{l|}{0.013/0.032}  & \multicolumn{1}{l|}{0.013/0.019}  & \multicolumn{1}{l|}{0.013/0.019}& 0.019/0.019  & \multicolumn{1}{l|}{0.019/0.038}  & \multicolumn{1}{l|}{0.000/0.019}  & \multicolumn{1}{l|}{0.000/0.009}  & \multicolumn{1}{l|}{0.000/0.019}& 0.009/0.019                      \\ \cline{3-13} 
\multicolumn{1}{|c|}{}                                                  &                                               & \textbf{Port}                                       & \multicolumn{1}{l|}{0.039/0.065}  & \multicolumn{1}{l|}{0.019/0.032}  & \multicolumn{1}{l|}{0.006/0.006}  & \multicolumn{1}{l|}{0.039/0.052}& 0.039/0.045  & \multicolumn{1}{l|}{0.019/0.028}  & \multicolumn{1}{l|}{0.019/0.019}  & \multicolumn{1}{l|}{0.019/0.019}  & \multicolumn{1}{l|}{0.019/0.019}& 0.009/0.009                 \\ \cline{3-13}

% \multicolumn{1}{|c|}{}                                                  &                                               & \textbf{Port\_Alt}                                   & \multicolumn{1}{l|}{0.155}       & \multicolumn{1}{l|}{0.038}       & \multicolumn{1}{l|}{0.064}       & \multicolumn{1}{l|}{0.051}       & 0.136                            & \multicolumn{1}{l|}{0.169}       & \multicolumn{1}{l|}{0.047}       & \multicolumn{1}{l|}{0.056}       & \multicolumn{1}{l|}{0.066}       & 0.047                            \\ 
\cline{3-13} 
% \multicolumn{1}{|c|}{}                                                  &                                               & \textbf{Port\_Best}                                  & \multicolumn{1}{l|}{0.181}       & \multicolumn{1}{l|}{0.058}       & \multicolumn{1}{l|}{0.071}       & \multicolumn{1}{l|}{0.071}       & 0.162                            & \multicolumn{1}{l|}{0.188}       & \multicolumn{1}{l|}{0.066}       & \multicolumn{1}{l|}{0.075}       & \multicolumn{1}{l|}{0.075}       & 0.056                            \\ 
\cline{2-13} 
\multicolumn{1}{|c|}{}                                                  & \multirow{6}{*}{\textbf{De}}                  & \textbf{Rel}                                        & \multicolumn{1}{l|}{0.158/0.204}  & \multicolumn{1}{l|}{0.382/0.474}  & \multicolumn{1}{l|}{0.138/0.178}  & \multicolumn{1}{l|}{0.112/0.132}& 0.118/0.164  & \multicolumn{1}{l|}{0.029/0.077}  & \multicolumn{1}{l|}{0.250/0.298}  & \multicolumn{1}{l|}{0.048/0.067}  & \multicolumn{1}{l|}{0.038/0.058}& 0.048/0.048                          \\ \cline{3-13} 
\multicolumn{1}{|c|}{}                                                  &                                               & \textbf{Gen}                                        
  & \multicolumn{1}{l|}{0.125/0.171}  & \multicolumn{1}{l|}{0.184/0.243}  & \multicolumn{1}{l|}{0.138/0.164}  & \multicolumn{1}{l|}{0.105/0.118}& 0.086/0.125  & \multicolumn{1}{l|}{0.058/0.067}  & \multicolumn{1}{l|}{0.106/0.115}  & \multicolumn{1}{l|}{0.048/0.067}  & \multicolumn{1}{l|}{0.038/0.048}& 0.038/0.058                   \\ \cline{3-13} 
\multicolumn{1}{|c|}{}                                                  &                                               & \textbf{Loc}                                        & \multicolumn{1}{l|}{0.020/0.026}  & \multicolumn{1}{l|}{0.007/0.026}  & \multicolumn{1}{l|}{0.013/0.020}  & \multicolumn{1}{l|}{0.013/0.020}& 0.020/0.020  & \multicolumn{1}{l|}{0.019/0.029}  & \multicolumn{1}{l|}{0.000/0.010}  & \multicolumn{1}{l|}{0.000/0.010}  & \multicolumn{1}{l|}{0.000/0.019}& 0.010/0.019                            \\ \cline{3-13} 
\multicolumn{1}{|c|}{}                                                  &                                               & \textbf{Port}                                       & \multicolumn{1}{l|}{0.039/0.066}  & \multicolumn{1}{l|}{0.020/0.039}  & \multicolumn{1}{l|}{0.013/0.013}  & \multicolumn{1}{l|}{0.007/0.020}& 0.020/0.033  & \multicolumn{1}{l|}{0.010/0.019}  & \multicolumn{1}{l|}{0.000/0.000}  & \multicolumn{1}{l|}{0.000/0.000}  & \multicolumn{1}{l|}{0.010/0.010}& 0.000/0.000                           \\ \cline{3-13} 
% \multicolumn{1}{|c|}{}                                                  &                                               & \textbf{Port\_Alt}                                   & \multicolumn{1}{l|}{0.125}       & \multicolumn{1}{l|}{0.059}       & \multicolumn{1}{l|}{0.046}       & \multicolumn{1}{l|}{0.052}       & 0.092                            & \multicolumn{1}{l|}{0.096}       & \multicolumn{1}{l|}{0.038}       & \multicolumn{1}{l|}{0.038}       & \multicolumn{1}{l|}{0.019}       & 0.038                            \\ 
\cline{3-13} 
% \multicolumn{1}{|c|}{}                                                  &                                               & \textbf{Port\_Best}                                  & \multicolumn{1}{l|}{0.144}       & \multicolumn{1}{l|}{0.078}       & \multicolumn{1}{l|}{0.059}       & \multicolumn{1}{l|}{0.059}       & 0.098                            & \multicolumn{1}{l|}{0.105}       & \multicolumn{1}{l|}{0.038}       & \multicolumn{1}{l|}{0.038}       & \multicolumn{1}{l|}{0.028}       & 0.038                            \\ 
\cline{2-13} 
\multicolumn{1}{|c|}{}                                                  & \multirow{6}{*}{\textbf{It}}                  & \textbf{Rel}                                         & \multicolumn{1}{l|}{0.144/0.176}  & \multicolumn{1}{l|}{0.157/0.196}  & \multicolumn{1}{l|}{0.425/0.503}  & \multicolumn{1}{l|}{0.144/0.183}& 0.163/0.216  & \multicolumn{1}{l|}{0.019/0.038}  & \multicolumn{1}{l|}{0.038/0.067}  & \multicolumn{1}{l|}{0.248/0.286}  & \multicolumn{1}{l|}{0.067/0.086}& 0.095/0.124                            \\ \cline{3-13} 
\multicolumn{1}{|c|}{}                                                  &                                               & \textbf{Gen}                                         & \multicolumn{1}{l|}{0.105/0.150}  & \multicolumn{1}{l|}{0.085/0.118}  & \multicolumn{1}{l|}{0.255/0.307}  & \multicolumn{1}{l|}{0.144/0.183}& 0.105/0.157  & \multicolumn{1}{l|}{0.029/0.067}  & \multicolumn{1}{l|}{0.048/0.076}  & \multicolumn{1}{l|}{0.162/0.200}  & \multicolumn{1}{l|}{0.038/0.057}& 0.048/0.067                           \\ \cline{3-13} 
\multicolumn{1}{|c|}{}                                                  &                                               & \textbf{Loc}                                         & \multicolumn{1}{l|}{0.020/0.026}  & \multicolumn{1}{l|}{0.007/0.026}  & \multicolumn{1}{l|}{0.013/0.020}  & \multicolumn{1}{l|}{0.013/0.020}& 0.020/0.020  & \multicolumn{1}{l|}{0.019/0.029}  & \multicolumn{1}{l|}{0.000/0.019}  & \multicolumn{1}{l|}{0.000/0.010}  & \multicolumn{1}{l|}{0.000/0.029}& 0.010/0.019                            \\ \cline{3-13} 
\multicolumn{1}{|c|}{}                                                  &                                               & \textbf{Port}                                        & \multicolumn{1}{l|}{0.046/0.072}  & \multicolumn{1}{l|}{0.007/0.033}  & \multicolumn{1}{l|}{0.013/0.033}  & \multicolumn{1}{l|}{0.020/0.033}& 0.020/0.033  & \multicolumn{1}{l|}{0.000/0.010}  & \multicolumn{1}{l|}{0.010/0.019}  & \multicolumn{1}{l|}{0.019/0.029}  & \multicolumn{1}{l|}{0.010/0.010}& 0.000/0.000                         \\ \cline{3-13} 
% \multicolumn{1}{|c|}{}                                                  &                                               & \textbf{Port\_Alt}                                   & \multicolumn{1}{l|}{0.130}       & \multicolumn{1}{l|}{0.039}       & \multicolumn{1}{l|}{0.045}       & \multicolumn{1}{l|}{0.039}       & 0.091                            & \multicolumn{1}{l|}{0.114}       & \multicolumn{1}{l|}{0.057}       & \multicolumn{1}{l|}{0.047}       & \multicolumn{1}{l|}{0.038}       & 0.057                            \\ 
\cline{3-13} 
% \multicolumn{1}{|c|}{}                                                  &                                               & \textbf{Port\_Best}                                  & \multicolumn{1}{l|}{0.163}       & \multicolumn{1}{l|}{0.045}       & \multicolumn{1}{l|}{0.052}       & \multicolumn{1}{l|}{0.052}       & 0.098                            & \multicolumn{1}{l|}{0.114}       & \multicolumn{1}{l|}{0.066}       & \multicolumn{1}{l|}{0.066}       & \multicolumn{1}{l|}{0.038}       & 0.057                            \\ 
\cline{2-13} 
\multicolumn{1}{|c|}{}                                                  & \multirow{6}{*}{\textbf{Fr}}                  & \textbf{Rel}                                       & \multicolumn{1}{l|}{0.139/0.172}  & \multicolumn{1}{l|}{0.099/0.152}  & \multicolumn{1}{l|}{0.166/0.238}  & \multicolumn{1}{l|}{0.397/0.497}& 0.119/0.166  & \multicolumn{1}{l|}{0.048/0.077}  & \multicolumn{1}{l|}{0.048/0.067}  & \multicolumn{1}{l|}{0.038/0.077}  & \multicolumn{1}{l|}{0.269/0.346}& 0.019/0.058                       \\ \cline{3-13} 
\multicolumn{1}{|c|}{}                                                  &                                               & \textbf{Gen}                                        & \multicolumn{1}{l|}{0.152/0.212}  & \multicolumn{1}{l|}{0.079/0.139}  & \multicolumn{1}{l|}{0.139/0.185}  & \multicolumn{1}{l|}{0.185/0.272}& 0.093/0.139  & \multicolumn{1}{l|}{0.019/0.038}  & \multicolumn{1}{l|}{0.029/0.048}  & \multicolumn{1}{l|}{0.048/0.077}  & \multicolumn{1}{l|}{0.144/0.173}& 0.010/0.019                        \\ \cline{3-13} 
\multicolumn{1}{|c|}{}                                                  &                                               & \textbf{Loc}                                         & \multicolumn{1}{l|}{0.020/0.026}  & \multicolumn{1}{l|}{0.013/0.033}  & \multicolumn{1}{l|}{0.013/0.020}  & \multicolumn{1}{l|}{0.013/0.020}& 0.020/0.020  & \multicolumn{1}{l|}{0.019/0.029}  & \multicolumn{1}{l|}{0.000/0.019}  & \multicolumn{1}{l|}{0.000/0.010}  & \multicolumn{1}{l|}{0.000/0.019}& 0.010/0.010                          \\ \cline{3-13} 
\multicolumn{1}{|c|}{}                                                  &                                               & \textbf{Port}                                       & \multicolumn{1}{l|}{0.060/0.079}  & \multicolumn{1}{l|}{0.020/0.033}  & \multicolumn{1}{l|}{0.020/0.020}  & \multicolumn{1}{l|}{0.040/0.060}& 0.040/0.053  & \multicolumn{1}{l|}{0.019/0.019}  & \multicolumn{1}{l|}{0.010/0.010}  & \multicolumn{1}{l|}{0.000/0.010}  & \multicolumn{1}{l|}{0.029/0.029}& 0.000/0.000                          \\ \cline{3-13} 
% \multicolumn{1}{|c|}{}                                                  &                                               & \textbf{Port\_Alt}                                   & \multicolumn{1}{l|}{0.152}       & \multicolumn{1}{l|}{0.039}       & \multicolumn{1}{l|}{0.052}       & \multicolumn{1}{l|}{0.046}       & 0.125                            & \multicolumn{1}{l|}{0.086}       & \multicolumn{1}{l|}{0.028}       & \multicolumn{1}{l|}{0.019}       & \multicolumn{1}{l|}{0.028}       & 0.028                            \\ 
\cline{3-13} 
% \multicolumn{1}{|c|}{}                                                  &                                               & \textbf{Port\_Best}                                  & \multicolumn{1}{l|}{0.185}       & \multicolumn{1}{l|}{0.059}       & \multicolumn{1}{l|}{0.066}       & \multicolumn{1}{l|}{0.066}       & 0.132                            & \multicolumn{1}{l|}{0.105}       & \multicolumn{1}{l|}{0.038}       & \multicolumn{1}{l|}{0.019}       & \multicolumn{1}{l|}{0.057}       & 0.028                            \\ 
\cline{2-13} 
\multicolumn{1}{|c|}{}                                                  & \multirow{6}{*}{\textbf{Es}}                  & \textbf{Rel}& \multicolumn{1}{l|}{0.107/0.153}  & \multicolumn{1}{l|}{0.073/0.106}  & \multicolumn{1}{l|}{0.166/0.213}  & \multicolumn{1}{l|}{0.147/0.186}& 0.373/0.493  & \multicolumn{1}{l|}{0.058/0.087}  & \multicolumn{1}{l|}{0.038/0.058}  & \multicolumn{1}{l|}{0.087/0.115}  & \multicolumn{1}{l|}{0.058/0.106}& 0.240/0.337\\ \cline{3-13} 
\multicolumn{1}{|c|}{}                                                  &                                               & \textbf{Gen}                                        & \multicolumn{1}{l|}{0.087/0.256}  & \multicolumn{1}{l|}{0.087/0.106}  & \multicolumn{1}{l|}{0.140/0.173}  & \multicolumn{1}{l|}{0.093/0.146}& 0.220/0.286  & \multicolumn{1}{l|}{0.048/0.087}  & \multicolumn{1}{l|}{0.058/0.087}  & \multicolumn{1}{l|}{0.087/0.115}  & \multicolumn{1}{l|}{0.058/0.087}& 0.163/0.202                          \\ \cline{3-13} 
\multicolumn{1}{|c|}{}                                                  &                                               & \textbf{Loc}                                        & \multicolumn{1}{l|}{0.020/0.026}  & \multicolumn{1}{l|}{0.007/0.026}  & \multicolumn{1}{l|}{0.013/0.020}  & \multicolumn{1}{l|}{0.013/0.020}& 0.020/0.020  & \multicolumn{1}{l|}{0.019/0.029}  & \multicolumn{1}{l|}{0.000/0.019}  & \multicolumn{1}{l|}{0.000/0.010}  & \multicolumn{1}{l|}{0.000/0.019}& 0.010/0.019                            \\ \cline{3-13} 
\multicolumn{1}{|c|}{}                                                  &                                               & \textbf{Port}                                         & \multicolumn{1}{l|}{0.033/0.060}  & \multicolumn{1}{l|}{0.007/0.013}  & \multicolumn{1}{l|}{0.027/0.033}  & \multicolumn{1}{l|}{0.033/0.046}& 0.027/0.040  & \multicolumn{1}{l|}{0.010/0.010}  & \multicolumn{1}{l|}{0.000/0.000}  & \multicolumn{1}{l|}{0.010/0.010}  & \multicolumn{1}{l|}{0.019/0.019}& 0.010/0.019                       \\ \cline{3-13} 
% \multicolumn{1}{|c|}{}                                                  &                                               & \textbf{Port\_Alt}                                   & \multicolumn{1}{l|}{0.120}       & \multicolumn{1}{l|}{0.046}       & \multicolumn{1}{l|}{0.053}       & \multicolumn{1}{l|}{0.033}       & 0.106                            & \multicolumn{1}{l|}{0.134}       & \multicolumn{1}{l|}{0.038}       & \multicolumn{1}{l|}{0.067}       & \multicolumn{1}{l|}{0.028}       & 0.048                            \\ 
\cline{3-13} 
% \multicolumn{1}{|c|}{}                                                  &                                               & \textbf{Port\_Best}                                  & \multicolumn{1}{l|}{0.126}       & \multicolumn{1}{l|}{0.053}       & \multicolumn{1}{l|}{0.073}       & \multicolumn{1}{l|}{0.060}       & 0.113                            & \multicolumn{1}{l|}{0.144}       & \multicolumn{1}{l|}{0.038}       & \multicolumn{1}{l|}{0.076}       & \multicolumn{1}{l|}{0.048}       & 0.057                            \\ 
\hline
\end{tabular}
}
\caption{Comparison of reliability (Rel), generalization (Gen), locality (Loc), and portability (Port) scores for multiple language models evaluated using the ZsRE dataset and the MEMIT editing method. The second column indicates the language in which each model was edited.}
\label{tab:zsre_MMIT}
\end{table*}

\end{document}